\documentclass[preprint,12pt]{elsarticle}

\usepackage{amssymb}
\usepackage[margin=1in]{geometry}
\usepackage{amsmath}
\usepackage{soul}
\usepackage[cal=cm,scr=euler]{mathalpha}
\usepackage{algorithm}
\usepackage{algpseudocode}
\usepackage{booktabs}
\usepackage{multirow}
\usepackage{subfigure}
\usepackage{hyperref}
\biboptions{numbers,sort&compress}

\newcommand{\argmin}{\mathop{\mathrm{argmin}}}
\newcommand{\kl}{\mathop{\mathrm{KL}}}

\usepackage{tikz}
\usepackage{etoolbox}
\definecolor{c0}{RGB}{31, 119, 180}
\definecolor{c1}{RGB}{255, 127, 14}
\definecolor{c3}{RGB}{214, 39, 40}
\definecolor{m0}{HTML}{0072BD}
\definecolor{m1}{HTML}{D95319}

\newrobustcmd*{\myline}[2]{\tikz[baseline=-0.6ex]\draw[#1,thick,#2] (0,0)--(0.5,0);}
\newrobustcmd*{\mycircle}[1]{\tikz{\filldraw[draw=#1,fill=#1] (0,0) circle [radius=0.1cm];}}
\newrobustcmd*{\mytriangle}[1]{\tikz{\filldraw[draw=#1,fill=#1] (0,0)--(0.2cm,0) -- (0.1cm,0.2cm);}}
 
\algdef{SE}[SUBALG]{Indent}{EndIndent}{}{\algorithmicend\ }%
\algtext*{Indent}
\algtext*{EndIndent}

\journal{}

\begin{document}
\begin{frontmatter}
\newpage

\title{Accelerating Hamiltonian Monte Carlo for Bayesian Inference in Neural Networks and Neural Operators}

\author[inst1,inst2,inst3]{Ponkrshnan Thiagarajan}
\author[inst1,inst2]{Tamer A. Zaki}
\author[inst1,inst3]{Michael D. Shields}

\affiliation[inst1]{organization={Hopkins Extreme Materials Institute, Johns Hopkins University},
            city={Baltimore},
            state={MD},
            country={USA}}
\affiliation[inst2]{organization={Department of Mechanical Engineering, Johns Hopkins University},
            city={Baltimore},
            state={MD},
            country={USA}}
\affiliation[inst3]{organization={Department of Civil and Systems Engineering, Johns Hopkins University},
            city={Baltimore},
            state={MD},
            country={USA}}            

\begin{abstract}
Hamiltonian Monte Carlo (HMC) is a powerful and accurate method to sample from the posterior distribution in Bayesian inference. However, HMC techniques are computationally demanding for Bayesian neural networks due to the high dimensionality of the network's parameter space and the non-convexity of their posterior distributions. Therefore, various approximation techniques, such as variational inference (VI) or stochastic gradient MCMC, are often employed to infer the posterior distribution of the network parameters. Such approximations introduce inaccuracies in the inferred distributions, resulting in unreliable uncertainty estimates. In this work, we propose a hybrid approach that combines inexpensive VI and accurate HMC methods to efficiently and accurately quantify uncertainties in neural networks and neural operators.  The proposed approach leverages an initial VI training on the full network. We examine the influence of individual parameters on the prediction uncertainty, which shows that a large proportion of the parameters do not contribute substantially to uncertainty in the network predictions. This information is then used to significantly reduce the dimension of the parameter space, and HMC is performed only for the subset of network parameters that strongly influence prediction uncertainties. This yields a framework for accelerating the full batch HMC for posterior inference in neural networks. We demonstrate the efficiency and accuracy of the proposed framework on deep neural networks and operator networks, showing that inference can be performed for large networks with tens to hundreds of thousands of parameters. We show that this method can effectively learn surrogates for complex physical systems by modeling the operator that maps from upstream conditions to wall-pressure data on a cone in hypersonic flow.
\end{abstract}


\end{frontmatter}

\setcounter{page}{1}
\section{Introduction}
\label{sec:Introduction}
Quantifying uncertainties in machine learning (ML) model predictions is crucial to building confidence in these models. Erroneous predictions without a measure of confidence may lead to catastrophic results in precision-critical applications such as autonomous driving, medical diagnosis, and extreme engineering regimes. Various methods to quantify uncertainties in neural network predictions have been proposed in the literature \cite{kabir2018neural,psaros2023uncertainty, he2023survey}. Among these methods, Bayesian methods are mathematically rigorous and can provide accurate estimates of uncertainties. 

In a Bayesian neural network (BNN), the parameters are stochastic and treated as random variables whose distributions are learned from the training data.  A prior distribution for the parameters is assumed, and the posterior distribution of parameters given training data is inferred using Bayes' rule. The posterior distribution of the parameters is often intractable because it is very high-dimensional (deep neural networks may have thousands to billions of parameters), non-convex, and possesses non-linear dependencies.  Therefore, it often needs to be approximated. Techniques such as variational inference (VI) \cite{graves2011practical,blundell2015weight,hoffman2013stochastic}, Markov chain Monte Carlo methods (MCMC) \cite{neal2011mcmc,chen2014stochastic,zhang2020cyclical},  Monte Carlo dropout \cite{srivastava2014dropout,gal2016dropout,gal2017concrete}, Laplace approximations \cite{mackay1992practical,ritter2018scalable} and ensemble learning \cite{lakshminarayanan2017simple,fort2019deep} methods have been widely used to approximate the posterior distribution of parameters in neural networks. This work focuses on two of the most commonly used techniques to approximate the posterior distribution of parameters in neural networks: MCMC and VI.

MCMC methods comprise a set of algorithms to draw samples from arbitrary and intractable probability distributions. MCMC approaches can converge closely to the true posterior, but are computationally demanding. The potential for the MCMC technique, specifically the Hamiltonian Monte Carlo (HMC) technique, in the context of BNNs was explored by Neal \cite{neal2011mcmc}. Their method performed very well for small BNN models but failed to scale for larger models or datasets. To overcome this computational difficulty, Welling and Teh~\cite{welling2011bayesian} introduced Stochastic Gradient Langevin Dynamics that improved scaling for large models. Following this, multiple stochastic gradient-based MCMC methods have been proposed in the literature, see~\cite{chen2014stochastic,zhang2020cyclical,ahn2012bayesian,ahn2014distributed,ma2015complete,ding2014bayesian,zhang2020amagold,garriga2021exact}. Despite these efficiency improvements, most of these methods are biased due to the lack of the Metropolis-Hastings correction step and subsampling of the data \cite{izmailov2021bayesian}. 

Building on the work of Neal \cite{neal2011mcmc}, Cobb and Jalaian \cite{cobb2021scaling} proposed a new integration scheme by splitting the data into batches that preserve the Hamiltonian. This method allowed posterior sampling using HMC for large datasets without the need to compute stochastic gradients. Given their proven convergence to the true posterior, multiple attempts have been made to understand Bayesian posteriors \cite{izmailov2021bayesian,wenzel2020good} and quantify uncertainties \cite{pasparakis2025bayesian} using HMC. However, the computational cost associated with performing HMC is still a significant challenge, which restricts the use of the method to relatively small networks and data sets.          

In contrast to MCMC methods, VI approximates the posterior distribution by a tractable form using a known family of parameterized distributions, such as the Gaussian. Assuming a simple variational posterior significantly reduces the computational cost of Bayesian inference. In VI, the parameters of the variational posterior are optimized to minimize the dissimilarity between the variational posterior and the true posterior. In this way, expensive MCMC sampling is avoided altogether. Hinton and Van Camp \cite{hinton1993keeping} introduced an information-theoretic approach to learn the posterior distribution of neural network parameters, which is conceptually similar to variational inference (VI). Their method factorized the posterior distribution of the parameters, disregarding correlations between them. Barber and Bishop \cite{barber1998ensemble} later extended this work to incorporate full parameter correlations. Both approaches focused on analytical representations of the optimization objective, limiting their applicability to networks with a single hidden layer. To improve scalability for large networks and datasets, Graves \cite{graves2011practical} proposed a data sub-sampling technique within the VI framework. Additionally, their method optimized the evidence lower bound (ELBO) to estimate the network parameters. Following these developments, several stochastic variational inference methods have been proposed in the literature with various advantages \cite{blundell2015weight,hernandez2015probabilistic,hoffman2013stochastic,kingma2015variational, molchanov2017variational, louizos2017multiplicative, khan2018fast, zhang2018noisy, wu2018deterministic,osawa2019practical,dusenberry2020efficient}.

Despite these advancements, the posterior distribution estimated by VI remains inaccurate due to the fundamental approximations involved in the method. This inaccuracy is evident in the uncertainties quantified in BNN predictions using VI, which tend to be overestimated compared to HMC and often do not reflect basic qualitative behaviors that one would expect from uncertainty predictions, such as an increase in uncertainty in regions of sparse data and a decrease in uncertainty in regions where data are abundant. In summary, HMC offers high accuracy at the expense of high computational cost, while VI provides a more efficient but less reliable alternative.

The present work combines the advantages of both VI and HMC to introduce a hybrid VI-HMC method that is both computationally efficient and accurate.  We hypothesize that the contributions of stochastic network parameters to prediction uncertainties are sparse. In other words, a relatively small number of stochastic network parameters exhibit a strong influence on prediction uncertainty, and many (indeed, a majority) of the parameters have negligible influence. Using this hypothesis, we learn the posterior distribution of the entire set of parameters using VI and examine the contribution of individual network parameters to uncertainty quantification through sensitivity analysis. We then apply HMC to infer the precise distribution of the small fraction of parameters that contribute significantly to prediction uncertainties, thereby reducing the dimension of the parameter space significantly to accelerate HMC.

A few works in the literature have explored accelerating HMC methods for neural networks, and even fewer have focused on combining VI and MCMC techniques.   Salimans et al. \cite{salimans2015markov} proposed a Markov chain variational inference (MCVI) method to include MCMC steps in the VI framework. Their goal, much like this work, was to combine the two methods to obtain accurate and computationally efficient approximations of the posterior distribution. Specifically, MCMC methods are used in MCVI to obtain an unbiased estimate of the gradient of the evidence lower bound (ELBO), which is then utilized in the optimization framework of VI to learn the variational parameters. In doing so, the optimization is improved, providing accurate estimates of the variational parameters while still maintaining the low cost benefit of VI methods. Following this, Wolf et al \cite{wolf2016variational} explored and improved the use of HMC in the MCVI method. They introduced the Metropolis-Hastings acceptance step in the MCVI algorithm, ensuring convergence to the true posterior.  Although these two works combined VI and HMC methods with the same goal as ours, they are fundamentally different from our hybrid approach. Unlike our work that focuses on reducing the computational cost of the HMC method using the information obtained from VI posteriors, MCVI \cite{salimans2015markov,wolf2016variational} focused on improving the accuracy of VI optimization by incorporating MCMC to compute gradients of ELBO.

In terms of accelerating HMC, Patel et al.~\cite{patel2024multi} proposed a multi-fidelity HMC method that computes the acceptance probability using a computationally efficient surrogate model along with an expensive high-fidelity model to evaluate the posterior only when the proposal is accepted. Their approach was more efficient compared to the standard HMC method. Wang and Wibisono \cite{wang2022accelerating} considered accelerating the HMC method by using a time-varying time integration scheme based on the roots of Chebyshev polynomials. Finally, Dhulipala et al.~\cite{dhulipala2023efficient} integrated a Hamiltonian neural network to accelerate time integration in HMC for high-dimensional Bayesian inference with HMC. 

In the broader context of dimensionality reduction in Bayesian inference, several studies have focused on active subspaces methods. Cui et al \cite{cui2014likelihood} proposed a dimensionality reduction technique for Bayesian inverse problems by identifying a likelihood-informed subspace, determined by the relative influence of the prior and the likelihood on the posterior distribution. Constantine et al. \cite{constantine2016accelerating} accelerated MCMC methods by finding an active subspace defined by the gradient of the negative log-likelihood. Following this, Tripathy and Bilionis \cite{tripathy2019deep} proposed a gradient free approach in formulating the active subspaces. Izmailov et al. \cite{izmailov2020subspace}  constructed low-dimensional parameter subspaces for inference in Bayesian deep learning and demonstrated performance improvements on Bayesian neural networks using different inference methods, including VI and No U-turn samplers. Along these lines, Daxberger et al. \cite{daxberger2021bayesian} proposed sampling a subnetwork to preserve the model uncertainties of the full network, where the subnetwork was selected based on the Wasserstein distance between the approximate posterior for the full network and the approximate posterior for the subnetwork. While these efforts, along with other developments in active subspaces \cite{schuster2017exact,ripoli2024sequential, jantre2024learning,cui2022unified}, pursue the common objective of efficient Bayesian inference through dimensionality reduction, the methods they adopt differ from the approach proposed in this work. 

The remaining sections of the manuscript are organized as follows. The VI and HMC methods used for posterior inference in Bayesian networks are briefly introduced in section \ref{sec:VI_and_HMC_intro}. The proposed hybrid approach that combines VI and HMC is described in section \ref{sec:VI_HMC}. The efficiency and accuracy of the proposed VI-HMC method are demonstrated in section \ref{sec:examples} via experiments on Bayesian neural networks (section \ref{sec:BNN_experiments}) and Bayesian operator networks (section \ref{sec:operator_examples}). In sections \ref{sec:bnn1} and \ref{sec:bnn2}, the VI-HMC approach is demonstrated on two Bayesian neural networks to learn sinusoidal functions. A Bayesian operator network learning the solution of the Burgers equation using the hybrid VI-HMC approach is presented in section \ref{sec:burgers_example}, and a Bayesian operator network that maps from upstream conditions to wall-pressure data on a cone in hypersonic flow is demonstrated in section \ref{sec:cone_example}. The computational cost of VI, HMC, and the proposed VI-HMC approach is compared in section \ref{sec:cost_comparison}. Limitations of the VI-HMC method are discussed in section \ref{sec:limitations}, followed by conclusions in section \ref{sec:conclusions}.
\section{Bayesian Inference in Neural Networks} \label{sec:VI_and_HMC_intro}
The goal of a data-driven ML model is to learn a function $\boldsymbol{f}: \mathbb{R}^{d_x} \rightarrow \mathbb{R}^{d_y}$ or an operator $\mathcal{G}: \mathcal{U} \rightarrow \mathcal{V}$ given a set of training data $\mathbb{D}$. Here, $\mathcal{U} $ and $ \mathcal{V}$ are Banach spaces of vector-valued functions.  When learning a function $\boldsymbol{f}$, the training data consists of a set of input-output points,   $\mathbb{D} = \{x_i, y_i\}_{i=1}^{N_d}$, where $x \in \mathbb{R}^{d_x}$ is the vector of inputs and $y \in \mathbb{R}^{d_y}$ is the vector of outputs. Similarly, while learning an operator $\mathcal{G}$, the data consist of input-output pairs, $\mathbb{D} = \{u_i, v_i\}_{i=1}^{N_d}$, where $u \in \mathcal{U}$ is the input function and $v \in \mathcal{V}$ is the output function. The data-driven ML model then learns an approximation $\mathcal{F}_{\boldsymbol{\Theta}}$ parameterized by  $\boldsymbol{\Theta} \subset \mathbb{R}^{N_{\Theta}}$ for the function or the operator described above.

In a stochastic setting, the parameters $\boldsymbol{\Theta}$ are random variables, and we wish to estimate the posterior distribution of the parameters given the training data $\mathbb{D}$, $\mathscr{P}(\boldsymbol{\Theta} | \mathbb{D})$. This can be achieved by applying Bayes' rule
\begin{equation}
    \label{eq:bayes_rule}
    \mathscr{P}(\boldsymbol{\Theta} | \mathbb{D}) = \frac{\mathscr{P}( \mathbb{D} | \boldsymbol{\Theta} ) \mathscr{P}(\boldsymbol{\Theta})}{\mathscr{P}(\mathbb{D}) },
\end{equation}
where $\mathscr{P}( \mathbb{D} | \boldsymbol{\Theta} )$ is the likelihood of observing the data given a set of parameters and $\mathscr{P}(\boldsymbol{\Theta})$ is the prior distribution of the parameters. The term $\mathscr{P}(\mathbb{D})$ is the model evidence, and it is intractable to compute since it involves a high-dimensional integral. Therefore, the posterior distribution of the parameters in Eq.~\eqref{eq:bayes_rule} needs to be approximated using techniques such as VI or sampled using MCMC. These methods are briefly introduced next.

\subsection{Variational Inference (VI)}
In VI, the intractable posterior in Eq.~\eqref{eq:bayes_rule} is approximated by a simpler parametric distribution called the variational posterior $q(\boldsymbol{\Theta}|\lambda)$, where $\lambda$ is the set of parameters defining the distribution $q$. The variational posterior is generally assumed to be from a known family of distributions, such as the Gaussian, where the parameters $\lambda =\{\mu, \sigma\}$ are the mean and standard deviation. Under these assumptions, the goal is to minimize the dissimilarity between the true posterior $\mathscr{P}(\boldsymbol{\Theta}|\mathbb{D})$ and the variational posterior $q(\boldsymbol{\Theta}|\lambda)$ to learn the parameters $\lambda$ that best approximate the true posterior distribution. Thus, variational inference converts the inference problem (inferring the true posterior) into an optimization problem (finding $\lambda$ that best approximates the true posterior).

The most commonly used metric to measure dissimilarity is the Kullback-Leibler (KL) divergence. However, other divergences such as alpha divergence \cite{li2016renyi}, $\chi$ divergence \cite{dieng2017variational}, F-divergence \cite{wan2020f} and Jensen-Shanon divergence \cite{deasy2020constraining,thiagarajan2024jensen} have been proposed in the literature. Using the KL divergence as the dissimilarity metric, the optimization problem can be formulated as 

\begin{equation}
\label{eq:optimization}
    \hat{\boldsymbol{\Theta}} = \argmin_{\boldsymbol{\Theta}} \kl \left(q(\boldsymbol{\Theta}|\lambda) \: || \: \mathscr{P}(\boldsymbol{\Theta}|\mathbb{D}) \right),
\end{equation}
where the KL divergence between the variational and the true posterior is computed as 
\begin{equation}
\label{eq:kl}
    \kl \left( q(\boldsymbol{\Theta}|\lambda) \: || \: \mathscr{P}(\boldsymbol{\Theta}|\mathbb{D}) \right) = \int q(\boldsymbol{\Theta}|\lambda) \log \left[\frac{q(\boldsymbol{\Theta}|\lambda)}{\mathscr{P}(\boldsymbol{\Theta}|\mathbb{D})}\right] d\boldsymbol{\Theta}.
\end{equation}
Using Eq.~\eqref{eq:kl}, the optimization in Eq.~\eqref{eq:optimization} can be written as,
\begin{equation}
    \hat{\boldsymbol{\Theta}} = \argmin_{\boldsymbol{\Theta}} \kl \left( q(\boldsymbol{\Theta}|\lambda) \: || \: \mathscr{P}(\boldsymbol{\Theta})\right) - \mathbb{E}_{q(\boldsymbol{\Theta}|\lambda)} [\log \mathscr{P}(\mathbb{D}|\boldsymbol{\Theta}) ]
\end{equation}
which defines the loss function to be minimized to learn the distribution of parameters of the network. Importantly, this loss function becomes computationally tractable because the first term is the divergence between the variational posterior and the prior, which can be computed analytically for common variational posteriors, and the second term involves a simple expectation that can be efficiently estimated by Monte Carlo sampling from the posterior at each iteration.

\subsection{Hamiltonian Monte Carlo (HMC)}
HMC is an MCMC method in which samples from the posterior distribution of parameters are drawn by solving Hamilton's equations for conservation of energy. 
The Hamiltonian ($\mathcal{H}$) of a system is defined as the sum of its potential energy ($U$) and kinetic energy ($K$), $\mathcal{H} = U + K$. To draw samples from the arbitrary distribution $\mathscr{P}(\boldsymbol{\Theta}|\mathbb{D})$, the potential energy is defined as the negative log of the distribution given by
\begin{equation}
U = - \log \mathscr{P}(\boldsymbol{\Theta}|\mathbb{D}).
\end{equation}
where $\boldsymbol{\Theta}$ plays the role of the position vector in Hamiltonian dynamics. Using Bayes' rule (Eq.~\eqref{eq:bayes_rule}), this can be rewritten as
\begin{equation}
    U(\boldsymbol{\Theta}) = - \log [\mathscr{P}(\mathbb{D}|\boldsymbol{\Theta}) \mathscr{P}(\boldsymbol{\Theta})] + Z
\end{equation}
where $Z = \log \mathscr{P}(\mathbb{D})$ is independent of $\boldsymbol{\Theta}$.
The kinetic energy is then assumed to take the form 
\begin{equation}
    K(\boldsymbol{p}) = \boldsymbol{p}^\top \boldsymbol{M}^{-1} \boldsymbol{p} 
\end{equation}
where $\boldsymbol{p} \in \mathcal{R}^{N_\Theta}$ is a momentum variable and $\boldsymbol{M}$ is an assumed mass matrix. This form for the kinetic energy corresponds to the log probability of the Gaussian distribution on $\boldsymbol{p}$ with zero mean and covariance $\boldsymbol{M}$.  With these definitions of potential and kinetic energy, the Hamiltonian can be expressed as 
\begin{align}
    \label{eq:Hamiltonion}
    \mathcal{H}(\boldsymbol{\Theta},\boldsymbol{p}) &= U(\boldsymbol{\Theta}) + K(\boldsymbol{p}) \\
    & = - \log [\mathscr{P}(\mathbb{D}|\boldsymbol{\Theta}) \mathscr{P}(\boldsymbol{\Theta})] + Z + \boldsymbol{p}^\top \boldsymbol{M}^{-1} \boldsymbol{p} 
\end{align}
The evolution of the position $\boldsymbol{\Theta}$ and momentum $\boldsymbol{p}$ can then be described by Hamilton's equations given as,
\begin{equation}
    \begin{aligned}
        &\frac{\partial \Theta_i}{\partial t} = \frac{\partial \mathcal{H}}{\partial p_i};\\ &\frac{\partial p_i}{\partial t} = -\frac{\partial \mathcal{H}}{\partial \Theta_i}; &i = 1, \ldots,N_\Theta
    \end{aligned}
\end{equation}
These equations can be solved using numerical schemes such as Euler's method or Leapfrog integration. For any time interval $s$, these equations define a mapping $T_s$ from the state at any time $t$ to the state at time $t+s$.

Using the canonical distribution from statistical mechanics, the joint distribution for $\boldsymbol{\Theta}$ and $\boldsymbol{p}$ can be written as
\begin{equation}
    \begin{aligned}
        \mathscr{P}(\boldsymbol{\Theta},\boldsymbol{p}) &= \frac{1}{\mathcal{Z}} \exp \left(\frac{- \mathcal{H}(\boldsymbol{\Theta},\boldsymbol{p})}{T}\right) \\
        & = \frac{1}{\mathcal{Z}} \exp \left(\frac{-U(\boldsymbol{\Theta})}{T}\right)\exp\left(\frac{-K(\boldsymbol{p})}{T}\right) 
    \end{aligned}
\end{equation}
Assuming temperature $T = 1$ and combining all the constants into $\hat{\mathcal{Z}}$, the joint distribution can be expressed 
\begin{align}
\mathscr{P}(\boldsymbol{\Theta},\boldsymbol{p}) & = \frac{1}{\hat{\mathcal{Z}}}\mathscr{P}(\mathbb{D}|\boldsymbol{\Theta}) \mathscr{P}(\boldsymbol{\Theta}) \mathcal N(\boldsymbol{p}|0,\boldsymbol{M})
\end{align}
Thus, sampling from the joint distribution $\mathscr{P}(\boldsymbol{\Theta},\boldsymbol{p})$ yields independent samples from the unnormalized posterior of parameters, $\mathscr{P}(\mathbb{D}|\boldsymbol{\Theta}) \mathscr{P}(\boldsymbol{\Theta})$, and the Gaussian momentum distribution $\mathcal N(\boldsymbol{p}|0,\boldsymbol{M})$. Therefore, drawing samples from the joint distribution $\mathscr{P}(\boldsymbol{\Theta},\boldsymbol{p})$ and discarding the samples of $\boldsymbol{p}$ gives samples from the required unnormalized posterior distribution of parameters $\boldsymbol{\Theta}$. 

To draw samples from the joint distribution $\mathscr{P}(\boldsymbol{\Theta},\boldsymbol{p})$, a Metropolis-Hastings algorithm is used as described in Algorithm~\ref{algo:HMC}. 
\begin{algorithm}
\caption{Algorithm to draw samples using the HMC method}
\label{algo:HMC}
\begin{algorithmic}
\Require  Prior: $\mathscr{P}(\boldsymbol{\Theta}$), Initial $\boldsymbol{\Theta}$
\While {$n < $ number of samples}
 \State $\boldsymbol{p} \gets \mathcal{N}(0,\boldsymbol{M})$ \Comment{Sample momentum from a Gaussian distribution}
 \State $ \boldsymbol{\Theta}^*, \boldsymbol{p}^* \gets T_s(\boldsymbol{\Theta},\boldsymbol{p})$ \Comment{Solve the Hamiltonian equations (Eq.~\eqref{eq:Hamiltonion})}
 \State Perform a Metropolis-Hastings step:
  \Indent
 \begin{align*}
     & r \gets \frac{\exp (-\mathcal{H}(\boldsymbol{\Theta}^*,\boldsymbol{p}^*))}{\exp (-\mathcal{H}(\boldsymbol{\Theta},\boldsymbol{p}))} = \frac{\mathscr{P}(\mathbb{D}|\boldsymbol{\Theta}^*) \mathscr{P}(\boldsymbol{\Theta}^*) \mathcal N(\boldsymbol{p}^*|0,\boldsymbol{M})}{\mathscr{P}(\mathbb{D}|\boldsymbol{\Theta}) \mathscr{P}(\boldsymbol{\Theta}) \mathcal N(\boldsymbol{p}|0,\boldsymbol{M})} &
 \end{align*}
    \State $u \gets U[0,1] $ \Comment{Draw a sample from the uniform distribution}
  \If {$r>u$} 
 \State Accept $ \boldsymbol{\Theta}^*, \boldsymbol{p}^*$
 \State $\boldsymbol{\Theta} \gets \boldsymbol{\Theta}^*$
 \Else
 \State Reject  $ \boldsymbol{\Theta}^*, \boldsymbol{p}^*$
 \EndIf
 \EndIndent
\EndWhile
\end{algorithmic}
\end{algorithm}

\section{Accelerated HMC using VI sensitivities} \label{sec:VI_HMC}
Approximating the posterior distribution using VI is computationally efficient compared to HMC. However, the uncertainties obtained by VI can be erroneous. In this section, we describe a hybrid VI-HMC approach that takes advantage of VI's efficiency while retaining the accuracy of HMC. 

We hypothesize that not all parameters of the network contribute equally to quantifying uncertainties in the neural network predictions and, moreover, that the contribution of each parameter to the prediction uncertainty can be assessed through a sensitivity analysis. Prediction uncertainties can then be effectively quantified by estimating the true joint distribution of a modest number of influential neural network parameters using HMC. The proposed VI-HMC method is conducted in three steps. In the first step, the distribution of the parameters is approximated using VI. In the second step, the sensitivities of the network predictions to each neural network parameter around the mean of the VI parameters are estimated. In the third step, the network parameters are classified into two subsets\textemdash a low sensitivity and a high sensitivity subset\textemdash based on their contribution to the output variance. The parameters in the high-sensitivity subset are sampled using HMC, while the parameters in the low-sensitivity subsets are fixed at their mean values obtained from VI.  This drastically reduces the dimension of the parameter space from which the samples need to be drawn during HMC, thereby accelerating HMC. This process is described in greater detail in Algorithm~\ref{algo:VI_HMC}.

\subsection{Sensitivity to network parameters}
\label{sec:sensitivity}

In VI, a Gaussian distribution $\mathcal{N}(\mu_i,\sigma_i)$ is learned for each network parameter $\theta_i \in \boldsymbol{\Theta}$, where $i, \ldots, N_\Theta$, and $\mu_i$, $\sigma_i$ are the mean and the standard deviation of the parameter, respectively. Let us begin by considering the first-order Taylor series expansion of $\mathcal{F}_{\Theta}(x)$ evaluated at the mean parameter vector $\boldsymbol{\mu}$ given by
\begin{equation} \label{eq:FOT}
    \mathcal{F}_{\Theta}(x) \approx \mathcal{F}_{\boldsymbol{\mu}}(x) + \sum_{i=1}^{N_{\Theta}} (\theta_i - \mu_i) \dfrac{\partial \mathcal{F}_{\boldsymbol{\mu}}(x)}{\partial \theta_i}.
\end{equation}
Here, $\mathcal{F}_{\boldsymbol{\mu}}(x)$ is the neural network or the neural operator evaluated at the mean parameters $\mu$ for an input $x$.
Evaluating the variance of this expression yields
\begin{equation} 
    \text{Var}[\mathcal{F}_{\Theta}(x)] \approx  \sum_{i=1}^{N_{\Theta}} \sigma_i^2 \left(\dfrac{\partial \mathcal{F}_{\boldsymbol{\mu}}(x)}{\partial \theta_i}\right)^2
\end{equation}
This approximates the total variance of the output given uncertainty in the neural network parameters for a given input $x$. 

The average variance over $N_d$ training data can then be written as,
\begin{align}
    \frac{1}{N_d}\sum_{j=1}^{N_d}  \text{Var}[\mathcal{F}_{\Theta}(x_j)] &\approx  \frac{1}{N_d}\sum_{j=1}^{N_d} \sum_{i=1}^{N_{\Theta}} \sigma_i^2 \left(\dfrac{\partial \mathcal{F}_{\boldsymbol{\mu}}(x_j)}{\partial \theta_i}\right)^2 \\
    &= \sum_{i=1}^{N_{\Theta}}\frac{\sigma_i^2}{N_d}\sum_{j=1}^{N_d} \left(\dfrac{\partial \mathcal{F}_{\boldsymbol{\mu}}(x_j)}{\partial \theta_i}\right)^2
\end{align}

We define the sensitivity $S_i^2$ to parameter $\theta_i$ as the contribution of that parameter to the total variance averaged over the $N_d$ training data as
\begin{equation}
    S_{i}^2 = \frac{\sigma_i^2}{N_d}\sum_{j=1}^{N_d} \left(\dfrac{\partial \mathcal{F}_{\boldsymbol{\mu}}(x_j)}{\partial \theta_i}\right)^2
    \label{eqn:sensitivities}
\end{equation}
where the input vector $x_j$ corresponds to the input vector for neural networks, or may be replaced by the input function $u_j$ for neural operators. 

Computing the sensitivities $S_i^2$ from Eq.~\eqref{eqn:sensitivities}, we then rank the parameters according to their sensitivities. Ordering the parameters from most to least sensitive, we then truncate the summation to include $\hat{N}$ parameters based on an established threshold proportion of the variance captured, $\tau$, such that
\begin{equation}
   \hat{N} =  \max \left\{N_T: \dfrac{\sum_{i=1}^{N_T}S_i^2}{\sum_{j=1}^{N_\Theta} S_j^2} \le \tau \right\}.
    \label{eqn:threshold}
\end{equation}
That is, the number of sensitive parameters $\hat N$ is the maximum number that needs to be included in the summation such that the variance ratio is less than or equal to $\tau$. Although a first-order Taylor series expansion is proposed in Eq.\eqref{eq:FOT}, higher order terms can be included, resulting in a modification to the sensitivity scores in Eq.~\eqref{eqn:sensitivities}. These changes can be incorporated into the current framework without modifying the rest of the hybrid VI-HMC approach.

\subsection{HMC in reduced space}
The truncation criterion in Eq.~\eqref{eqn:threshold} implicitly defines a threshold $\hat{S}$ such that each parameter $\theta_i$ having sensitivity $S_i^2$ can be classified into one of two subsets of influential parameters and non-influential parameters as, 
\begin{align*}
    &\boldsymbol{\Theta}_{s} = \{ \theta_i : \theta_i \in \boldsymbol{\Theta} \text{ and } S_i^2 > \hat{S} \}; & \boldsymbol{\Theta}_{\sim s} = \{ \theta_i : \theta_i \in \boldsymbol{\Theta} \text{ and } S_i^2 \leq \hat{S} \}
\end{align*}
The set of parameters $\boldsymbol{\Theta}$ of the network can then be written as the disjoint union of the two subsets 
\begin{equation}
    \boldsymbol{\Theta} = \boldsymbol{\Theta}_{\sim s} \sqcup \boldsymbol{\Theta}_{s}
\end{equation}
Given a distribution of $\boldsymbol{\Theta}_{\sim s}$ from VI, we can find the conditional distribution of $\boldsymbol{\Theta}_{s}$ given the parameters $\boldsymbol{\Theta}_{\sim s}$ are set to their mean values $\boldsymbol{\mu}_{\sim s}$, $\mathscr{P}(\boldsymbol{\Theta}_s | \mathbb{D},\boldsymbol{\Theta}_{\sim s}=\boldsymbol{\mu}_{\sim s})$, using Bayes' rule  as,
\begin{equation}
    \mathscr{P}(\boldsymbol{\Theta}_s | \mathbb{D},\boldsymbol{\mu}_{\sim s}) = \frac{\mathscr{P}( \mathbb{D} | \boldsymbol{\Theta}_s,\boldsymbol{\mu}_{\sim s} ) \mathscr{P}(\boldsymbol{\Theta}_s|\boldsymbol{\mu}_{\sim s})}{\mathscr{P}(\mathbb{D}|\boldsymbol{\mu}_{\sim s}) }
\end{equation}
Samples from the distribution of the sensitive parameters $\mathscr{P}(\boldsymbol{\Theta}_s | \mathbb{D},\boldsymbol{\mu}_{\sim s})$ are then drawn using HMC as shown in Algorithm \ref{algo:VI_HMC}. 

\begin{algorithm}[!ht]
\caption{Algorithm to draw samples using the accelerated HMC}
\label{algo:VI_HMC}
\begin{algorithmic}
\State \textbf{Compute:} $ \mathscr{P}(\boldsymbol{\Theta}|\mathbb{D})$  \Comment{Use VI to obtain approximate distributions of $\boldsymbol{\Theta}$}
\State \textbf{Require:} Variance threshold, $\tau$ \Comment{Proportion of total variance to be included, Eq.~\eqref{eqn:threshold}}
\State \textbf{Compute:} $S_i^2, \hat S$ \Comment{Compute the sensitivities and the sensitivity threshold}
\State \textbf{Construct:} Subsets of parameters $\boldsymbol{\Theta}_{\sim s}, \boldsymbol{\Theta}_{ s}$
\State \textbf{Fix}: $\boldsymbol{\Theta}_{\sim s} \gets \mu_{\boldsymbol{\Theta}_{\sim s}}$ \Comment{Fix the parameters in $\boldsymbol{\Theta}_{\sim s}$ to their means}
\State \textbf{Require:} Prior: $\mathscr{P}(\boldsymbol{\Theta}_s|\boldsymbol{\Theta}_{\sim s})$, Initial $\boldsymbol{\Theta}_s$
\While {$n < $ number of samples}
 \State $\boldsymbol{p} \gets \mathcal{N}(0,\boldsymbol{M})$ \Comment{Sample momentum from a Gaussian distribution}
 \State $ \boldsymbol{\Theta}^*_s, \boldsymbol{p}^* \gets T_s(\boldsymbol{\Theta}_s,\boldsymbol{p})$ \Comment{Solve the Hamiltonian equations (Eq.~\eqref{eq:Hamiltonion})}
 \State Perform a Metropolis-Hastings step:
  \Indent
 \begin{align*}
     & r \gets \frac{\exp (-\mathcal{H}(\boldsymbol{\Theta}^*_s,\boldsymbol{p}^*))}{\exp (-\mathcal{H}(\boldsymbol{\Theta}_s,\boldsymbol{p}))} = \frac{\mathscr{P}(\mathbb{D}|\boldsymbol{\Theta}^*_s,\boldsymbol{\Theta}_{\sim s}) \mathscr{P}(\boldsymbol{\Theta}^*_s|\boldsymbol{\Theta}_{\sim s}) \mathcal N(\boldsymbol{p}^*|0,\boldsymbol{M})}{\mathscr{P}(\mathbb{D}|\boldsymbol{\Theta}_s,\boldsymbol{\Theta}_{\sim s}) \mathscr{P}(\boldsymbol{\Theta}_s|\boldsymbol{\Theta}_{\sim s}) \mathcal N(\boldsymbol{p}|0,\boldsymbol{M})} &
 \end{align*}
    \State $u \gets U[0,1] $ \Comment{Draw a sample from the uniform distribution}
  \If {$r>u$} 
 \State Accept $ \boldsymbol{\Theta}^*_s, \boldsymbol{p}^*$
 \State $\boldsymbol{\Theta}_s \gets \boldsymbol{\Theta}^*_s$
 \Else
 \State Reject  $ \boldsymbol{\Theta}^*_s, \boldsymbol{p}^*$
 \EndIf
 \EndIndent
\EndWhile
\end{algorithmic}
\end{algorithm}
\subsection{Computational Benefits} \label{sec:computational_benefits}
The computational cost associated with drawing an independent sample from a $D$-dimensional distribution is approximately $\mathcal{O}(D^{5/4})$, considering the level of allowable error accumulation in the numerical integration of the Hamiltonian equations to achieve a reasonable acceptance rate \cite{neal2011mcmc,hoffman2014no}. 
Therefore, reducing the dimension of the neural network's parameter space will result in significant computational savings, which is the goal of the proposed hybrid VI-HMC approach. For example, reducing the parameter dimension from $10^4$ to $10^3$ corresponds to a nearly 20x reduction in expected computational cost. In other words, thanks to the reduced dimensionality, the Hamiltonian equations can be solved with a larger step size in the proposed VI-HMC approach, resulting in fewer time integration steps to convergence. 

However, it is important to note that the hybrid VI-HMC approach does not change the cost per time-integration step in solving the Hamiltonian equations. The two main factors contributing to the cost of a single time integration step are the forward pass in the neural network and the gradient computation. Since the model complexity itself is not reduced in the hybrid approach, i.e.\, the number and size of the hidden layers remain the same, the cost for the forward pass and the gradient computation remain unchanged, yielding no change in the computational cost per integration step due to the reduced dimensions.  Additional cost savings could potentially be gained by circumventing gradient computations for the non-influential parameters, but this would require a fundamental rewriting of existing neural network training codes, which is challenging. 
 
\section{Numerical Examples} \label{sec:examples}
This section presents four numerical experiments that demonstrate the advantages of the proposed VI-HMC method. The first and second use BNNs to approximate functions from sparse data. The third and fourth examples apply Bayesian neural operators to learn complex partial differential equations. All numerical experiments were implemented in PyTorch \cite{paszke2019pytorch}. VI layers are implemented following \cite{shridhar2019comprehensive}. The HMC algorithm uses the hamiltorch\cite{cobb2021scaling} package, and the VI-HMC method is implemented by building on it. All the hyper-parameters used in the experiments are summarized in \ref{app:hyperparams}.

Code to implement the proposed VI-HMC approach can be found at this GitHub repository: \url{https://github.com/ponkrshnan/VI-HMC.git}

\subsection{Bayesian Neural Networks} \label{sec:BNN_experiments}
In this example, a BNN is trained to learn a simple noisy function of the form
\begin{align}
    y = a \sin (\omega_1 x + \phi_1) + b \sin  (\omega_2 x + \phi_2) + \epsilon
    \label{eq:sin_cos}
\end{align}
where $\epsilon$ is a zero mean Gaussian noise, $\epsilon \sim \mathcal{N}(0,\sigma_d)$. Two cases with different network architectures and function parameters ($a, b, \omega_1, \omega_2, \phi_1, \phi_2$) are considered to generate the data to study the proposed approach.

\subsubsection{Case I:} \label{sec:bnn1}
In the first case, the training data are generated using the parameters shown in Table~\ref{Tab:dnn_comp1} along with the noise parameter $\sigma_d = 1E-3$. Twenty points with $x \in [-1,-0.2] \cup [0.2,1]$ are generated for training (with an intentional gap in the range $(-0.2, 0.2)$) and 300 points with $x \in [-1.2,1.2]$ are generated for validation.
A single hidden-layer neural network with two neurons and a sine activation function is considered. With the bias added to the hidden layer, the network has six parameters to learn that mimic the dimension of the parameterization in Eq.~\eqref{eq:sin_cos}. 

\begin{figure}[!ht]
    \centering
    \subfigure[]{\includegraphics[width=0.49\linewidth]{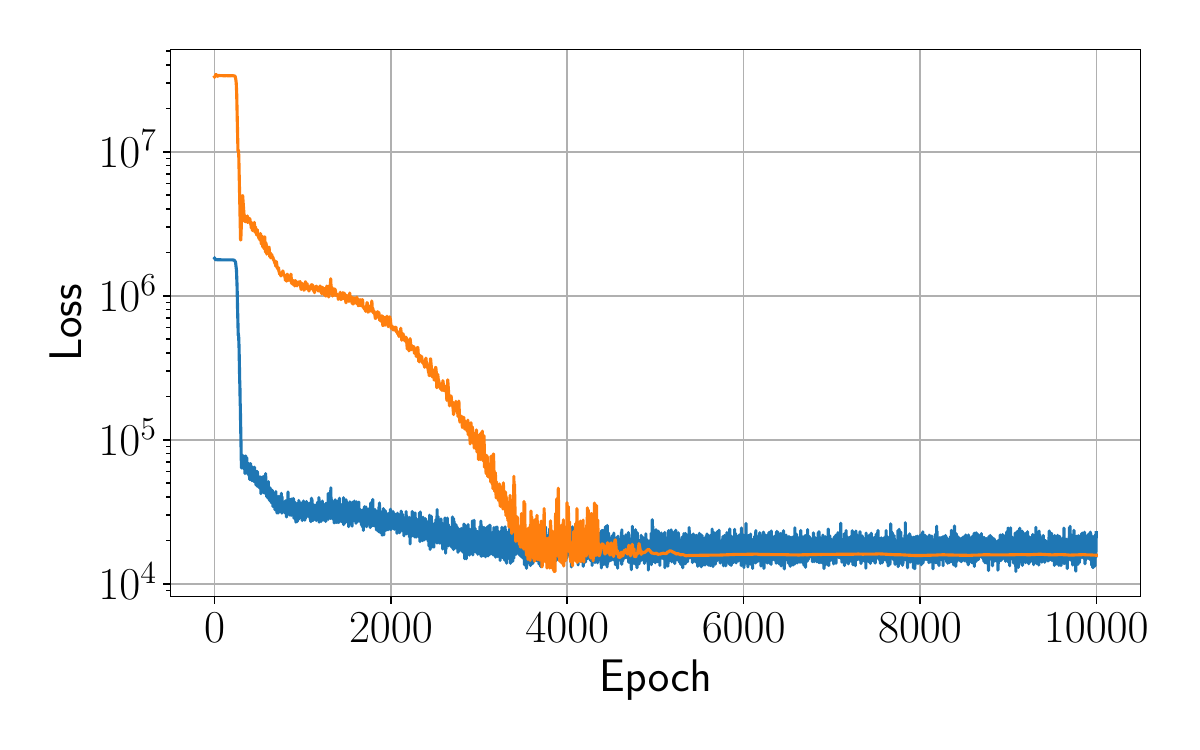}}
    \subfigure[]{\includegraphics[width=0.49\linewidth]{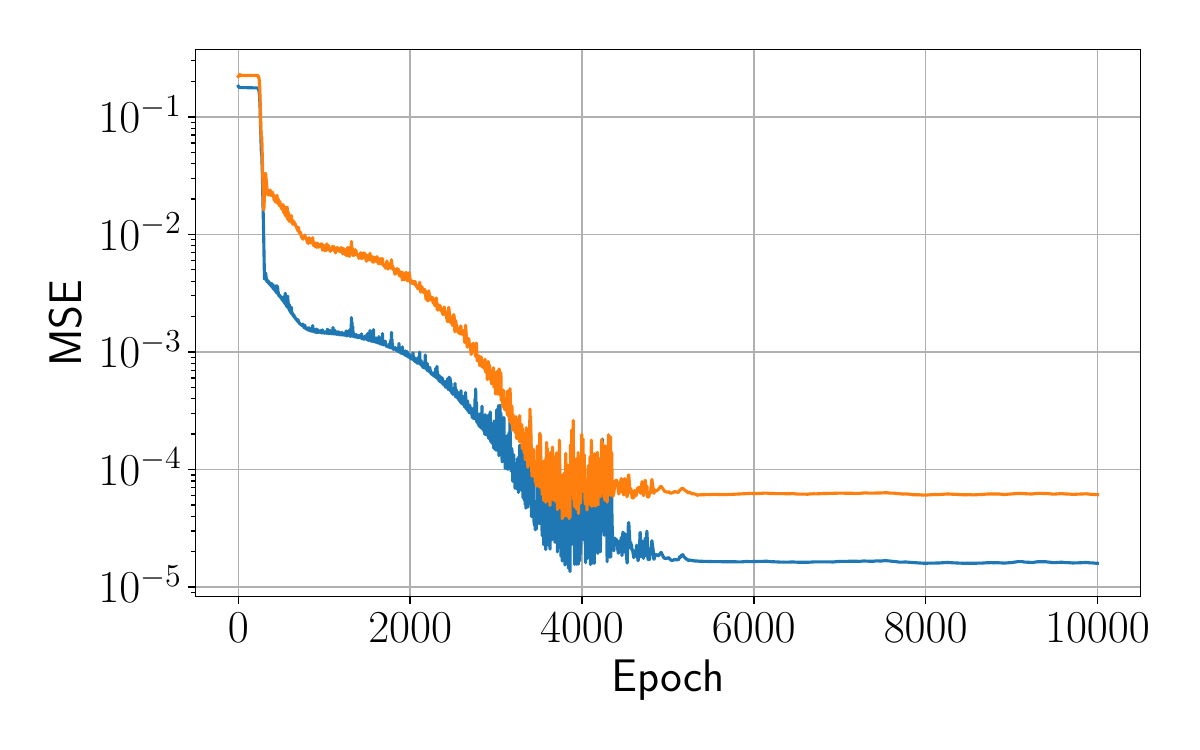}}
    \caption{Case I: Evolution of (a) ELBO loss function and (b) Mean square error in VI training. \myline{c0}{} Training, \myline{c1}{}Validation. }
    \label{fig:loss_convergence_case1}
\end{figure}

\begin{figure}[!ht]
    \centering
    \includegraphics[width=0.5\linewidth]{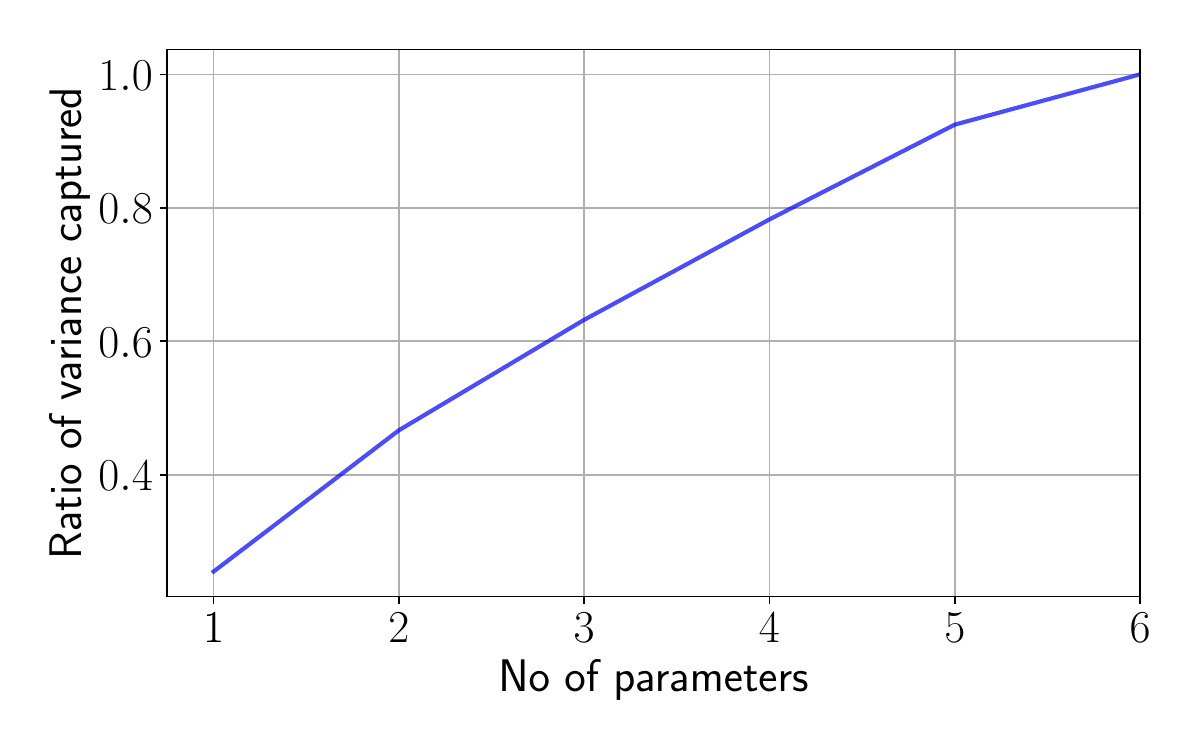}
    \caption{Case I: Ratio of variance captured for increasing number of BNN parameters. }
    \label{fig:sensitivity_case1}
\end{figure}

In the first step, the posterior distribution of the parameters is learned using VI with a normal variational distribution. The network is trained until the ELBO loss converges, as seen in Figure~\ref{fig:loss_convergence_case1}(a). Also, shown in Figure~\ref{fig:loss_convergence_case1}(b) is the mean square error (MSE) convergence during VI training. Subsequently, the neural network parameter sensitivities are estimated as described in Section~\ref{sec:sensitivity}. Figure~\ref{fig:sensitivity_case1} shows the proportion of variance in the output captured for an increasing number of neural network parameters. The first four parameters are sufficient for a sensitivity threshold of $\tau = 0.9$. Thus, in the VI-HMC step, the sample space reduces from 6 parameters to 4. The four most sensitive parameters are sampled using HMC, and the other two are fixed to their mean values obtained from VI.

The results obtained from learning the parameters using VI and standard HMC are compared with the results from the proposed VI-HMC method in Table~\ref{Tab:dnn_comp1}. We observe that the parameters sampled through VI-HMC closely match those obtained using the full HMC approach. Note that the phase angles sampled by HMC are $\phi_1 \approx 2\pi$, which is the same as $\phi_1 = 0$, and $\phi_2 \approx 2.5 \pi$, which is the same as $\phi_2 = 0.5 \pi$.Also, note that VI-HMC uses HMC to infer the distributions of the parameters $\omega_1$, $\phi_2$, $a$, and $b$ and selects $\omega_2$ and $\phi_1$ as their mean values from VI because their distributions contribute very little to the variance in the overall prediction. The sensitivity rankings of the parameters are verified by computing analytical derivatives of the sinusoidal function in Eq.~\eqref{eq:sin_cos}  with respect to the six parameters and computing the sensitivity scores according to Eq.~\eqref{eqn:sensitivities}.
\begin{table}[!ht]
    \centering
    \caption{Parameters learned via VI, HMC, and Hybrid VI-HMC}
    \begin{tabular}{ccccccc}
    \toprule
        \textbf{Method} & $\boldsymbol{\omega_1}$ & $\boldsymbol{\omega_2}$ & $\boldsymbol{\phi_1}$ & $\boldsymbol{\phi_2}$ & $\boldsymbol{a}$& $\boldsymbol{b}$  \\ \midrule
        True  &  4 & -3 & 0 & 1.57 & 0.4 & 0.5 \\ \midrule 
        \multirow{ 2}{*}{VI} &  -4.05   & 2.94 & -0.072   & -1.61   & -0.39   & -0.48   \\
         &$\pm$0.068 &$\pm$0.059 & $\pm$0.030 & $\pm$0.035 & $\pm$0.019 & $\pm$0.020  \\ \midrule   
         \multirow{ 2}{*}{HMC} & 4.00 & -3.00 & 6.29 & 7.86 & 0.40& 0.50  \\
         & $\pm$7.3E-3& $\pm$3.8E-3 & $\pm$5.3E-3 & $\pm$5.8E-3 & $\pm$2.6E-3 & $\pm$1.6E-3 \\ \midrule
         \multirow{ 2}{*}{VI-HMC} & -4.00 & 2.94 & -0.072 & -1.57  & -0.40 & -0.48  \\
         & $\pm$7.1E-3 & - & - & $\pm$5.8E-3 & $\pm$2.4E-3 & $\pm$4.1E-4 \\
        \bottomrule
    \end{tabular}
    \label{Tab:dnn_comp1}
\end{table}

\begin{figure}[!ht]
    \centering
    \subfigure[]{\includegraphics[width=0.49\linewidth]{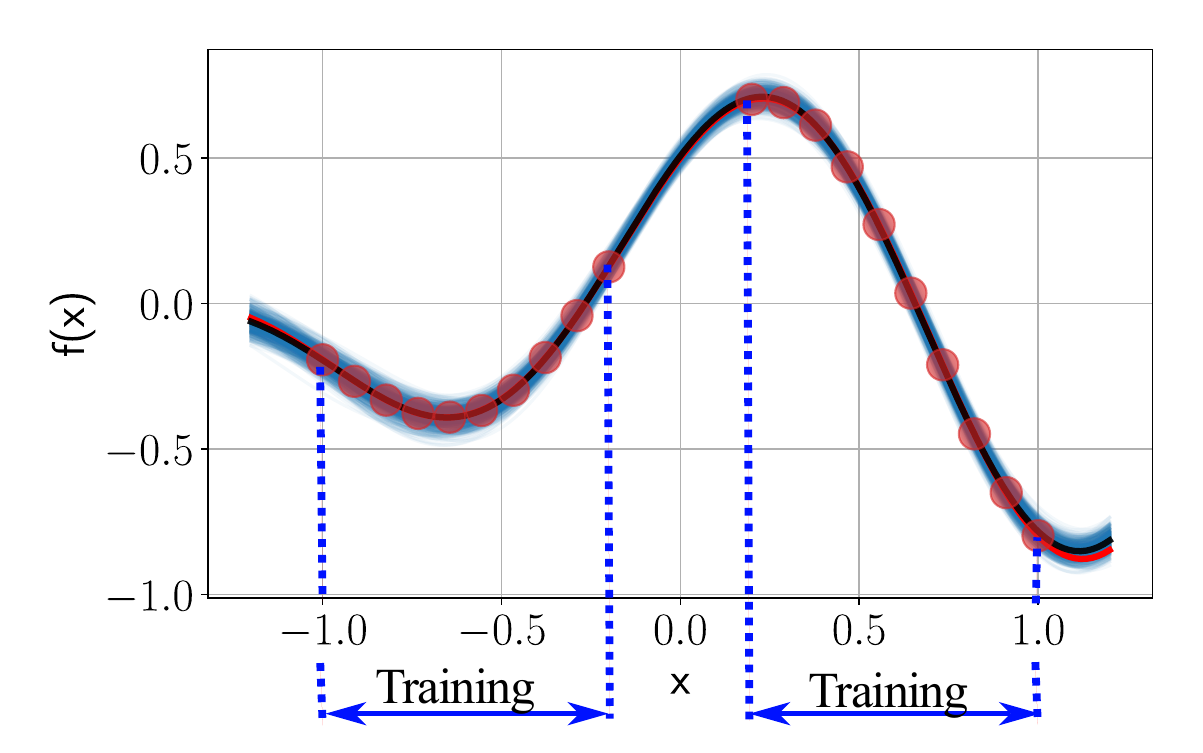}}
    \subfigure[]{\includegraphics[width=0.49\linewidth]{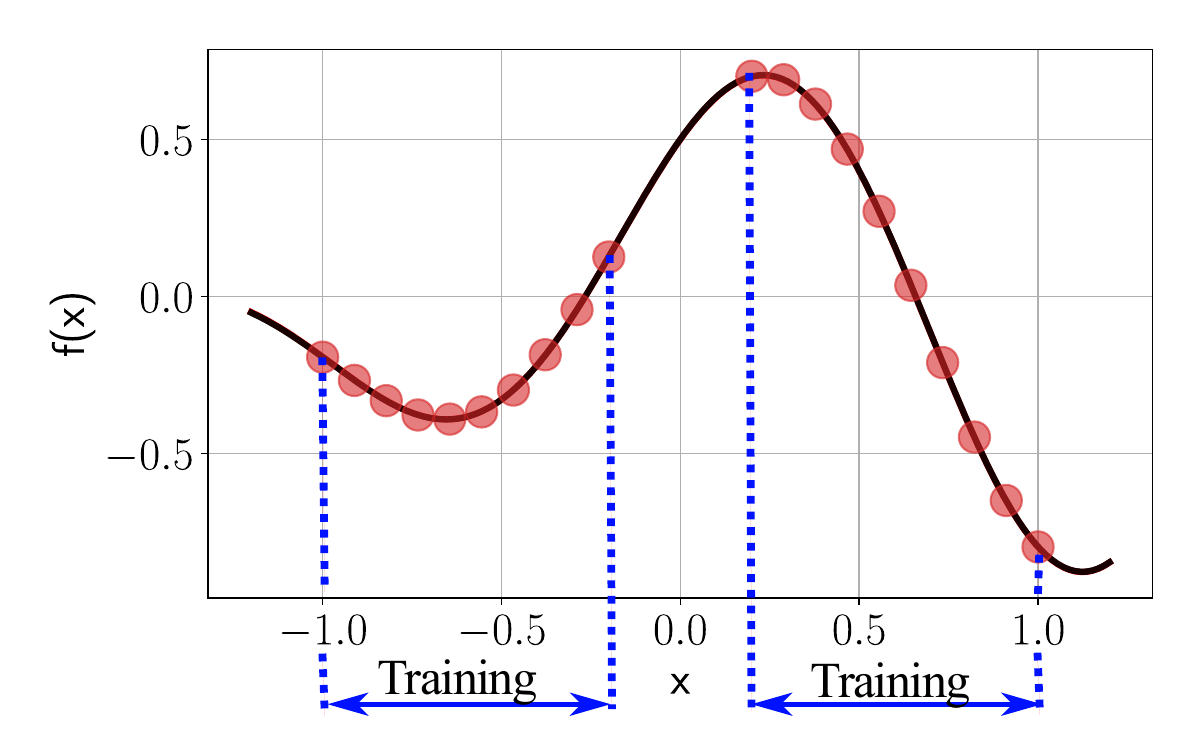}}
    \subfigure[]{\includegraphics[width=0.49\linewidth]{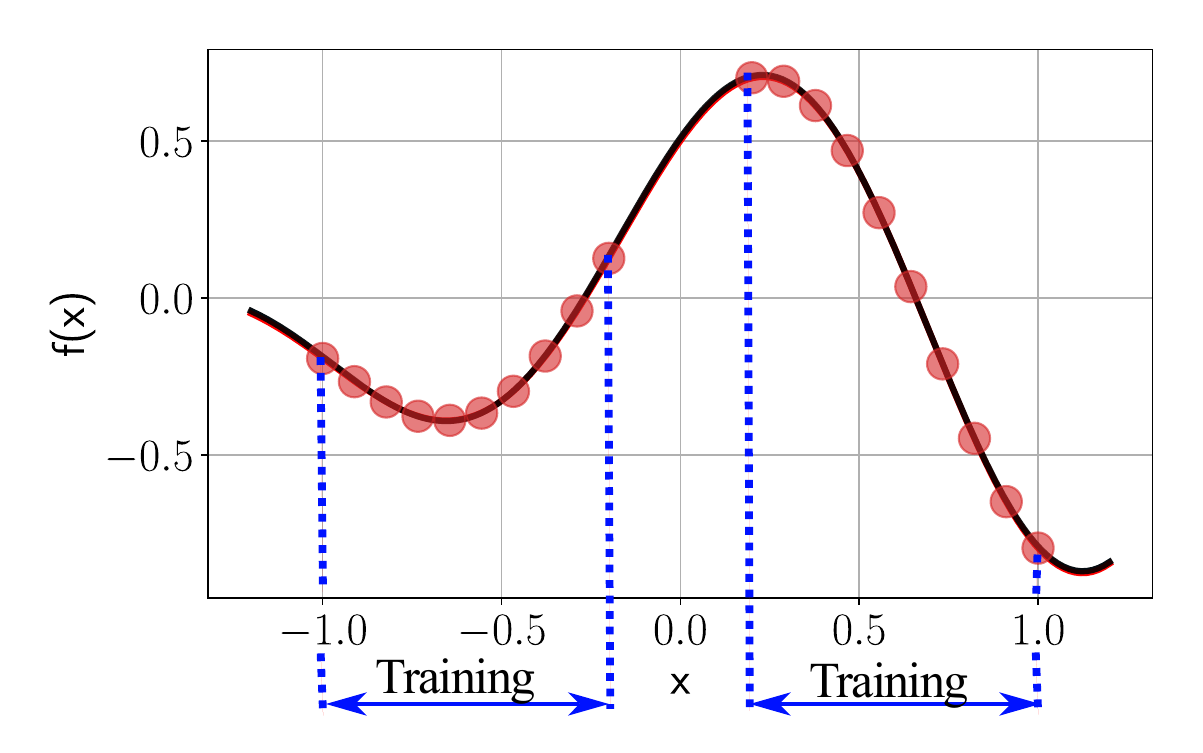}}
    \caption{Case I: BNN predictions and their uncertainties from (a) VI, (b) HMC, and (c) VI-HMC. \myline{c0}{}samples, \myline{black}{} mean prediction, \mycircle{c3!60}{ training data}, \myline{red}{}true function.}
    \label{fig:pred_comp_case1}
\end{figure}

BNN predictions and associated uncertainties obtained by VI, HMC, and VI-HMC are presented in Figure~\ref{fig:pred_comp_case1}. Note that VI overestimates the uncertainties compared to HMC, whereas the uncertainty estimates of VI-HMC closely resemble the benchmark uncertainties obtained from HMC. Note also that the VI uncertainties are unrealistic in that they show high uncertainty even in regions with abundant data.

\begin{figure}[!htb]
    \centering
    \subfigure[]{\includegraphics[width=0.45\linewidth]{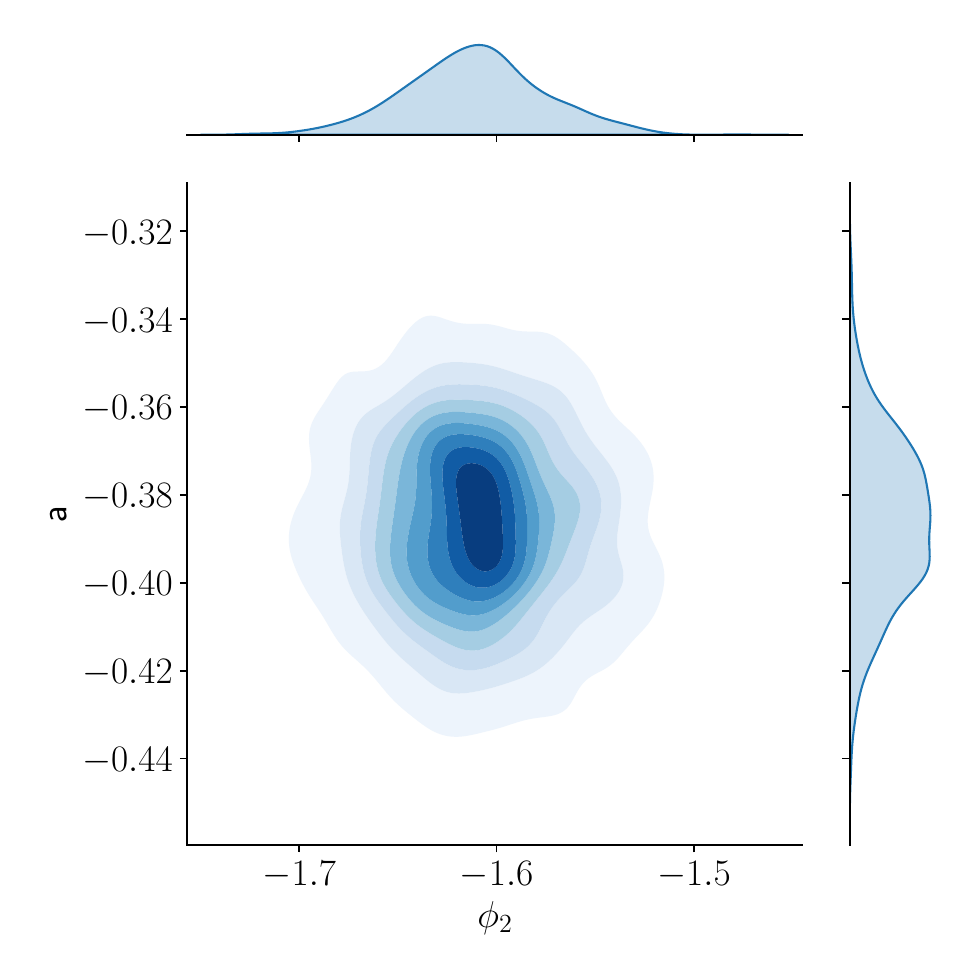}}
    \subfigure[]{\includegraphics[width=0.45\linewidth]{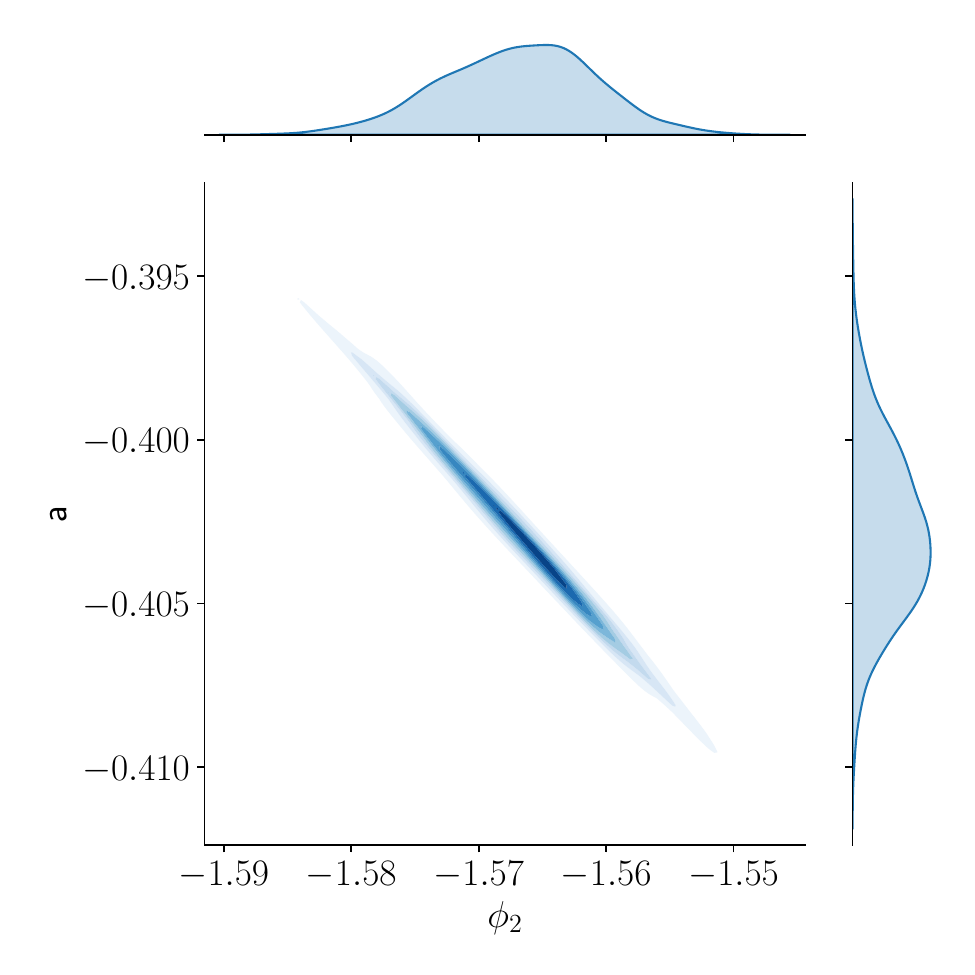}}
    \subfigure[]{\includegraphics[width=0.45\linewidth]{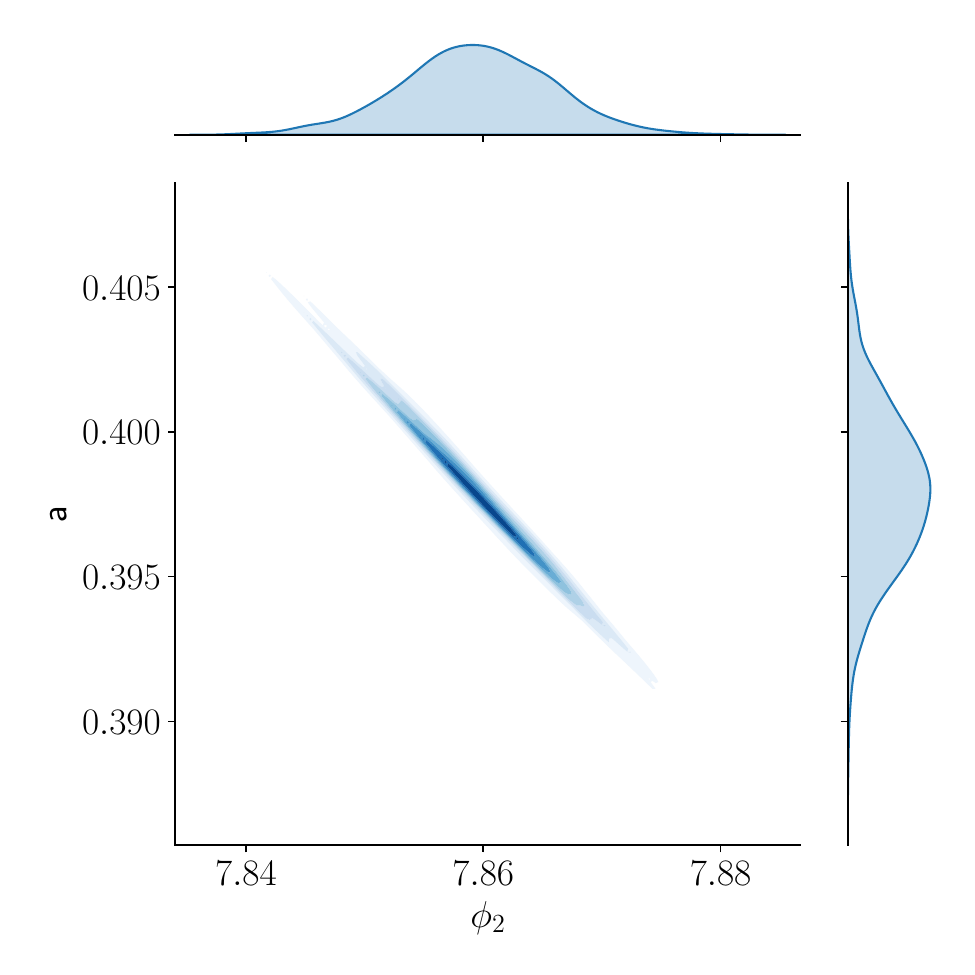}}
    \caption{Joint distributions of two parameters ($\phi_2$ and $a$) obtained by (a) VI, (b) VI-HMC, (c) HMC.}
    \label{fig:joint_dist}
\end{figure}

Finally, the joint distributions between two parameters $(\phi_2, a)$ obtained by VI, VI-HMC, and HMC are compared in Figure~\ref{fig:joint_dist}. Since the mean field approximation in VI assumes that the parameters of the network are independent, they show no correlation in Figure~\ref{fig:joint_dist}(a). However, the true joint distribution between these two parameters obtained by HMC shows that the parameters are highly correlated in Figure~\ref{fig:joint_dist}(c). The VI-HMC approach reflects this correlation structure in  Figure~\ref{fig:joint_dist}(b) while closely approximating the joint distribution obtained by HMC.

\subsubsection{Case II} \label{sec:bnn2}
\begin{figure}[!ht]
    \centering
    \subfigure[]{\includegraphics[width=0.49\linewidth]{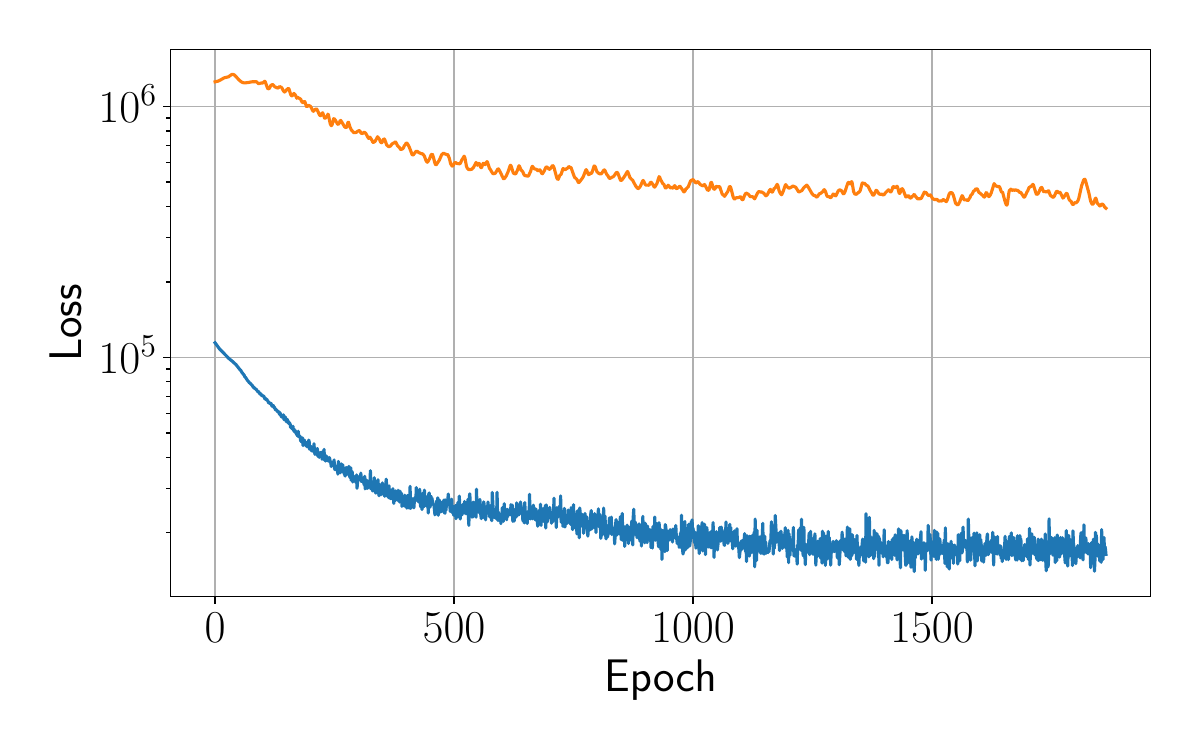}}
    \subfigure[]{\includegraphics[width=0.49\linewidth]{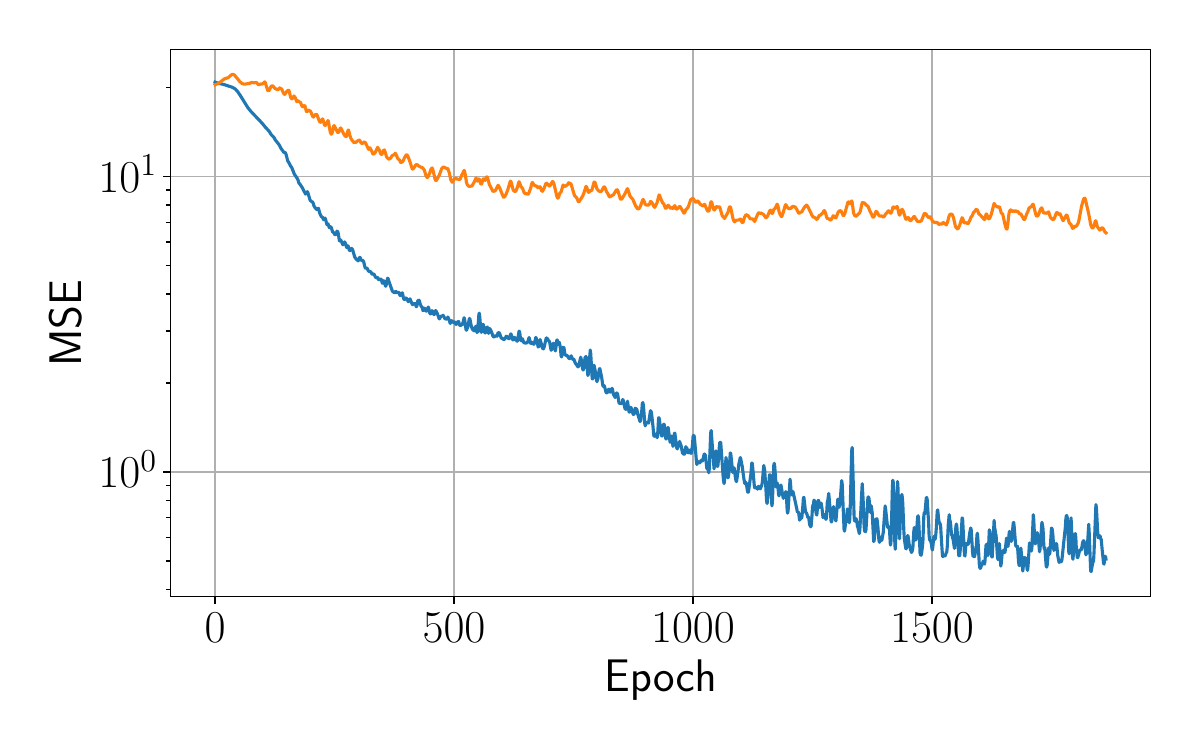}}
    \caption{Case II: Evolution of (a) ELBO loss function and (b) Mean square error in VI training. \myline{c0}{}Training, \myline{c1}{}Validation.}
    \label{fig:case2_loss}
\end{figure}

\begin{figure}[!ht]
    \centering
    \subfigure[]{\includegraphics[width=0.49\linewidth]{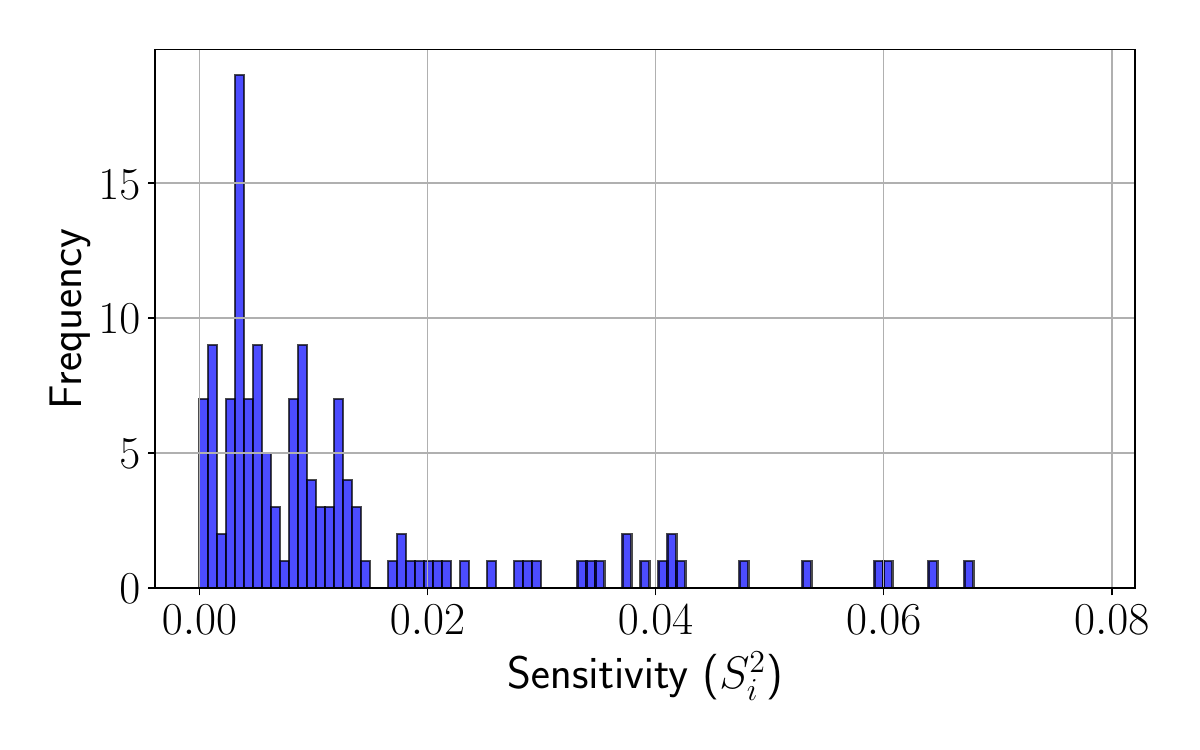}}
    \subfigure[]{\includegraphics[width=0.49\linewidth]{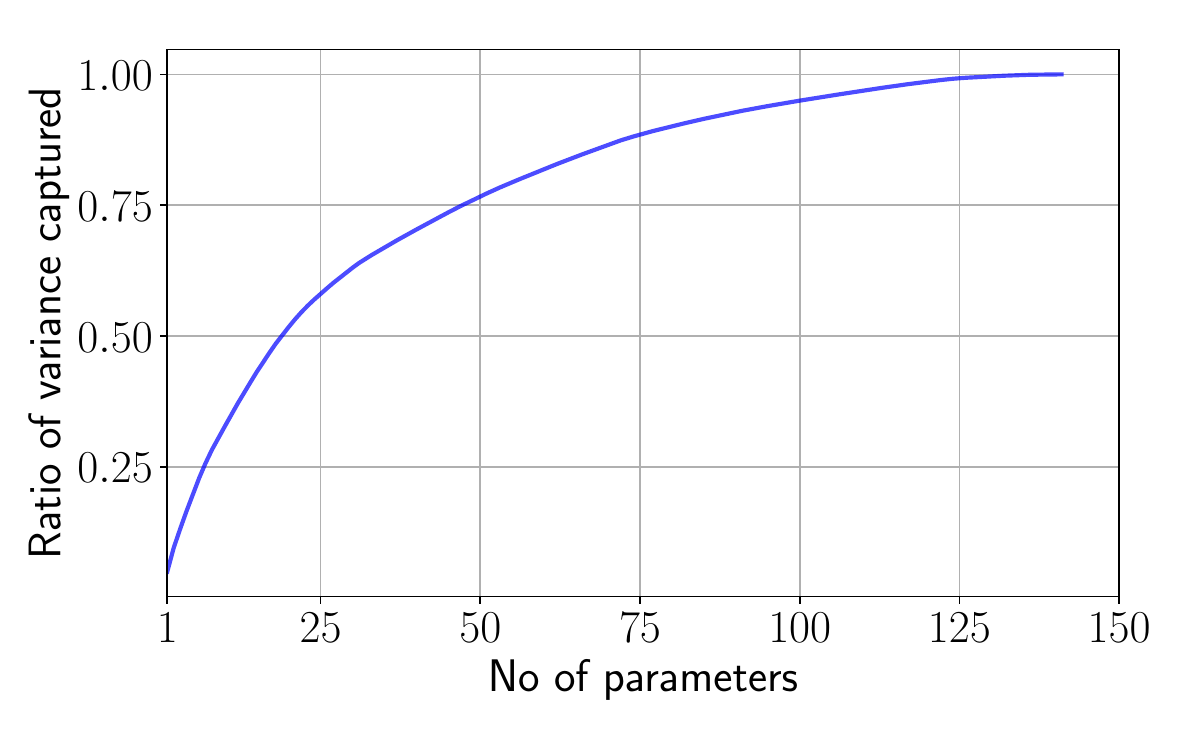}}
    \caption{Case II: BNN parameter sensitivities (a) Histogram of sensitivity values for the BNN parameters (x-axis is limited to the 99th percentile of $S_i^2$ for clarity). (b) Percentage of captured variance for an increasing number of BNN parameters.}
    \label{fig:case2_sensitivity}
\end{figure}

In this example, the parameters for Eq.~\eqref{eq:sin_cos} used to generate the training and test data are: $\omega_1 = 4, \omega_2 = -12, \phi_1= 0, \phi_2 = \pi/2, a =4, b=5, \sigma_d = 0.05 $. Again, twenty points with $x \in [-1,-0.2] \cup [0.2,1]$ are generated for training and 300 points with $x \in [-1.2,1.2]$ are generated for validation. In this case, a BNN with two hidden layers of 10 neurons each is considered with tanh activation functions, resulting in a total of 141 neural network parameters. 
The ELBO loss and MSE convergence during VI training are shown in Figure~\ref{fig:case2_loss}. Since the network is over-parameterized, the training loss is much lower than the validation loss. Once the posterior distribution is obtained from VI, the sensitivities of the parameters are analyzed to determine the most sensitive parameters to quantify uncertainties in the outputs. The histogram of sensitivities for the parameters and the percentage of variance captured are shown in Figure~\ref{fig:case2_sensitivity}. Here, the 79 parameters with the highest sensitivities are required to capture 90\% of the total variance in the network prediction.  Although the true function can be represented by a sine activation function with just six parameters, as seen in case I, the introduction of tanh as the activation function necessitates about 80 parameters to approximate the true function.  This case demonstrates that a meaningful reduction in the parameter space is achieved even for a small network size, such that, in the VI-HMC step, the parameter space to sample from reduces to about 56\% from the original space.

In the VI-HMC sampling, the 62 least sensitive parameters are fixed to their mean values obtained from VI. The remaining 79 parameters are sampled and compared with traditional HMC on the full parameter space. Ten parallel chains are initialized, and $10^4$ samples from each chain are drawn for both HMC and VI-HMC.

\begin{figure}[!ht]
    \centering
    \subfigure[]{\includegraphics[width=0.49\linewidth]{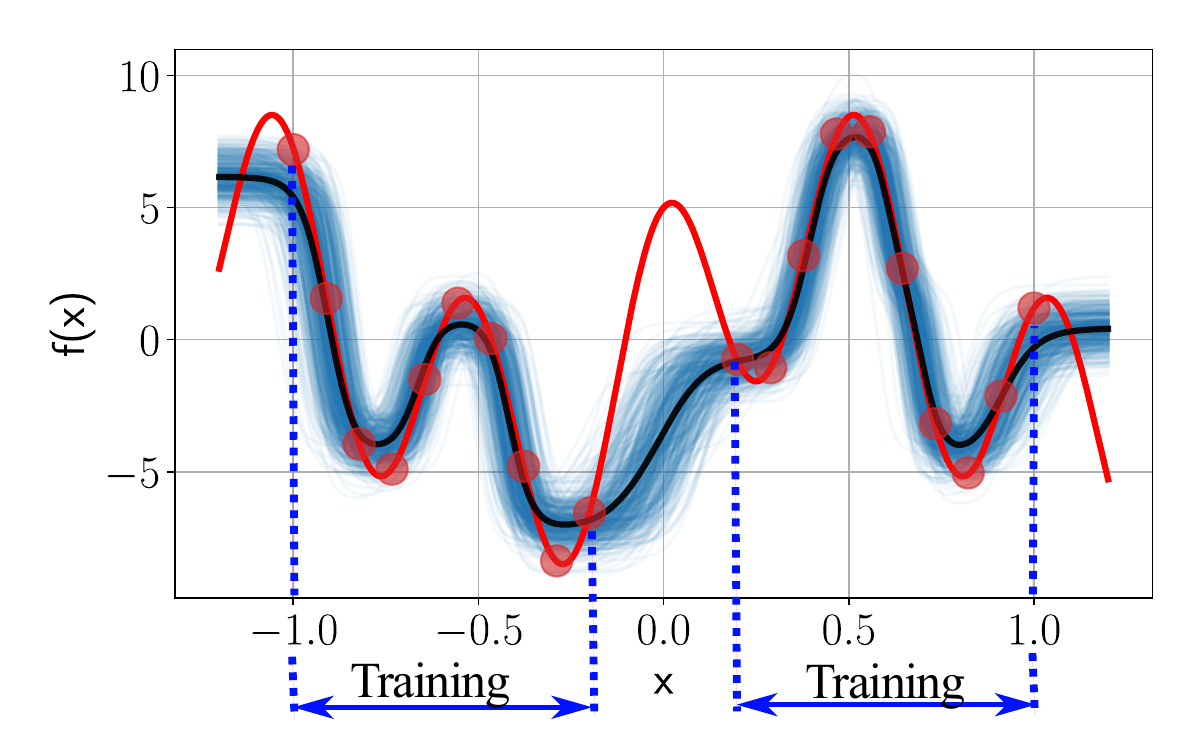}}
    \subfigure[]{\includegraphics[width=0.49\linewidth]{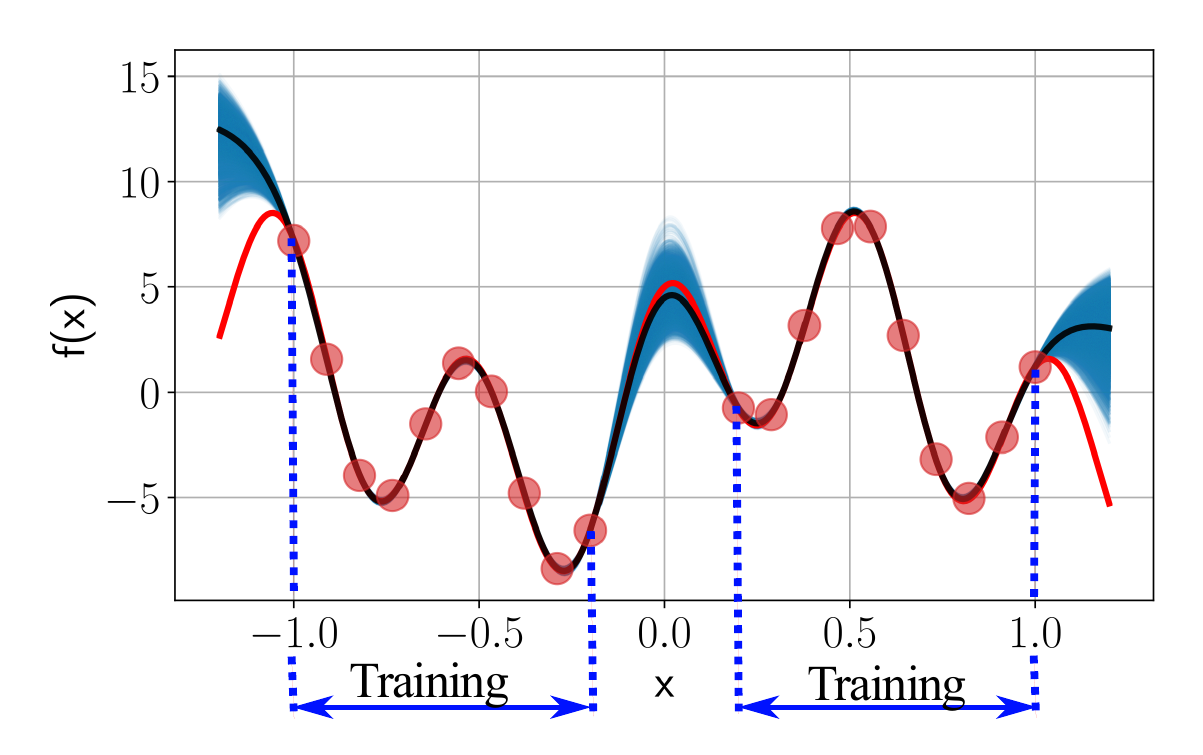}}
    \subfigure[]{\includegraphics[width=0.49\linewidth]{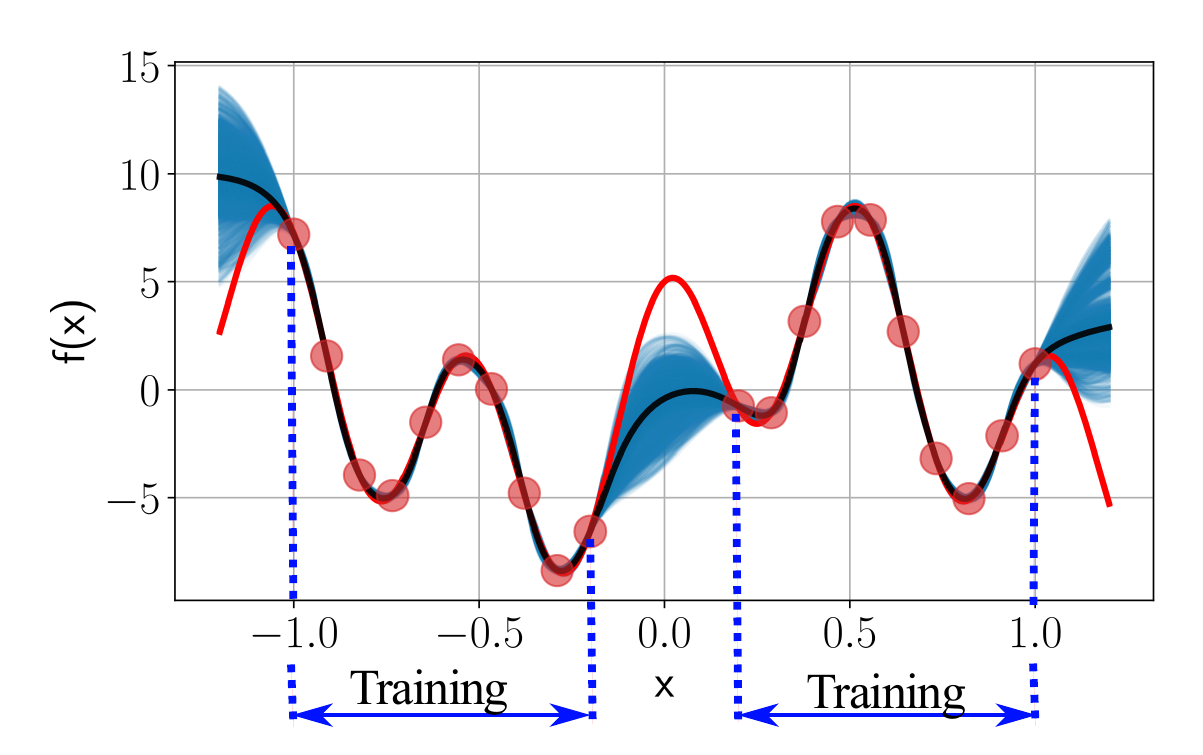}}
    \caption{ Case II: BNN predictions and their uncertainties from (a) VI, (b) HMC, and (c) VI-HMC.  \myline{c0}{}samples, \myline{black}{} mean prediction, \mycircle{c3!60}{ training data}, \myline{red}{}true function.}
    \label{fig:case2_predictions}
\end{figure}

The predictions obtained by VI, HMC, and VI-HMC are compared in Figure~\ref{fig:case2_predictions}. Compared to HMC, VI again overestimates the uncertainties where training data is present and underestimates the uncertainty in the interpolation region $x \in (-0.2,0.2)$ and the extrapolation regions $x \in [-1.2,-1) \cup (1,1.2]$. VI-HMC, on the other hand, provides uncertainties that better match the uncertainties from HMC. They are qualitatively better in their lower uncertainties near existing training data, while also yielding larger and more realistic uncertainties in the interpolation and extrapolation regimes. 
Further, since some parameters are fixed to be the means obtained from VI, the mean prediction of VI-HMC combines the mean predictions from VI and HMC. This effect can be seen in the interpolation regime $x \in (-0.2,0.2)$. In addition to these results, the robustness of the sensitivity scores for varying noise levels and training dataset size is discussed in \ref{app:robustness}

\subsection{Bayesian Neural Operators} \label{sec:operator_examples}
In this section, the VI-HMC method is applied to learn Bayesian neural operators. The neural operators are developed as Deep operator networks (DeepONets)~\cite{lu2021learning}, with uncertain parameters in both the branch and trunk networks.

\subsubsection{Burgers' equation} \label{sec:burgers_example}
In the first example, a DeepOnet (with architecture described in \ref{sec:App_Architecture}) is trained to learn the solution of the Burgers' equation benchmark investigated in \cite{li2020fourier}.  The 1D Burgers' equation is given by 
\begin{equation}
    \frac{\partial s}{\partial t} + s \frac{\partial s}{\partial x} - \nu \frac{\partial ^2 s}{\partial x^2} = 0, \quad (x,t) \in (0,1) \times (0,1] 
\end{equation}
A periodic boundary condition is considered here, such that
\begin{align}
    s(0,t) &= s(1,t) \\
    \frac{\partial s}{\partial x}(0,t) &= \frac{\partial s}{\partial x} (1,t)
\end{align}
and the initial condition is specified as $s(x,0)  = u(x)$ where $u(x)$ is a Gaussian random field with the covariance function defined in \cite{li2020fourier}. The viscosity is set to $\nu = 0.01$. A total of 2000 Gaussian random fields are generated as initial conditions, and the Burgers equation is solved using a split-step method on a $101\times101$ grid of points in space-time using the code provided by \cite{li2020fourier}. Once the solutions are obtained for the 2000 random initial conditions, the dataset is split 50-50\% into training and validation sets. Thus,  the training and validation datasets contain the solution at $1000\times101\times101$ points each (initial condition $\times$ grid points in space $\times$ grid points in time).  

\begin{figure}[!ht]
    \centering
    \subfigure[]{\includegraphics[width=0.49\linewidth]{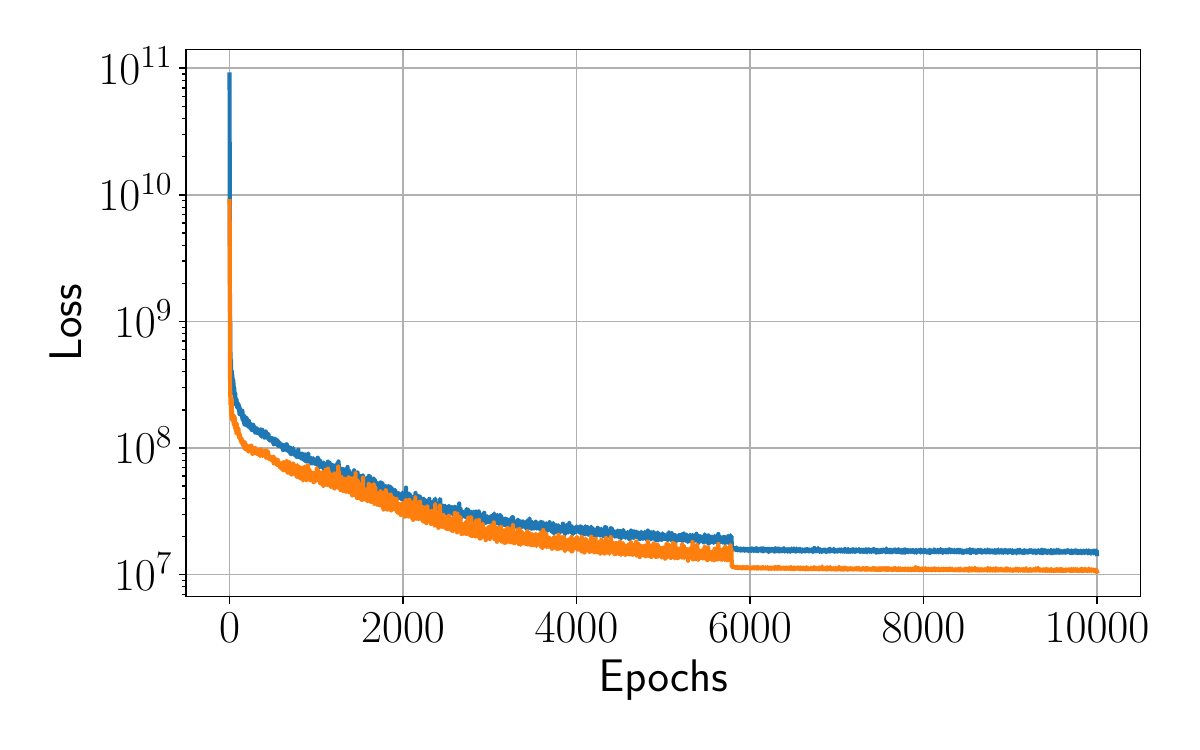}}
    \subfigure[]{\includegraphics[width=0.49\linewidth]{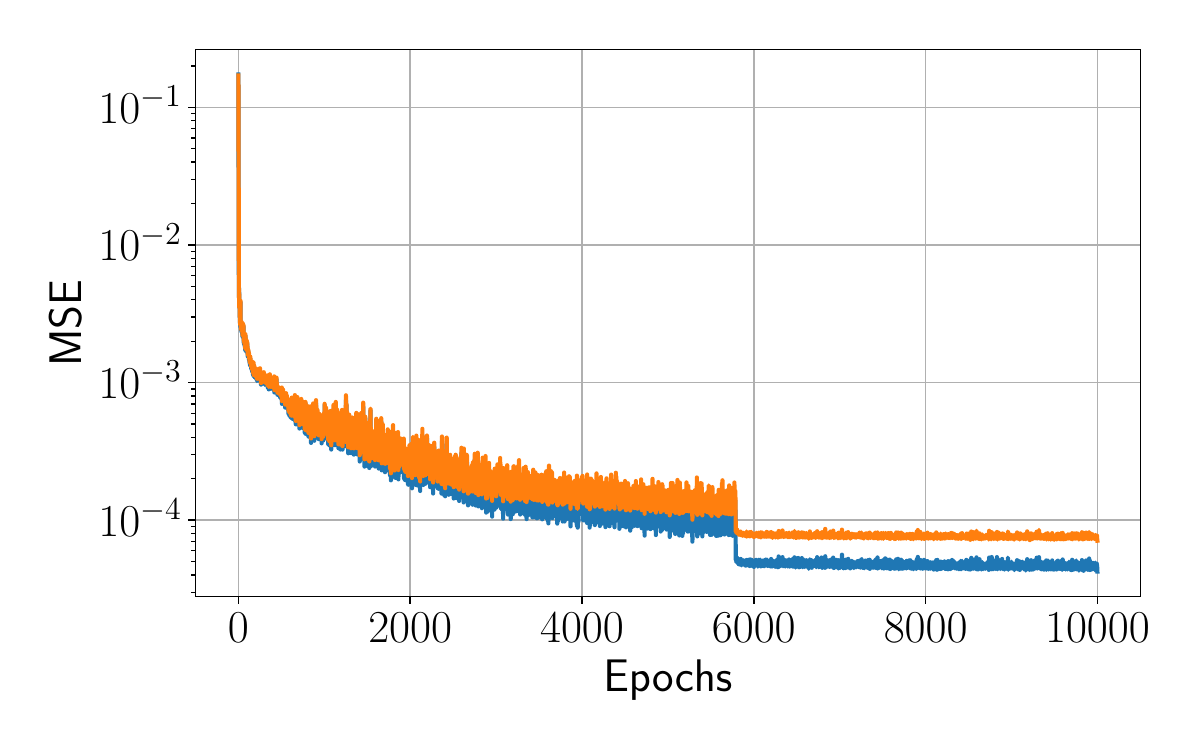}}
    \caption{Burgers' Equation: (a) ELBO loss and (b) MSE during DeepONet training using VI. \myline{c0}{}Training, \myline{c1}{}Validation}
    \label{fig:loss_burgers}
\end{figure}

\begin{figure}[!ht]
    \centering
    \subfigure[]{\includegraphics[width=0.49\linewidth]{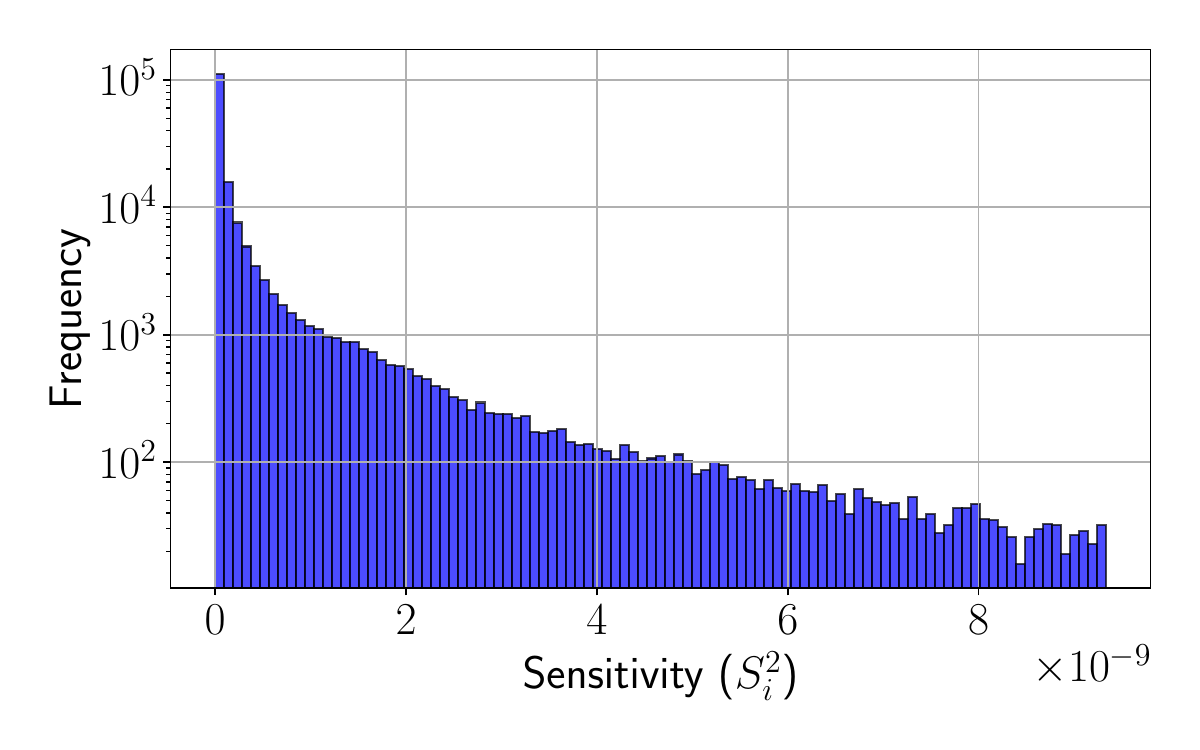}}
    \subfigure[]{\includegraphics[width=0.49\linewidth]{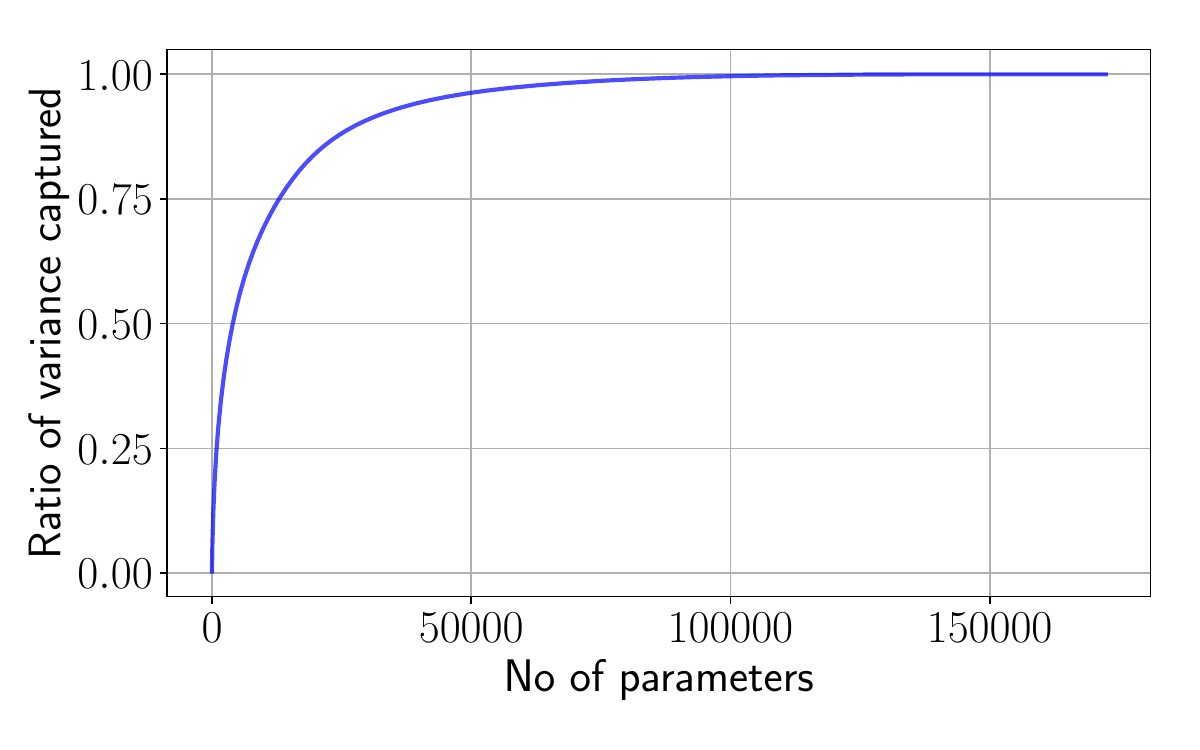}}
    \caption{Burgers' Equation: Bayesian DeepONet sensitivities. (a) Histogram of sensitivity values for the Bayesian DeepONet (x-axis is limited to the 99th percentile of $S_i^2$ for clarity). (b) Ratio of variance captured for increasing number of Bayesian DeepONet parameters, $\dfrac{\sum_{i=1}^{n}S_i^2}{\sum_{j=1}^{N_\Theta} S_j^2}$ vs $n$ .}
    \label{fig:senistivity_burger}
\end{figure}

The ELBO loss function and MSE during VI training are shown in Figure~\ref{fig:loss_burgers}. A learning rate scheduler is used to reduce the learning rate to 10\% of the current value when the validation loss plateaus. This results in the kinks observed near 6000 epochs in Figure~\ref{fig:loss_burgers}.  Following the VI training, parameter sensitivities are estimated as shown by the histogram in Figure~\ref{fig:senistivity_burger}(a). From Figure~\ref{fig:senistivity_burger}(b), we see that the number of parameters needed to capture 90\% of the variance is only 28,214 out of the 172,401 total parameters. This corresponds to a substantial dimension reduction, resulting in significant computational savings. Without this reduction, HMC is computationally prohibitive, and therefore, results from full HMC are not shown here. In addition, effect of varying the sensitivity threshold $\tau$ is shown in \ref{app:tau}  

\begin{figure}
    \centering
    \subfigure[]{
    \begin{minipage}[c]{0.49\linewidth}
        \includegraphics[width=0.9\linewidth]{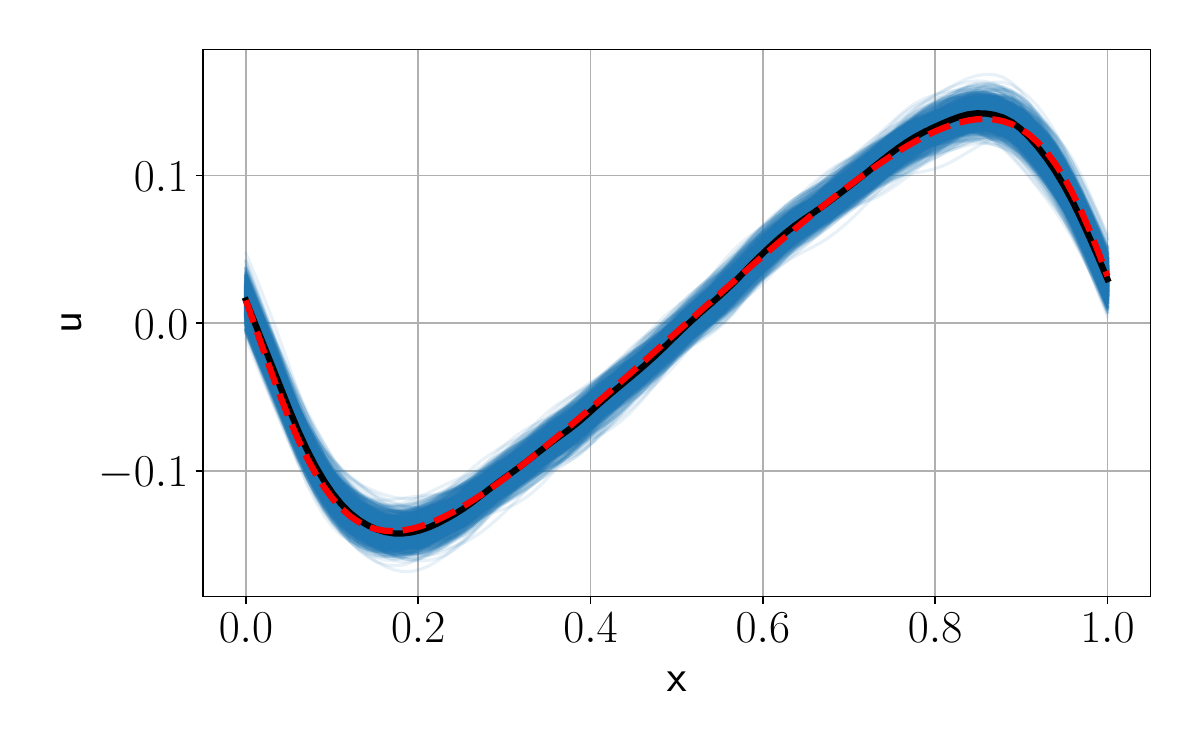}
    \end{minipage}
    \begin{minipage}[c]{0.49\linewidth}
        \includegraphics[width=0.9\linewidth]{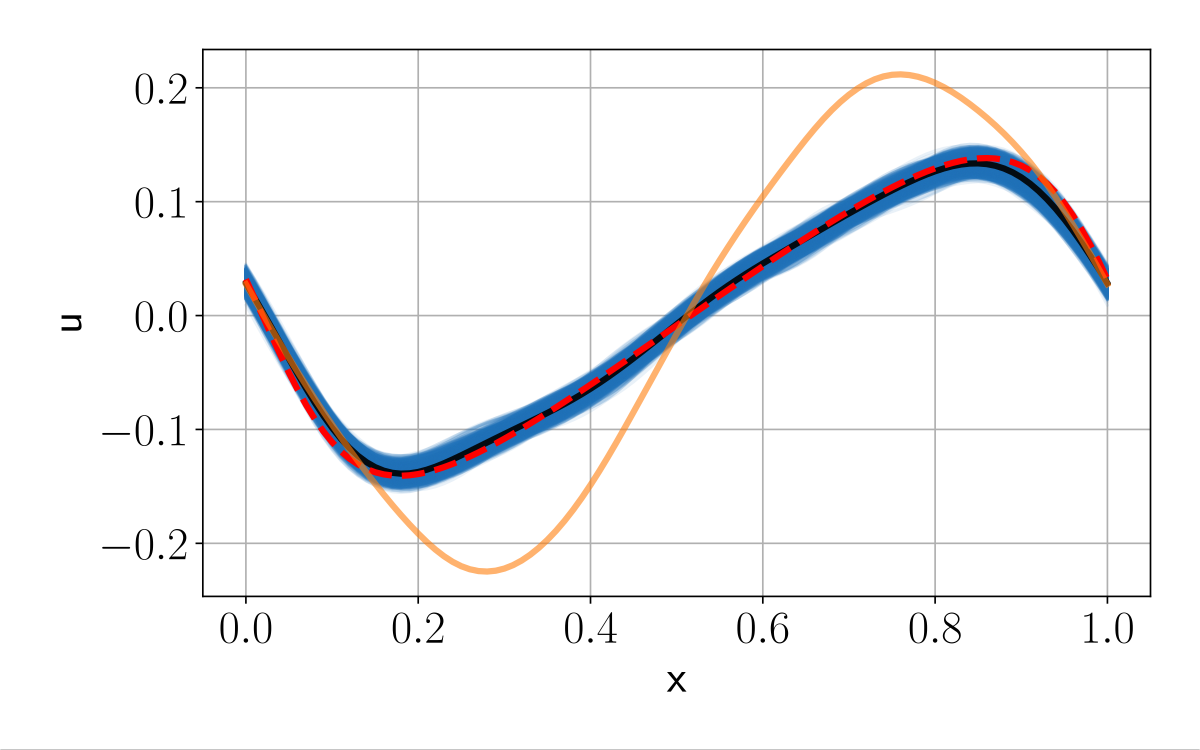}
    \end{minipage}
    }
    \subfigure[]{
    \begin{minipage}[c]{0.49\linewidth}
        \includegraphics[width=0.9\linewidth]{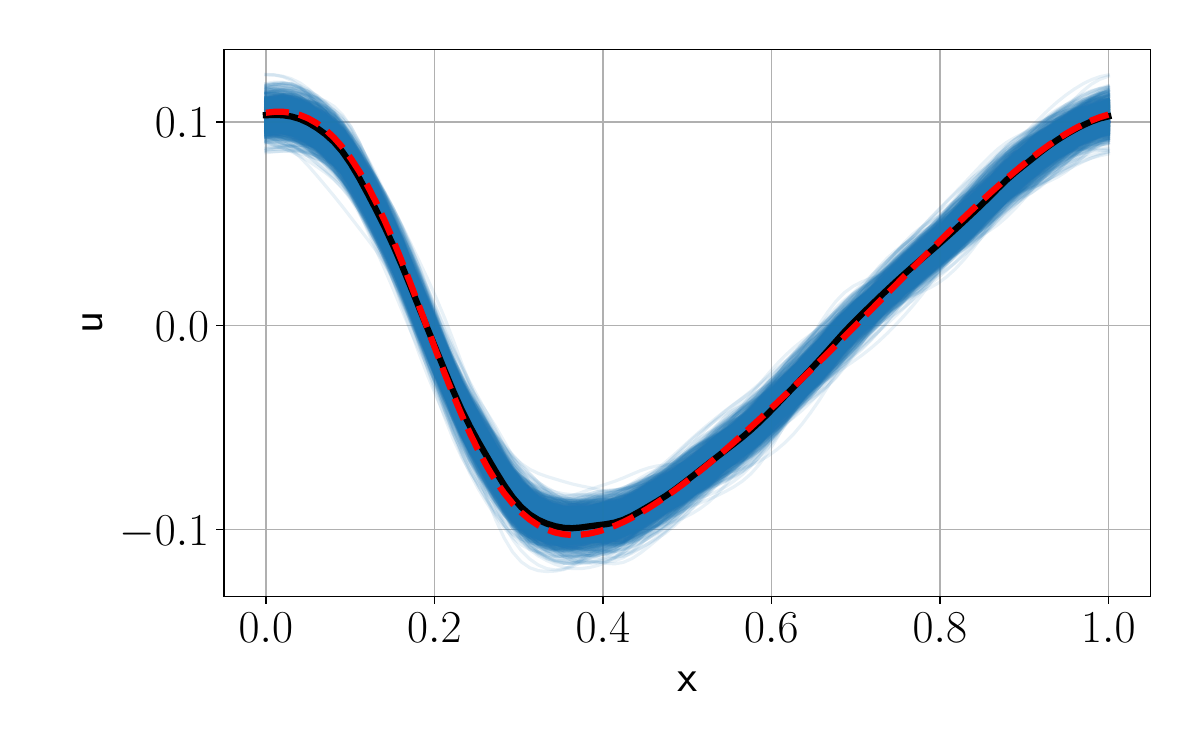}
    \end{minipage}
    \begin{minipage}[c]{0.49\linewidth}
        \includegraphics[width=0.9\linewidth]{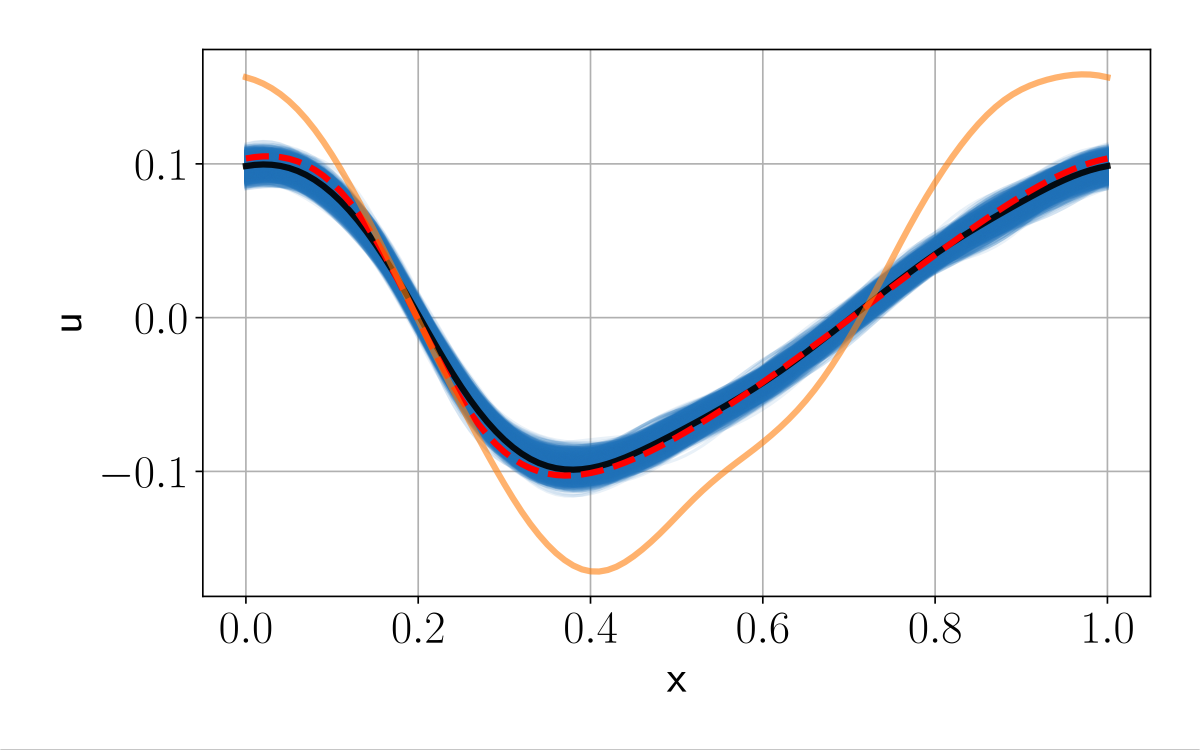}
    \end{minipage}
    }
    \subfigure[]{
    \begin{minipage}[c]{0.49\linewidth}
        \includegraphics[width=0.9\linewidth]{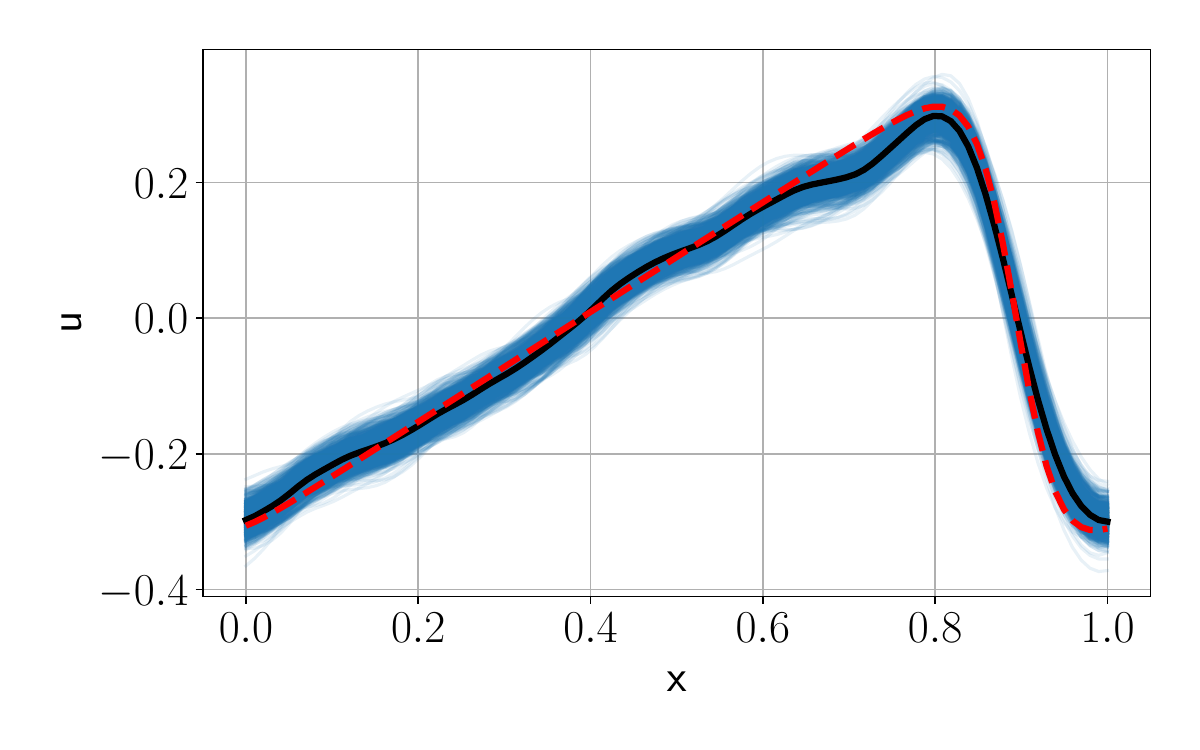}
    \end{minipage}
    \begin{minipage}[c]{0.49\linewidth}
        \includegraphics[width=0.9\linewidth]{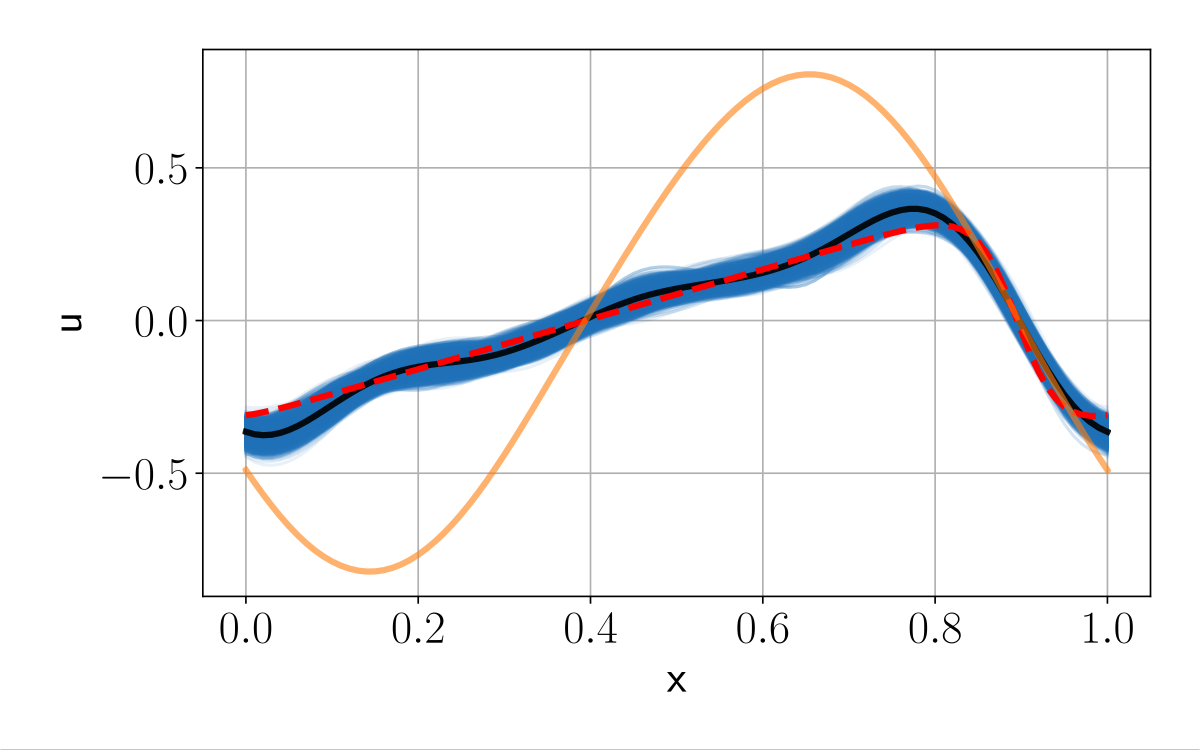}
    \end{minipage}
    }
    \subfigure[]{
    \begin{minipage}[c]{0.49\linewidth}
        \includegraphics[width=0.9\linewidth]{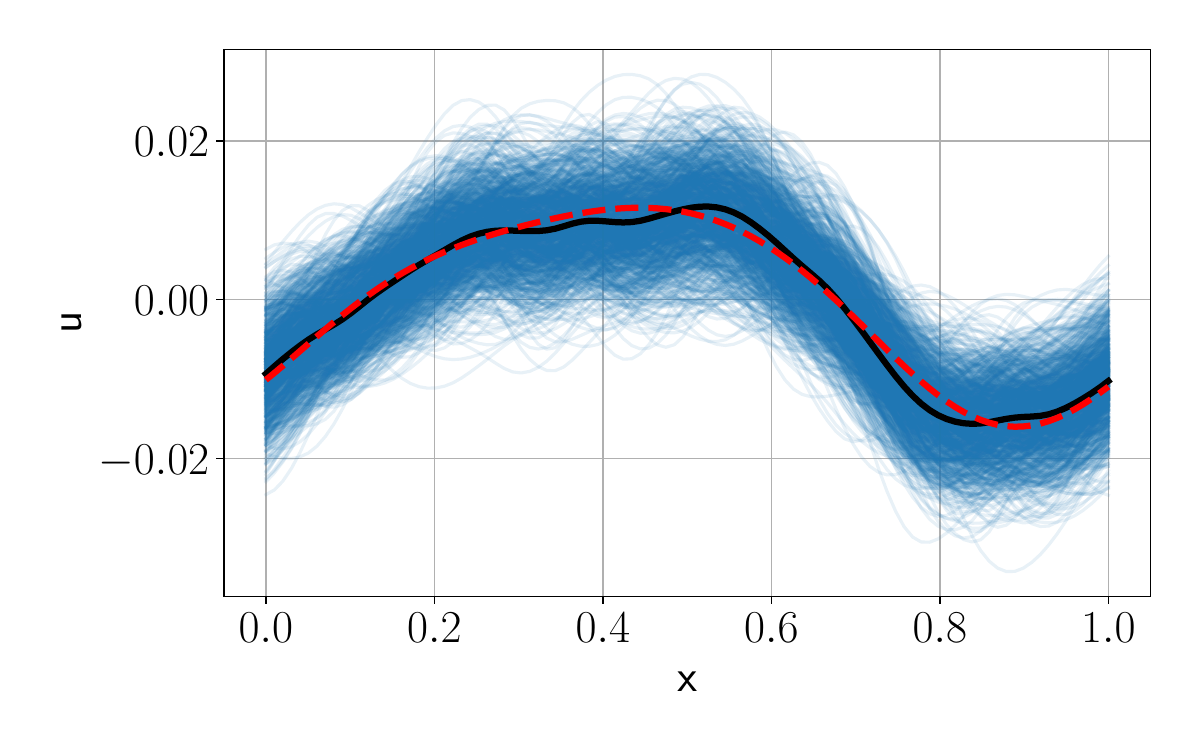}
    \end{minipage}
    \begin{minipage}[c]{0.49\linewidth}
        \includegraphics[width=0.9\linewidth]{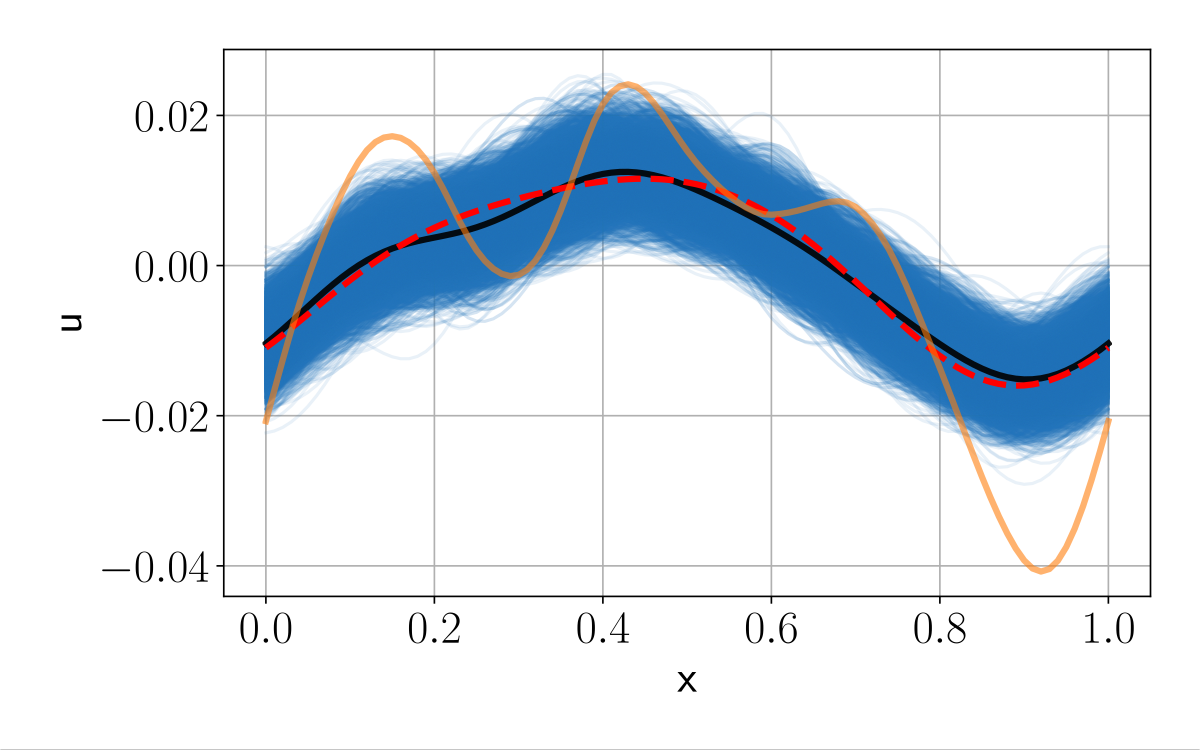}
    \end{minipage}
    }
    \caption{Comparing predictions between VI (left) and VI-HMC (right) for the Burgers equation for 4 different initial conditions at time $t = 1$. (a) Best prediction by VI, (b) Best prediction by VI-HMC, (c) Worst prediction by VI, and (d) Worst prediction by VI-HMC \myline{c0}{} samples, \myline{black}{}  mean prediction, \myline{red}{dashed} true solution, \myline{c1}{} initial condition.}
    \label{fig:prediction_unc_burgers}
\end{figure}

BNN predictions and the associated uncertainties are shown for the best and worst cases in Figure~\ref {fig:prediction_unc_burgers}. The mean predictions for the VI-HMC are reasonably accurate while showing well-correlated uncertainty estimates. The mean relative L2 errors, evaluated as $ \frac{1}{N} \sum \frac{||Y-Y_p||_2}{||Y||_2}$, for VI and VI-HMC are 3.8\% and 7.52\% respectively.

\subsubsection{Hypersonic flow over a cone} \label{sec:cone_example}
In the final example, a DeepONet is trained to learn the Navier-Stokes operator that maps the inlet conditions of a Mach 6, boundary layer flow over a cone to the wall pressure measurements. This problem is challenging due to the nonlinearity and chaotic nature of the Navier-Stokes operator.  Specifically, in the regime of interest, the flow is in the laminar-to-turbulence transitional regime, and as such, the downstream wall-pressure data are sensitive to small variations in the upstream disturbances \citep{park_zaki_2019,jahanbakhshi_zaki_2019,buchta_zaki_2021,buchta2022_cone,Wang_JFM2025}.  Previous attempts to learn the Navier-Stokes operator in the hypersonic transitional regime have only considered deterministic networks \citep{mao2021deepm,clarkdileoni2023_neural,hao2023instability}.  In light of the extreme sensitivity to uncertainty, a Bayesian approach is desirable. 
The data is obtained from Morra et al \cite{morra2024ml}. It is generated by solving the compressible Navier-Stokes equation by direct numerical simulations. The input to the DeepOnet is the inlet condition described by its power spectra represented as $\mathbf{c} \in \mathbb{R}^{11}$. The output is the pressure measurement $\mathrm{p}$ at given sensor locations and the frequency of the pressure spectra. 

\begin{figure}[!ht]
    \centering
    \subfigure[Loss]{\includegraphics[width=0.49\linewidth]{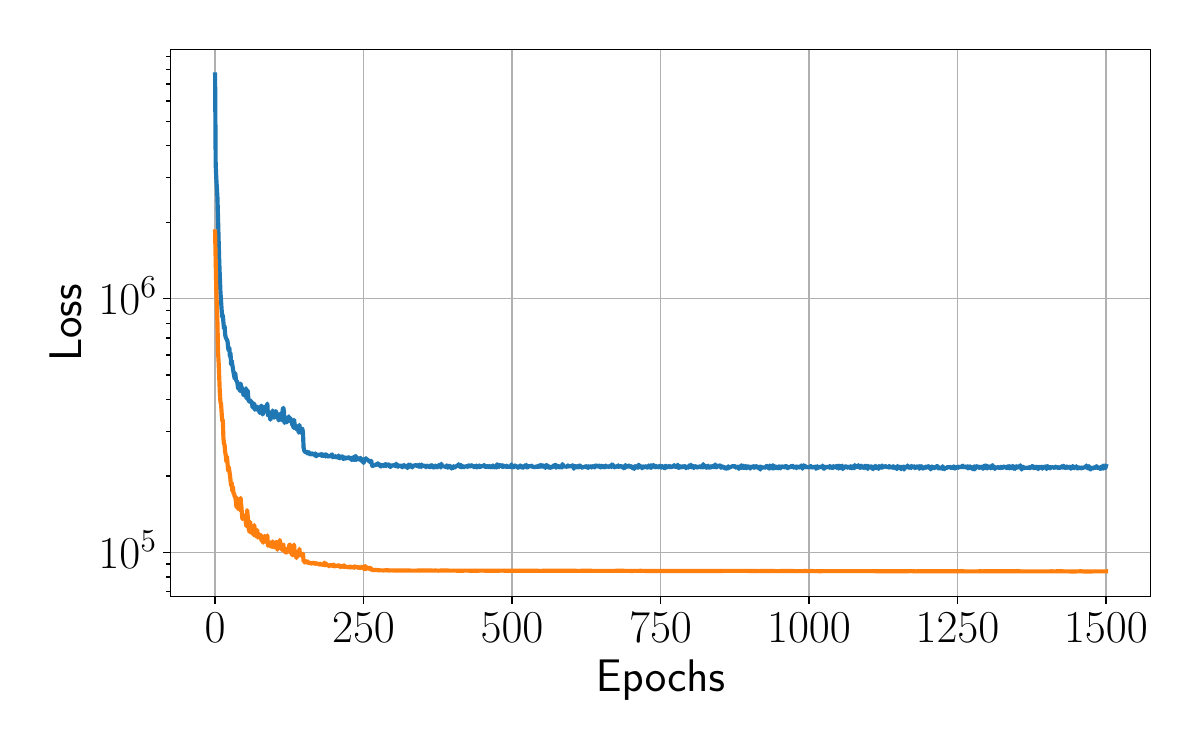}}
    \subfigure[MSE]{\includegraphics[width=0.49\linewidth]{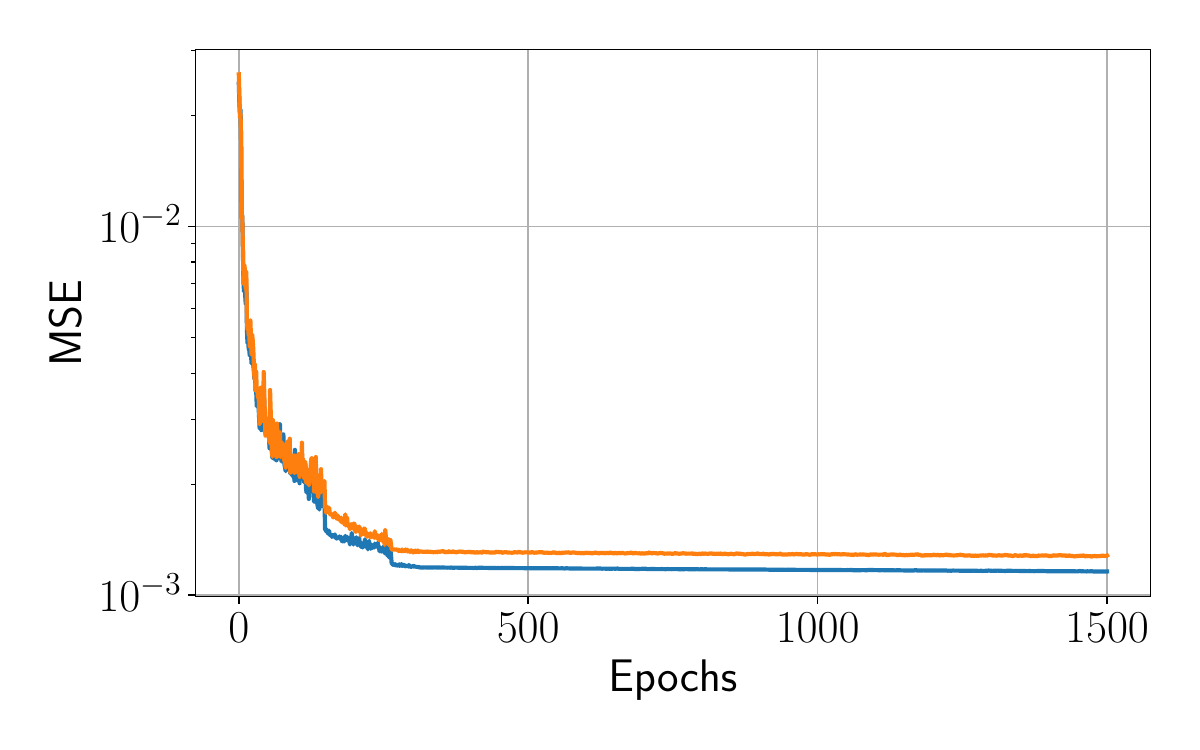}}
    \caption{ Flow over a cone, (a) ELBO loss and (b) MSE during DeepONet training using VI. \myline{c0}{}Training, \myline{c1}{}Validation}
    \label{fig:cone_loss}
\end{figure}

\begin{figure}[!ht]
    \centering
    \subfigure[]{\includegraphics[width=0.49\linewidth]{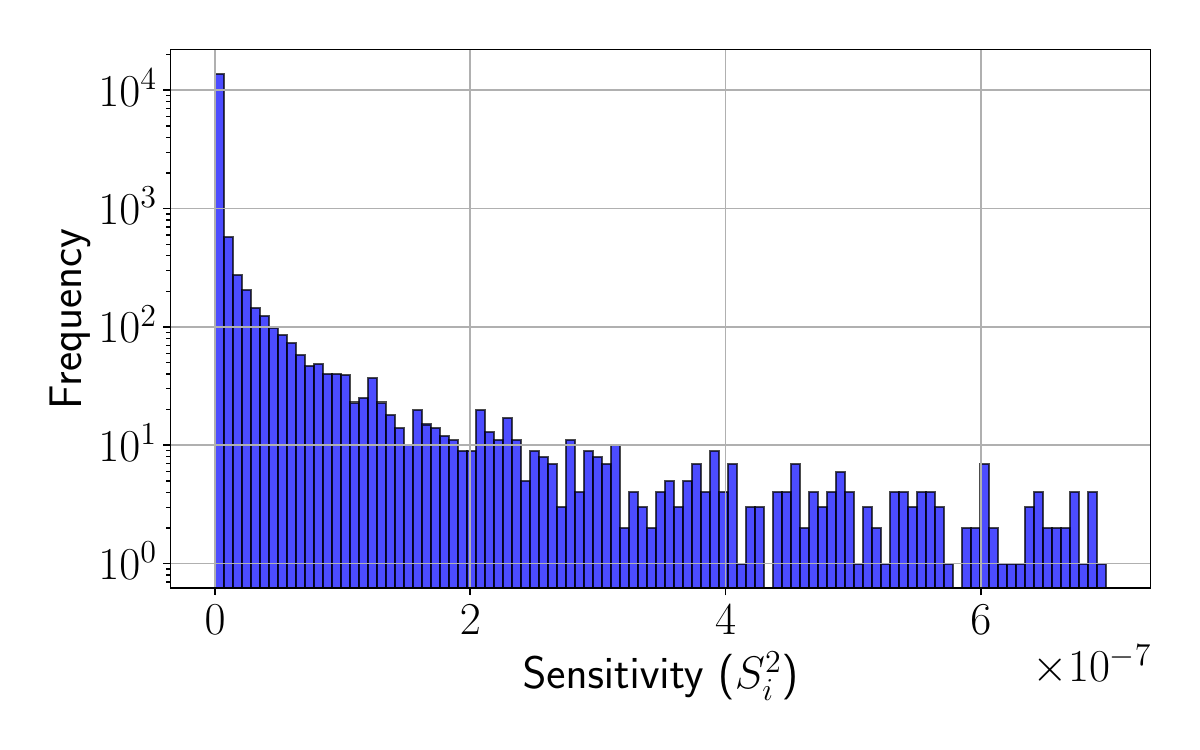}}
    \subfigure[]{\includegraphics[width=0.49\linewidth]{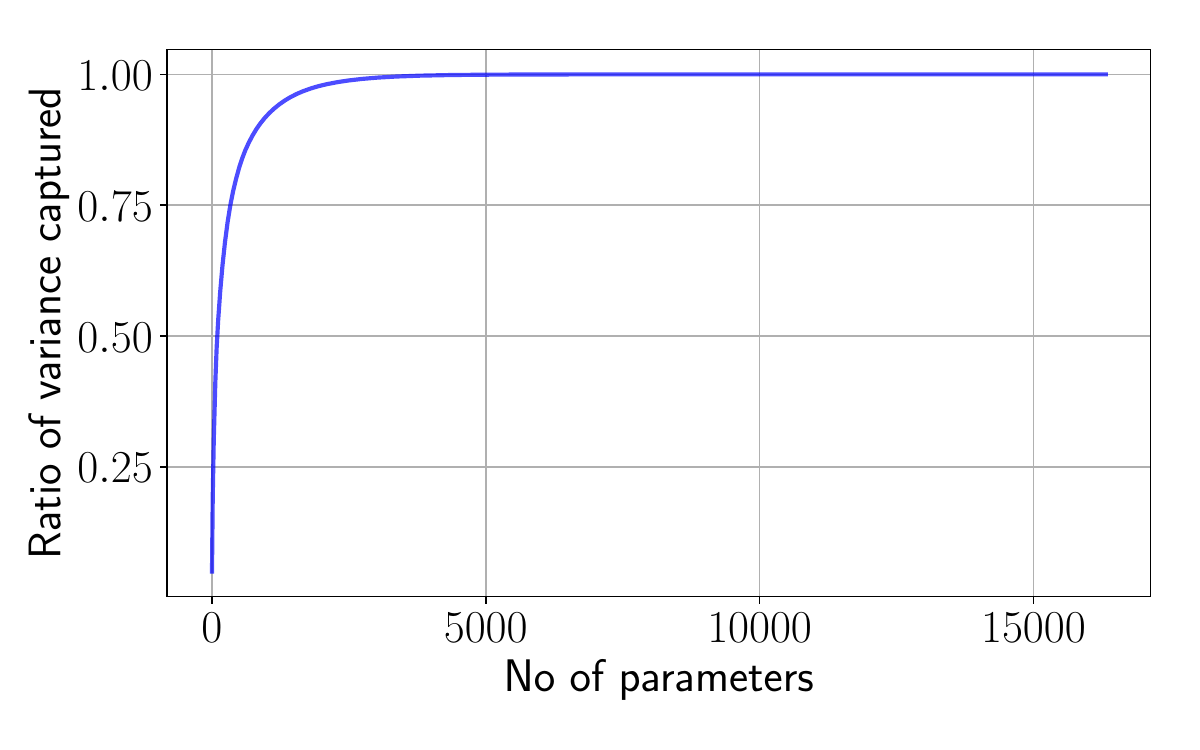}}
    \caption{ Flow over cone: Bayesian DeepONet sensitivities. (a) Histogram of sensitivity values for the Bayesian DeepONet (x-axis is limited to the 99th percentile of $S_i^2$ for clarity). (b) Ratio of variance captured for increasing number of Bayesian DeepONet parameters, $\dfrac{\sum_{i=1}^{n}S_i^2}{\sum_{j=1}^{N_\Theta} S_j^2}$ vs $n$.}
    \label{fig:cone_sensitivity}
\end{figure}

The architecture of the DeepOnet is presented in Table~\ref{Tab:hyperparams_deeponet}. The convergence of the ELBO loss for the VI training and the mean squared error loss during training is shown in Figure~\ref{fig:cone_loss}. Both losses show convergence after 300 epochs. The results of the sensitivity analysis are shown in Figure~\ref{fig:cone_sensitivity}. The total number of parameters in the DeepONet is 16,321. The sensitivity analysis shows that 840 parameters are enough to capture 90\% variance. Thus, there is more than a 19 times reduction in the dimension of the parameter space to perform HMC, which accelerates HMC significantly.  

\begin{figure}[!ht]
    \centering
     \subfigure[Sensor 1]{\includegraphics[width=0.32\linewidth]{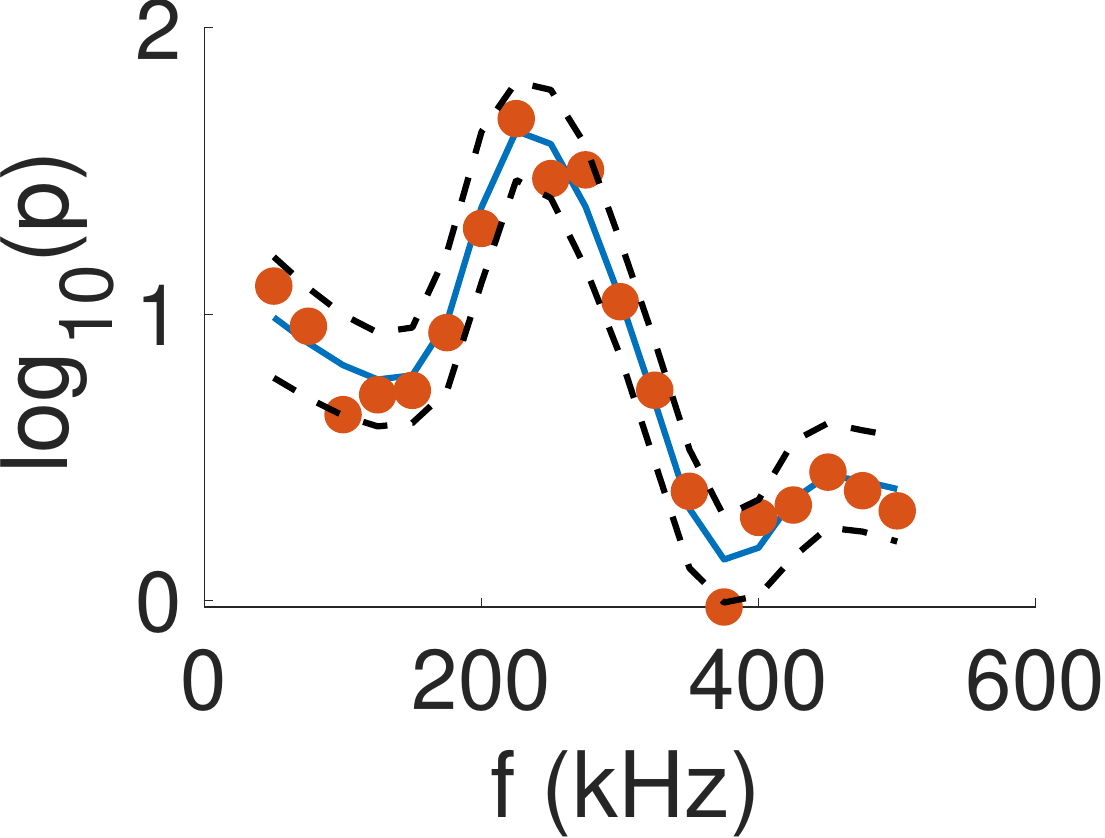}} \hfill
      \subfigure[Sensor 2]{\includegraphics[width=0.32\linewidth]{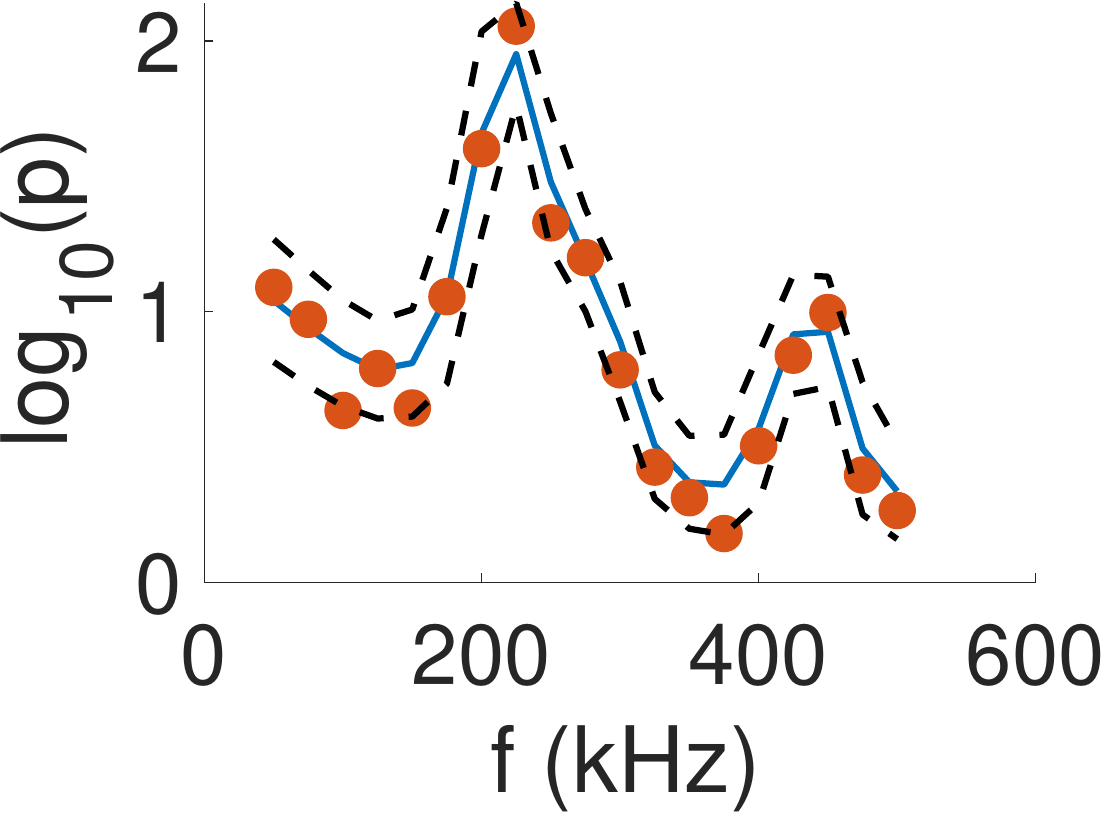}} \hfill
       \subfigure[Sensor 3]{\includegraphics[width=0.32\linewidth]{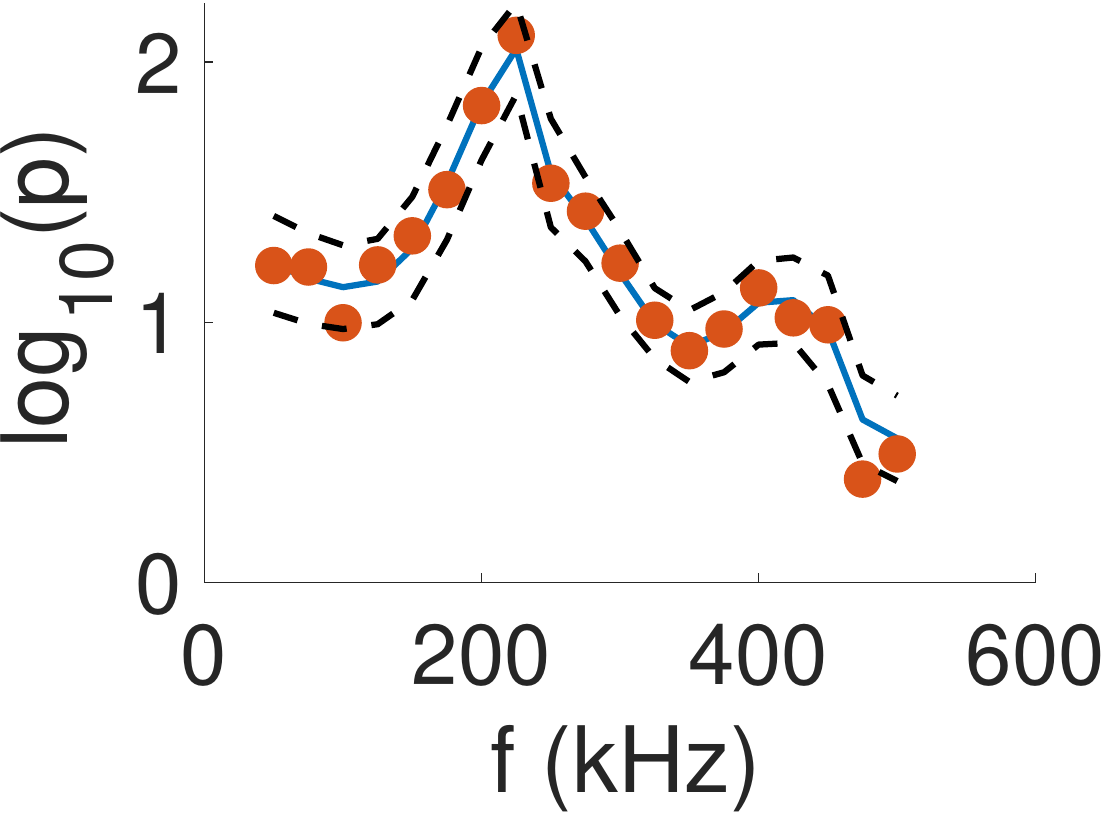}} 
        \subfigure[Sensor 4]{\includegraphics[width=0.32\linewidth]{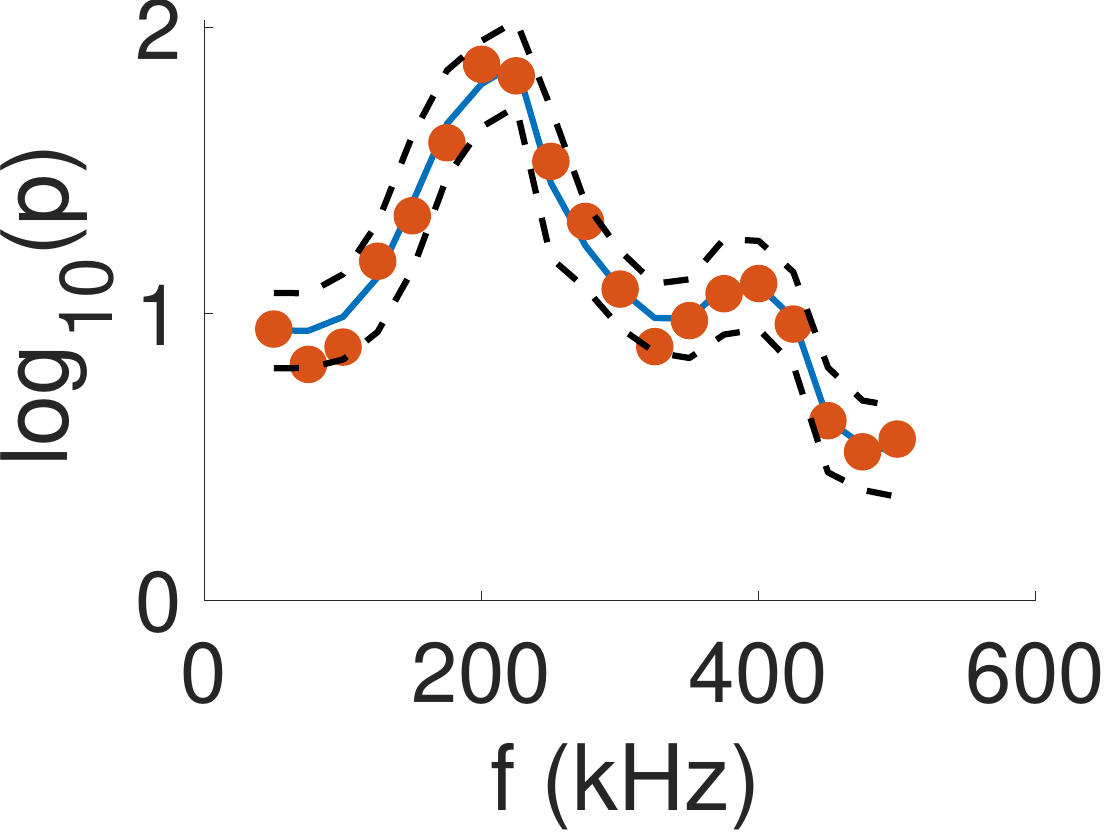}} \\
    \subfigure[Sensor 1]{\includegraphics[width=0.32\linewidth]{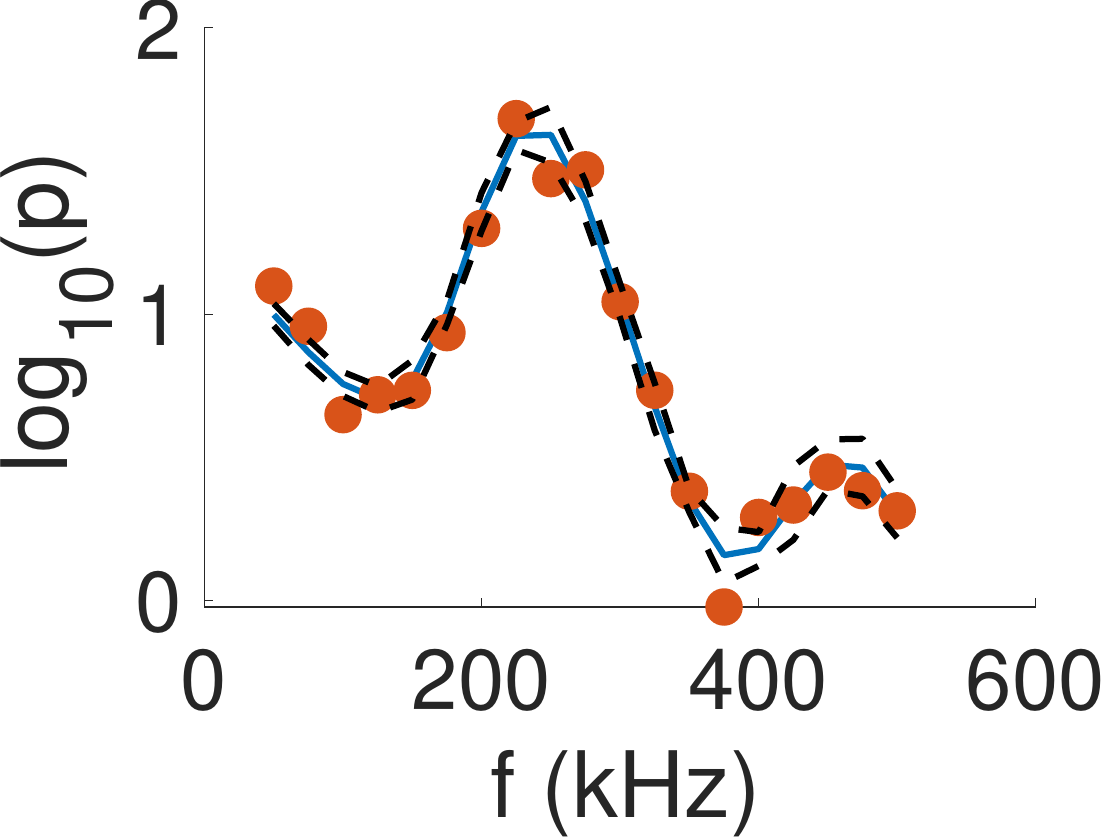}} \hfill
    \subfigure[Sensor 2]{\includegraphics[width=0.32\linewidth]{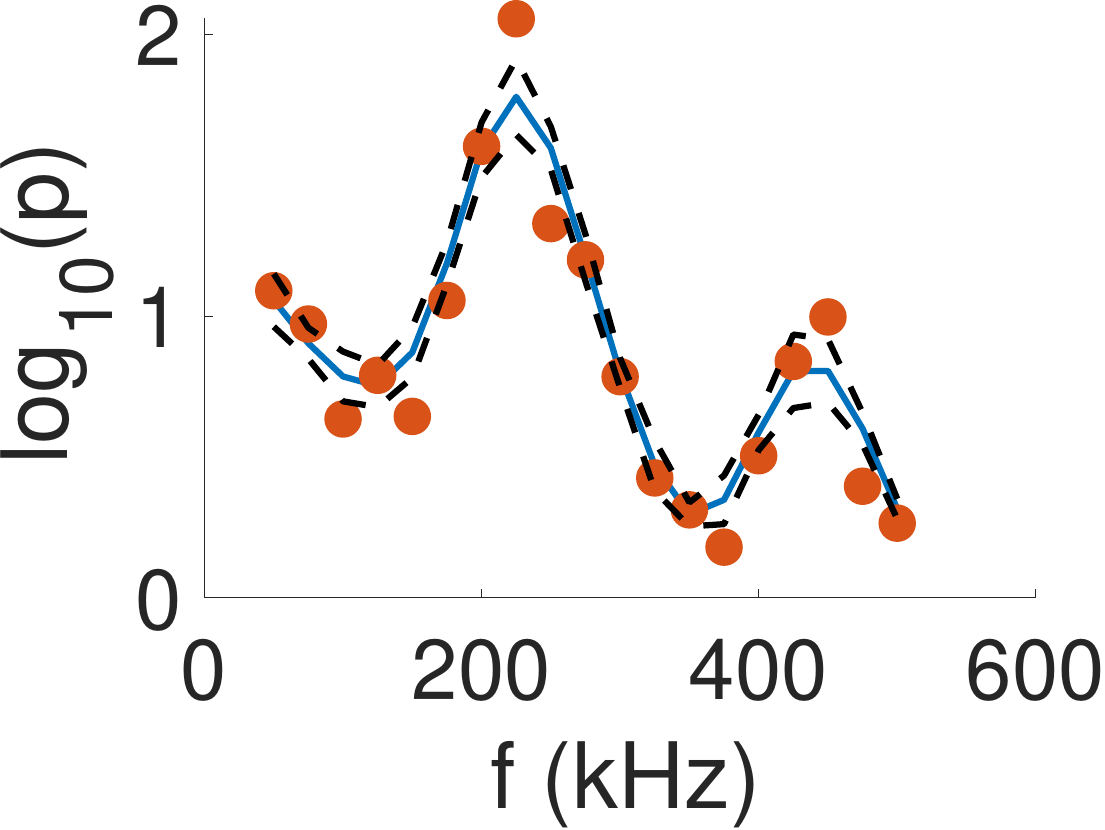}}\hfill
    \subfigure[Sensor 3]{\includegraphics[width=0.32\linewidth]{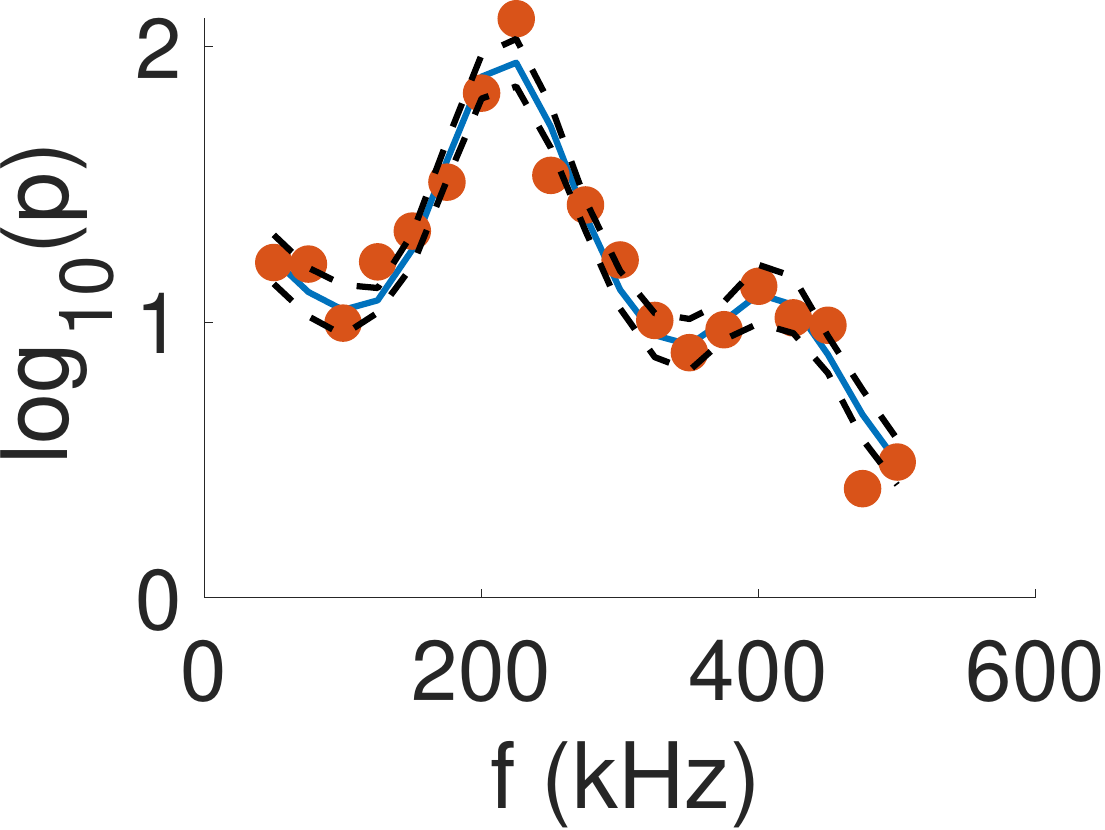}}
    \subfigure[Sensor 4]{\includegraphics[width=0.32\linewidth]{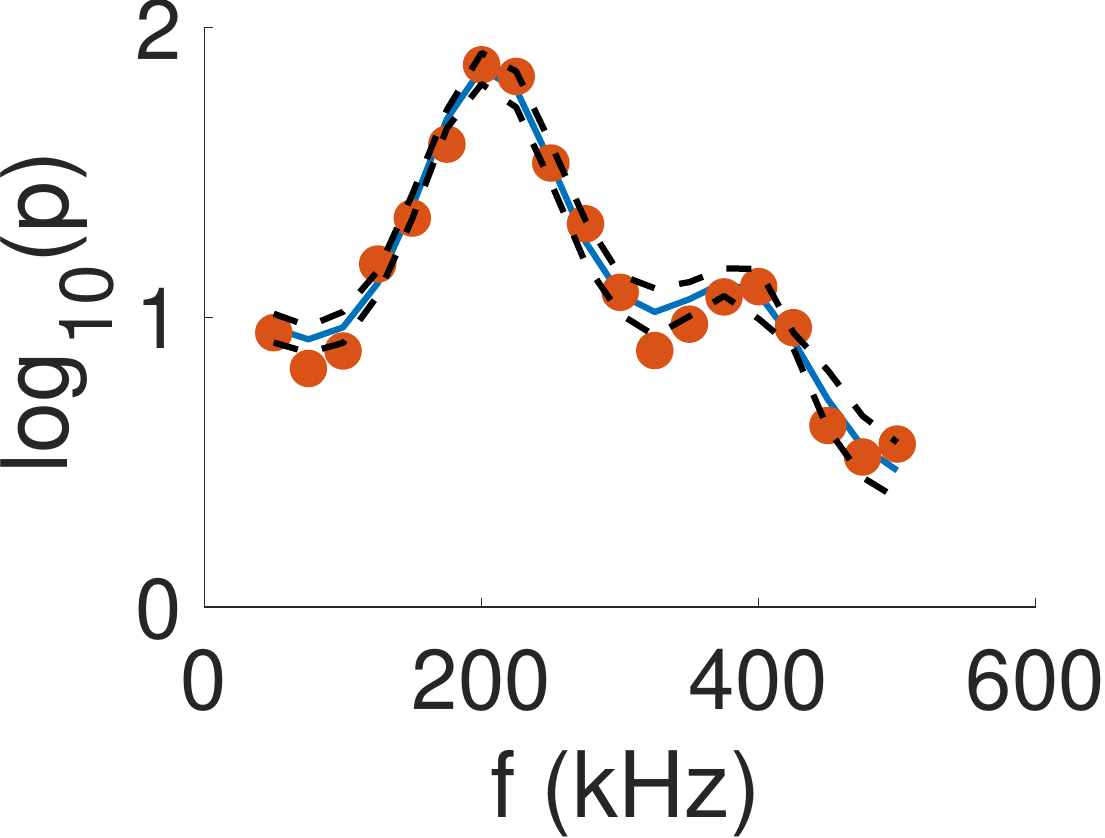}}
    \caption{ Flow over cone: Comparing uncertainties quantified by (a-d) VI and (e-h) VI-HMC. \mycircle{m1}  training data \myline{m0}{}DeepONet prediction, \myline{black}{dashed}uncertainty ($\pm 3\sigma$).}
    \label{fig:cone_uncertainty}
\end{figure}

VI-HMC is performed to sample the parameters in the reduced space, and a comparison of uncertainties is shown in Figure~\ref{fig:cone_uncertainty}. As we saw for the simpler BNNs, VI overestimates uncertainties when compared to the hybrid VI-HMC approach.

\subsubsection{Computational cost} \label{sec:cost_comparison}
A comparison of the computational gain achieved by the proposed VI-HMC method is elucidated in this section. Two different approaches are used to compare the computational benefits of the proposed method.

In the first approach, the integration of the Hamiltonian is performed using a fixed step size for a given problem when implementing the VI-HMC and the HMC methods. With the same step size, the acceptance rate, accuracy, and the time taken to draw one sample using the two approaches are compared in Table~\ref{Tab:cost_deeponet}. The proposed VI-HMC method has a better acceptance rate while taking significantly less time compared to the HMC approach. Specifically for the Burgers equation, there is more than $2$x reduction in computational cost for the proposed VI-HMC approach compared to HMC, which can be attributed to the dimension reduction of the parameter space. In addition, the acceptance rate of the HMC approach is merely 10\%, rendering the sampling inefficient in contrast to the 87\% acceptance for VI-HMC. Similarly, for the cone problem, there is a $2.5$x reduction in computational cost for the VI-HMC method. Similar to the Burgers problem, the acceptance rate for HMC for the cone problem is just 55\%, which implies poor sampling quality, compared to the 91\% acceptance rate for the VI-HMC approach. Although theoretically, no time improvement per sample is expected as mentioned in Section~\ref{sec:computational_benefits}, practical implementation of the full batch HMC results in gradient explosion, i.e, infinite values for the gradients of the parameters. Therefore, a symmetric splitting HMC is implemented following \cite{cobb2021scaling}, which results in the increased computational cost for the two problems implemented in this work. 

In the second approach, the acceptance rate for MCMC sampling is fixed at 80\% and the step size is computed using the step size adaptation method proposed in Algorithm 5 of \cite{hoffman2014no}. For the Burgers equation, the required step size for an 80\% acceptance rate with HMC is 5 times lower than the step size required for VI-HMC, which further demonstrates the computational advantage of the proposed approach. In the cone problem, the required acceptance step size for HMC is six orders of magnitude smaller than the step size required for VI-HMC, rendering the HMC approach infeasible to implement within an acceptable time frame. 

The overall computational benefits of the proposed approach originate from two sources: reduced error accumulation in leapfrog steps and reduced gradient pathologies. The primary source is the reduced error accumulation in the leapfrog integration, resulting from the dimensionality reduction of the parameter space. This reduced error accumulation results in better acceptance rates when the step size is fixed and allows for larger step sizes when the acceptance rate is fixed, yielding computational savings. Additionally, the larger parameter size, combined with large step sizes in leapfrog integration, leads to gradient pathologies, which adds to the computational cost in HMC. Whereas our approach reduces such gradient pathologies, resulting in additional computational benefits in some cases. The breakdown of these benefits can be seen from Table~\ref{Tab:cost_deeponet}. Improvements to the required step size at a fixed acceptance rate are a result of reduced error accumulation in the leapfrog integration. Whereas, a change in time/sample, if any, is a result of reduced gradient pathologies.
\begin{table}[htpb!]
    \centering
    \caption{Comparison of the computation cost between HMC and VI-HMC for Bayesian Neural Operators. The step size is fixed at $2E-4$ and $1E-5$ for the Burgers and the cone problems, respectively (for column 3), and the acceptance rate is fixed at 80\% for both problems (for column 4). }
    \begin{tabular}{cccccc}
    \toprule
     \multirow{2}{*}{\textbf{Problem}}& \multirow{2}{*}{\textbf{Method}} & \multicolumn{3}{c}{\textbf{Fixed step size}}  &  {\textbf{Fixed acceptance rate}}  \\
      && Acceptance rate & MSE & Time/sample & Required step size  \\ \hline
      \multirow{2}{*}{\textbf{Burgers}}& HMC & 10\% & -\footnotemark{} & 13.9 s \footnotemark & 1.0E-4 \\
      &VI-HMC & 87\%& 2.9E-4 & 6.27 s& 5.0E-4 \\ \hline
     \multirow{2}{*}{\textbf{Cone}}& HMC &55\% & 3.7E-3 & 0.63 s \footnotemark[\value{footnote}] &4.2E-11 \\
      &VI-HMC & 91\% & 2.7E-3 &0.25 s& 1.6E-5 \\
        \bottomrule
    \end{tabular}
    \label{Tab:cost_deeponet}
\end{table}
\footnotetext[1]{The low acceptance rate causes the samples to cluster near the initial point, making MSE uninformative.}
\footnotetext[2]{HMC is performed using a split Hamiltonian formulation, yielding a higher time per sample.}

\section{Limitations} \label{sec:limitations}
Since the proposed method uses the VI technique in its initial training, many of the limitations in training a BNN with VI are carried over to the proposed method. This is especially noteworthy in cases where training VI and sampling from HMC require contradictory problem setups. For example, we found that for Case 2 of the BNN example, where the data is limited and the network is large, it is faster and more accurate to train a neural network with VI by assuming a small standard deviation for the Gaussian likelihood (noise in the data). In contrast, HMC struggles to converge to a posterior distribution that yields good predictions under such a small likelihood standard deviation. On the other hand, when the likelihood standard deviation is large, HMC  converges faster while VI struggles to identify a distribution that accurately predicts the given data. Since changing the likelihood or introducing posterior tempering alters the underlying distribution that is approximated or sampled, one must take utmost care in optimizing the hyper-parameters, such as the likelihood variance or the posterior temperature, when implementing the proposed VI-HMC method.

\section{Conclusions} \label{sec:conclusions}
We propose a new hybrid approach that combines variational inference and Hamiltonian Monte Carlo for sampling from the posterior parameter distributions in Bayesian neural networks and neural operators. We demonstrate that different network parameters contribute to varying degrees in quantifying uncertainties in predictions and leverage this insight to formulate the hybrid approach. This approach uses sensitivity analysis on the posterior distributions learned from variational inference to significantly reduce the dimension of the parameter space. Hamiltonian Monte Carlo is then applied to this reduced parameter space, which greatly accelerates uncertainty quantification. The proposed approach provides accurate estimates of uncertainties while converging much faster than traditional HMC methods. We demonstrate the efficiency and accuracy of the framework by learning the posterior parameter distributions in both deep neural networks and operator networks and demonstrating their robust uncertainty estimates in the solution.  

We demonstrated the application of the method to learning functions and operators.  In the latter case, we demonstrated that we can accurately and efficiently predict the uncertainty of operator networks that represent the Burgers and Navier-Stokes equations. The present approach provides an important foundation for a number of future avenues, related to both forward and inverse problems.  In the former context, the trained operator networks can be evaluated at very high speeds and are therefore ideal for integration in control applications.  In the latter context, specifically in data assimilation \citep{Zaki2025}, the operator networks can accelerate the solution of the inverse problem by significantly reducing the cost of the forward model. This higher cost in the forward model stems from solving the nonlinear differential equation, and this cost is reduced by inexpensive sampling of the network parameters and evaluating the outputs. 

\section*{Acknowledgments}
This material is based upon work supported by the U.S. Department of Energy, Office of Science, Office of Advanced Scientific Computing Research, under Award Number DE-SC0024162.

\newpage

\appendix

\section{DeepONet Architectures}
\label{sec:App_Architecture}

\begin{table}[h!]
    \centering
    \caption{DeepOnet architecture for problems considered. Both trunk and branch networks are feed-forward neural networks.}
    \begin{tabular}{cccccccc}
    \toprule
      \multirow{2}{*}{\textbf{Problem}} & \multicolumn{3}{c}{\textbf{Branch}}  & \multicolumn{3}{c}{\textbf{Trunk}} & \multirow{2}{*}{\textbf{Activation}} \\
      & Input& Width & Depth & Input& Width & Depth & \\ \hline
      Burgers & 101& 100&9 & 5&100&9&tanh\\
      Cone &11& 44&6&2&44&5&tanh \\
        \bottomrule
    \end{tabular}
    \label{Tab:hyperparams_deeponet}
\end{table}

\section{Sensitivity in the parameters}
\begin{figure}[htpb!]
    \centering
    \includegraphics[width=\linewidth]{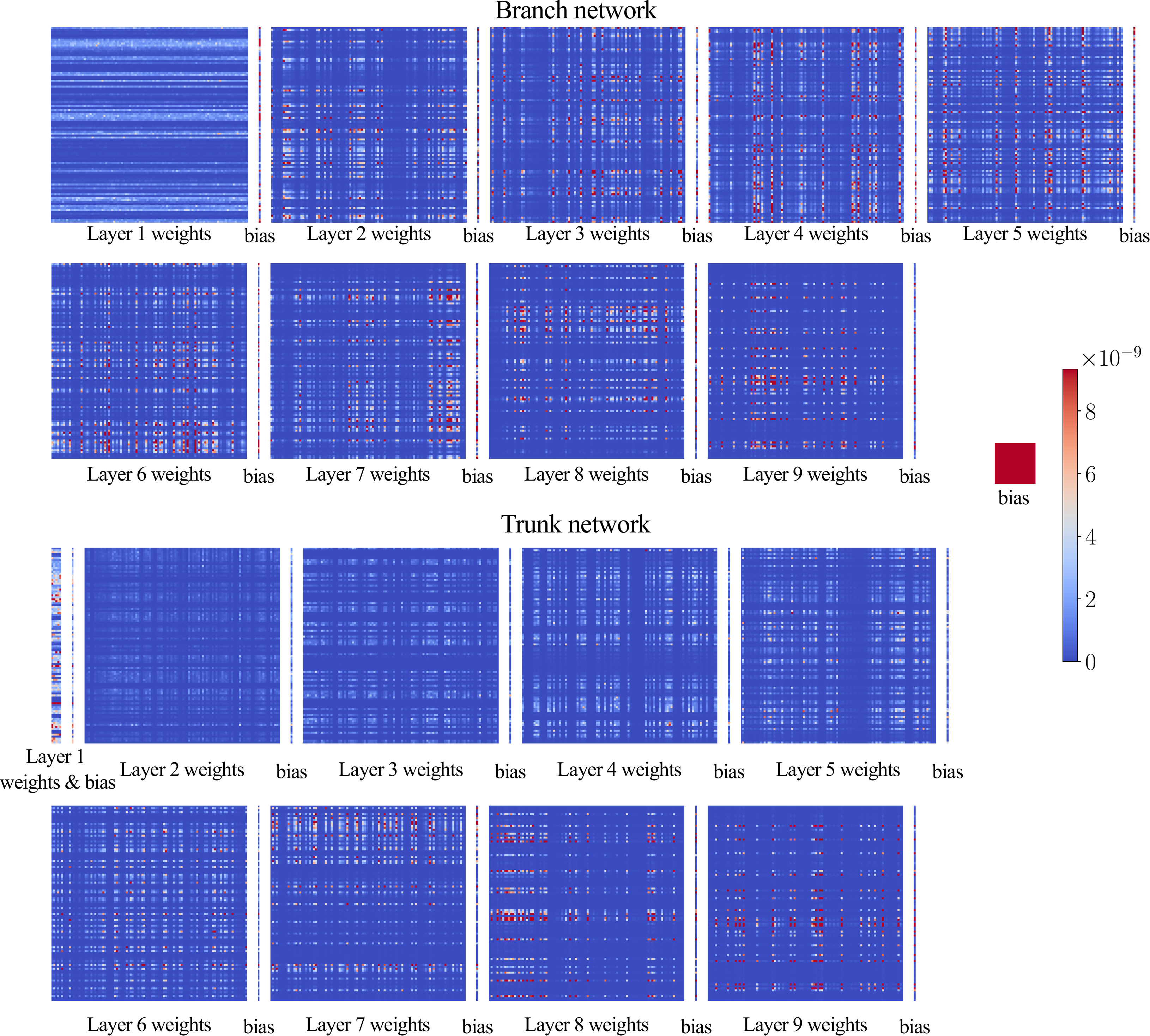}
    \caption{Sensitivity in the DeepONet parameters for the Burgers problem. Each $n \times m$ matrix corresponds to the sensitivity values of the parameters in a layer with $m$ inputs and $n$ outputs. Sensitivities of both weights and biases of a layer are shown. }
    \label{fig:burgers_sens}
\end{figure}

\begin{figure}
    \centering
    \includegraphics[width=\linewidth]{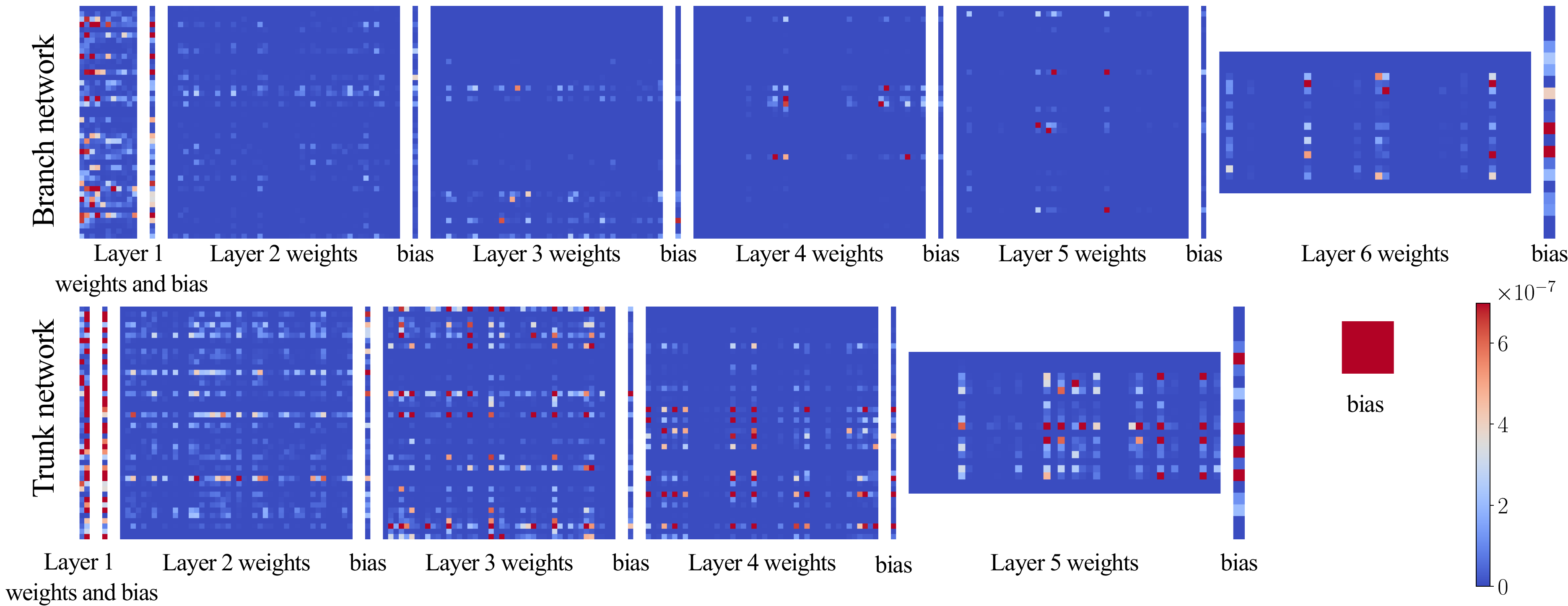}
    \caption{Sensitivity in the DeepONet parameters for the cone problem. Each $n \times m$ matrix corresponds to the sensitivity values of the parameters in a layer with $m$ inputs and $n$ outputs. Sensitivities of both weights and biases of a layer are shown.}
    \label{fig:cone_sens}
\end{figure}

Parameters Sensitivities for the Bayesian DeepONets are visualized in Figure~\ref{fig:burgers_sens},\ref{fig:cone_sens}. In the Burgers example, the trunk net is less sensitive in predicting uncertainties than the branch net. The trunk net can be viewed as approximating the basis of the output function, and the branch net as approximating the coefficients of the output function. Parameters in the branch net being more sensitive to the outputs implies that uncertainties in the output are more sensitive to uncertainties in the coefficients than the basis functions. Further, in the trunk net of the burgers problem, the later layers (layers close to the outputs) are more sensitive than the initial layers (layers close to the inputs), which might be attributed to vanishing gradients in the initial layers. 

In the cone problem, however, the trunk network has more sensitive parameters than the branch net, indicating that the basis functions are more sensitive in this case. Further analysis is necessary to understand the spread of sensitivities across the network weights.

\section{Hyper-parameters of HMC and VI-HMC} \label{app:hyperparams}
For the HMC approach, the step size and the number of leapfrog steps are important hyper-parameters that affect the sample quality. In all of the examples presented in this work, all hyper-parameters except for the step size for HMC and VI-HMC approaches are set the same values as shown in Table~\ref{Tab:hyperparams_hmc}.  The step size is manually tuned, and the number of steps is set to be  $\frac{\pi \times \text{posterior variance}}{2 \times \text{step size}}$. The posterior variance is taken as the maximum variance of parameters obtained from VI for the Burgers problem and the BNN's case I, and the median variance of parameters from VI for all other cases. 

10 parallel chains are randomly initialized for both HMC and the VI-HMC for all the problems considered in this work. For the Burgers problem, since the number of parameters is large even after reduction in the case of VI-HMC, 10 random initializations are done around the mean parameters obtained from VI for faster convergence. Therefore, the number of samples for Burgers is limited to 1000 per chain, in contrast to longer chains for the other examples.  

For all the examples, the prior and the likelihood are assumed to be zero-mean Gaussian and a Gaussian, respectively, with variances given in the Table~\ref{Tab:hyperparams_hmc}. Both the prior and likelihood are assumed to be the same for VI, HMC, and VI-HMC approaches to ensure sampling from the same posterior distribution of parameters.

\begin{table}[h!]
    \centering
    \caption{Hyper-parameters for the HMC and VI-HMC approaches.}
    \begin{tabular}{lcccc}
    \toprule
      \multirow{2}{*}{\textbf{Parameter}} & \multicolumn{2}{c}{\textbf{BNN}} & \multirow{2}{*}{\textbf{Burgers}} & \multirow{2}{*}{\textbf{Cone}}\\
      &Case I &Case II\\ \hline
        Step size (VI-HMC)&1E-5 &1E-3 &2E-4 &1E-5 \\ 
        Step Size (HMC)&1E-5 &1E-4 &- &-\\
        Posterior variance& $0.0679^2$&$0.2024^2$ &$0.1027^2$ &$0.0108^2$\\
        Parallel chains\footnotemark &1 &10 &10 &10\\
        Samples per chain&5,000&10,000 &1000 &100,000\\
        Burn in\footnotemark{} & 4,000&9,000 &100 &90,000\\
        Prior variance&1.0&1.0 &$0.1^2$ &$0.01^2$ \\
        Likelihood variance &(1E-3$)^2$  &(5E-2$)^2$ &1.0 &(5E-3$)^2$ \\
        
        \bottomrule
    \end{tabular}
    \label{Tab:hyperparams_hmc}
\end{table}
\footnotetext[3]{A few bad chains are removed before post-processing, where applicable.}
\footnotetext{The number of initial samples left out while post-processing and plotting results.}

\section{Robustness of sensitivity scores} \label{app:robustness}
The sensitivity scores are robust when the mean and standard deviation of the BNN parameters are unchanged for different training scenarios. When the choice of the dataset or the noise level affects the distribution of parameters learned by the network, the sensitivity scores are altered. This is expected since the sensitivity scores (in Eq.\ref{eqn:sensitivities}) rely on the gradient of predictions with respect to parameters and the standard deviation of parameters. Any change in these gradients or standard deviation results in different sensitivities of the parameters, resulting in different contributions from the individual parameters to the total sensitivities. 
To verify this, experiments with different datasets and noise levels are performed for the neural network case II, and the results of the sensitivity analysis are presented.
        
\begin{figure}
    \centering
    \subfigure[Varying noise level]{\includegraphics[width=0.9\linewidth]{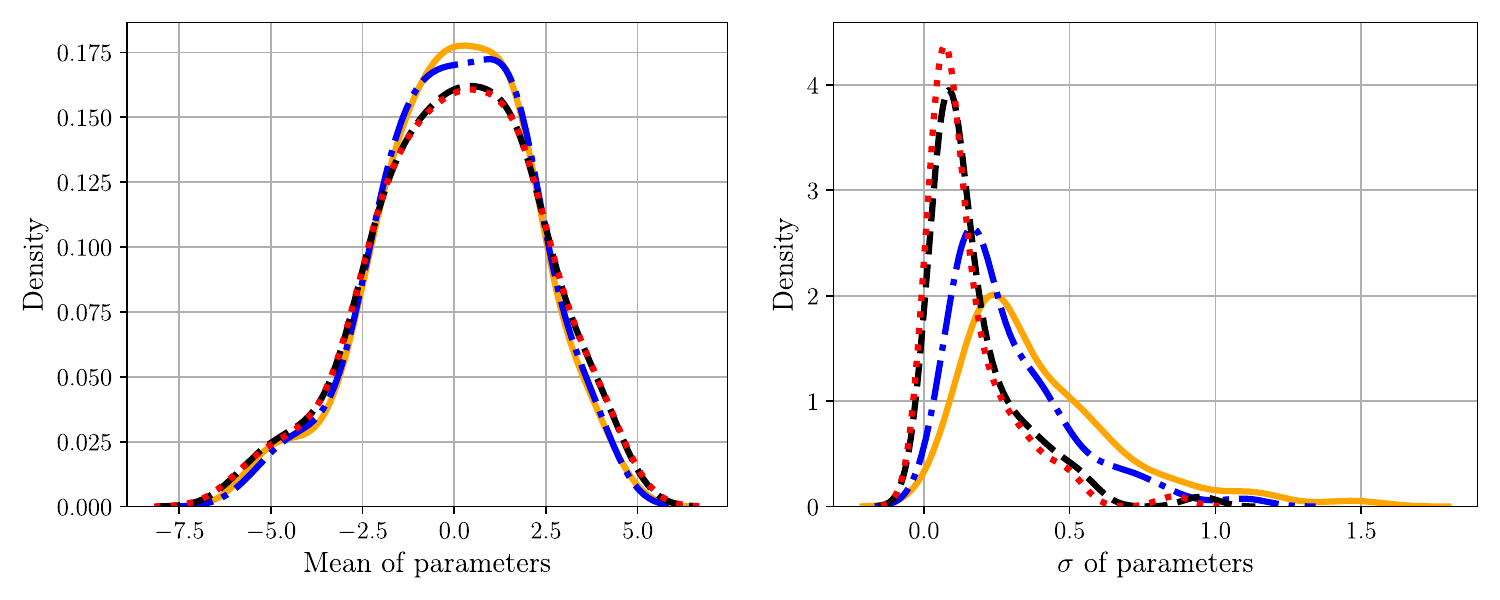}}
    \subfigure[Varying training data]{\includegraphics[width=0.9\linewidth]{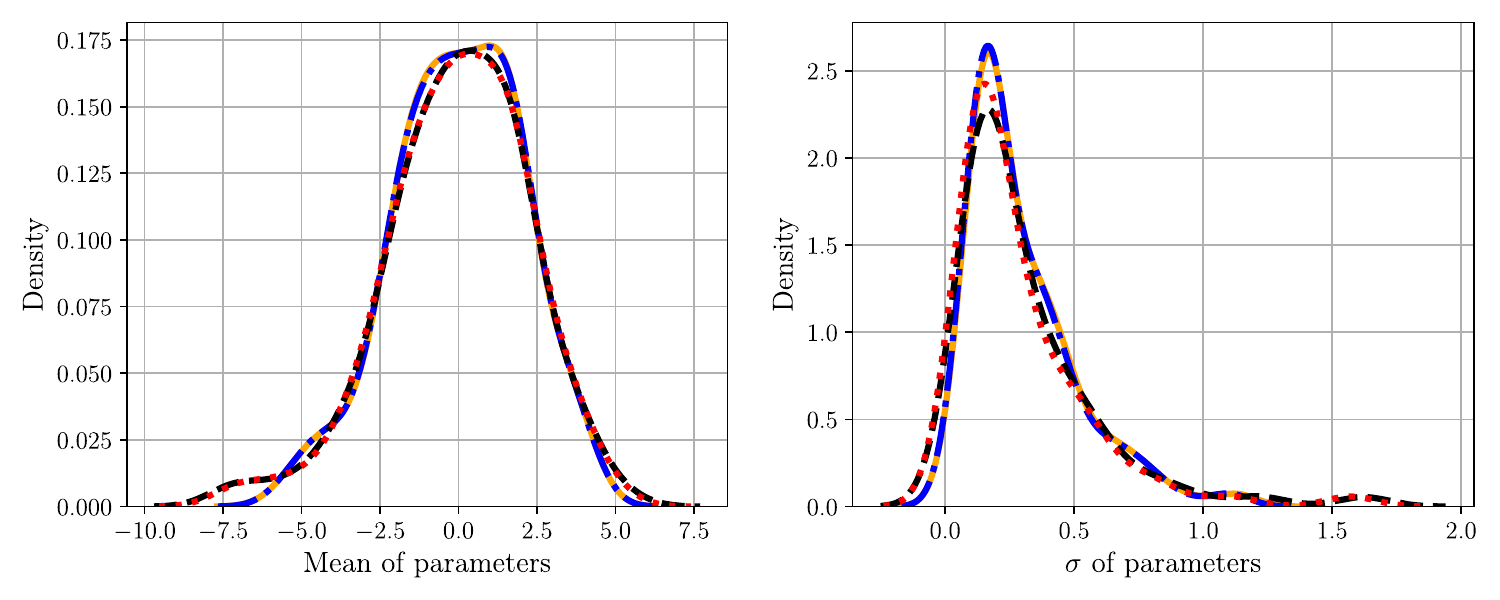}}
    \caption{Kernel Density estimates of the mean and standard deviation of BNN parameters learned for varying noise level $\sigma$ and number of training data points $N_d$. \myline{orange}{} $\sigma = 0.1, N_d=10$; \myline{blue}{dash dot} $\sigma = 0.05, N_d=20$; \myline{black}{dashed} $\sigma = 0.01, N_d=30$; \myline{red}{dotted} $\sigma = 0.005, N_d=40$ }
    \label{fig:hist_noise_nd}
\end{figure}

\begin{figure}[htpb]
    \centering
    \subfigure[Varying noise level]{\includegraphics[width=0.45\linewidth]{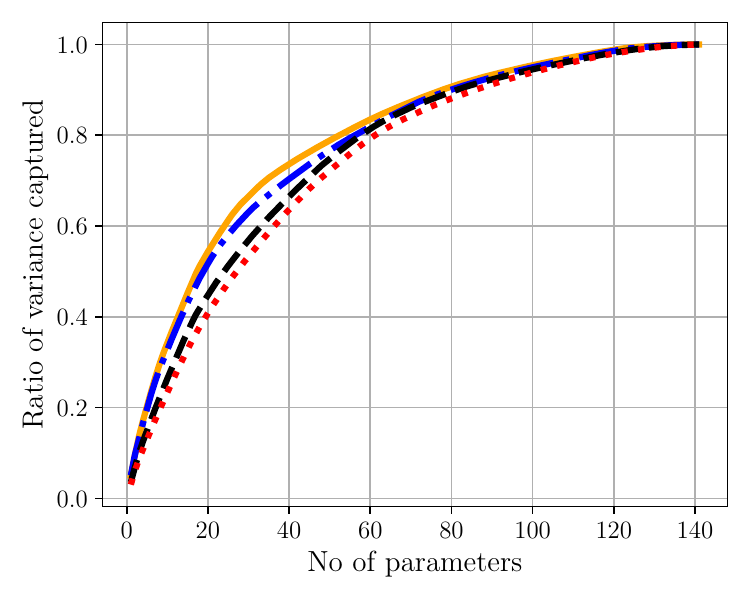}\label{fig:hist_noise}} \hfill
    \subfigure[Varying training data]{\includegraphics[width=0.45\linewidth]{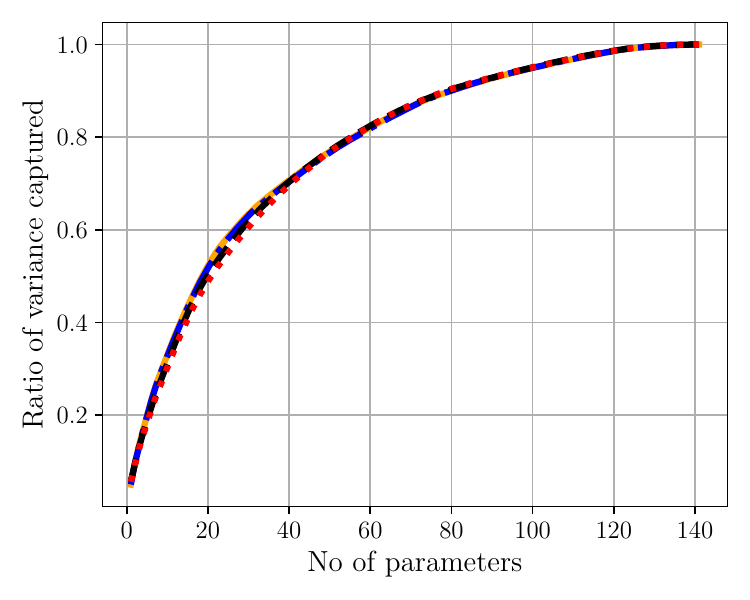}}
    \caption{Ratio of variance captured for different noise level $\sigma$ and number of training data points $N_d$. \myline{orange}{} $\sigma = 0.1, N_d=10$; \myline{blue}{dash dot} $\sigma = 0.05, N_d=20$; \myline{black}{dashed} $\sigma = 0.01, N_d=30$; \myline{red}{dotted} $\sigma = 0.005, N_d=40$ }
    \label{fig:caputred_variance_noise_nd}
\end{figure}

Varying the noise level changes the mean and significantly changes the standard deviation of parameters learned by the Bayesian neural network, whereas varying the number of training data points doesn't alter the distribution of parameters as much, as seen in Figure~\ref{fig:hist_noise_nd}. As a result of this, the ratio of variance captured is more sensitive to varying noise level than varying the training data, as seen in Figure~\ref{fig:caputred_variance_noise_nd}.

        
\section{Effect of the sensitivity threshold} \label{app:tau}
\begin{figure}[htpb]
        \centering
        \subfigure[$\tau = 0.9$]{\includegraphics[width=0.45\linewidth]{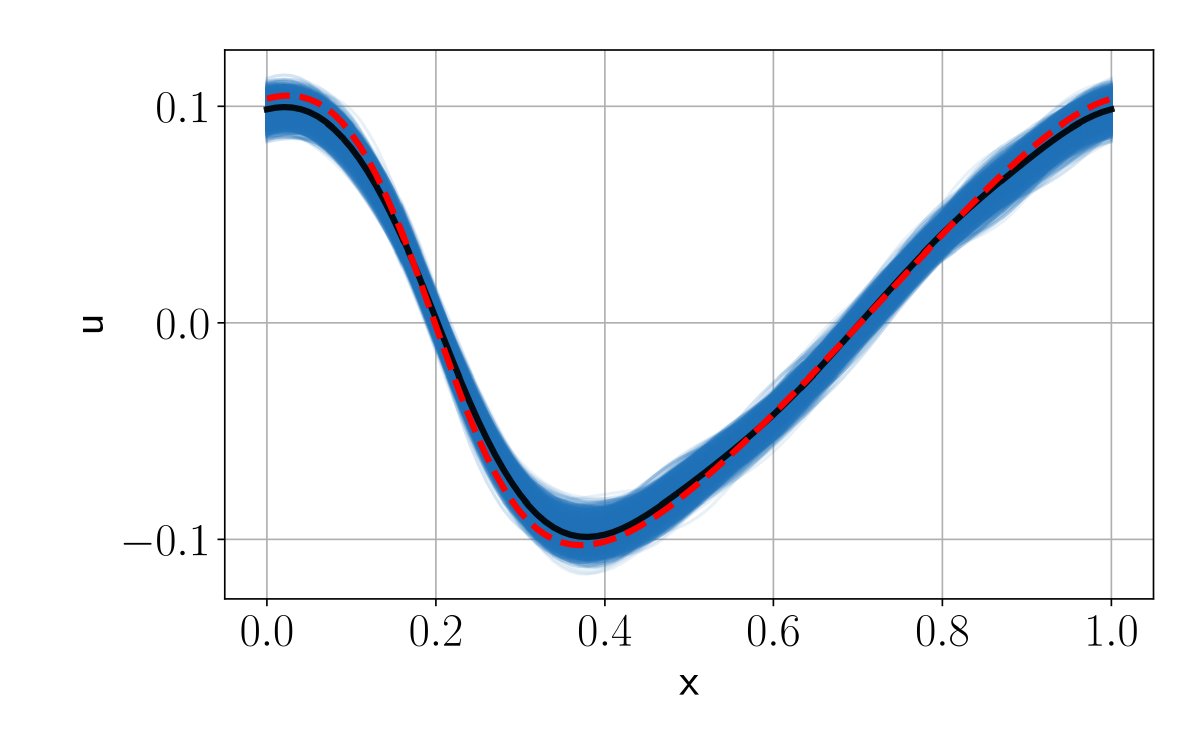}}
        \subfigure[$\tau = 0.3$]{\includegraphics[width=0.45\linewidth]{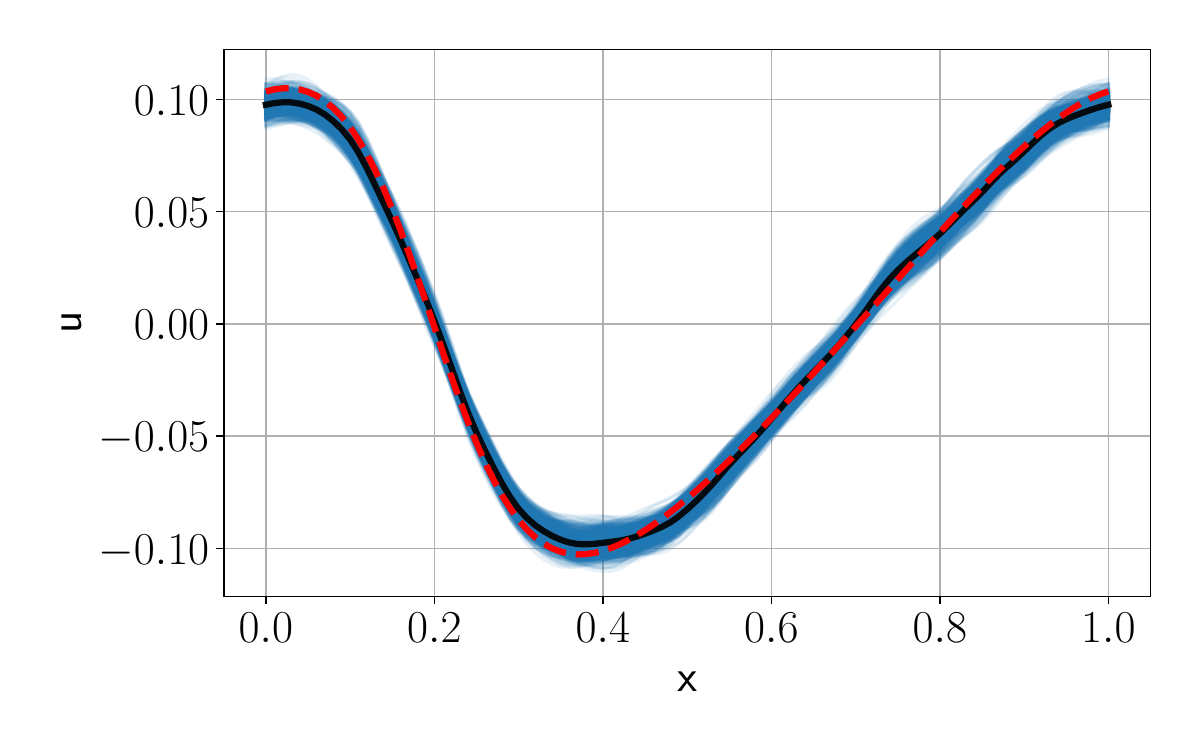}}
        \subfigure[Random]{\includegraphics[width=0.45\linewidth]{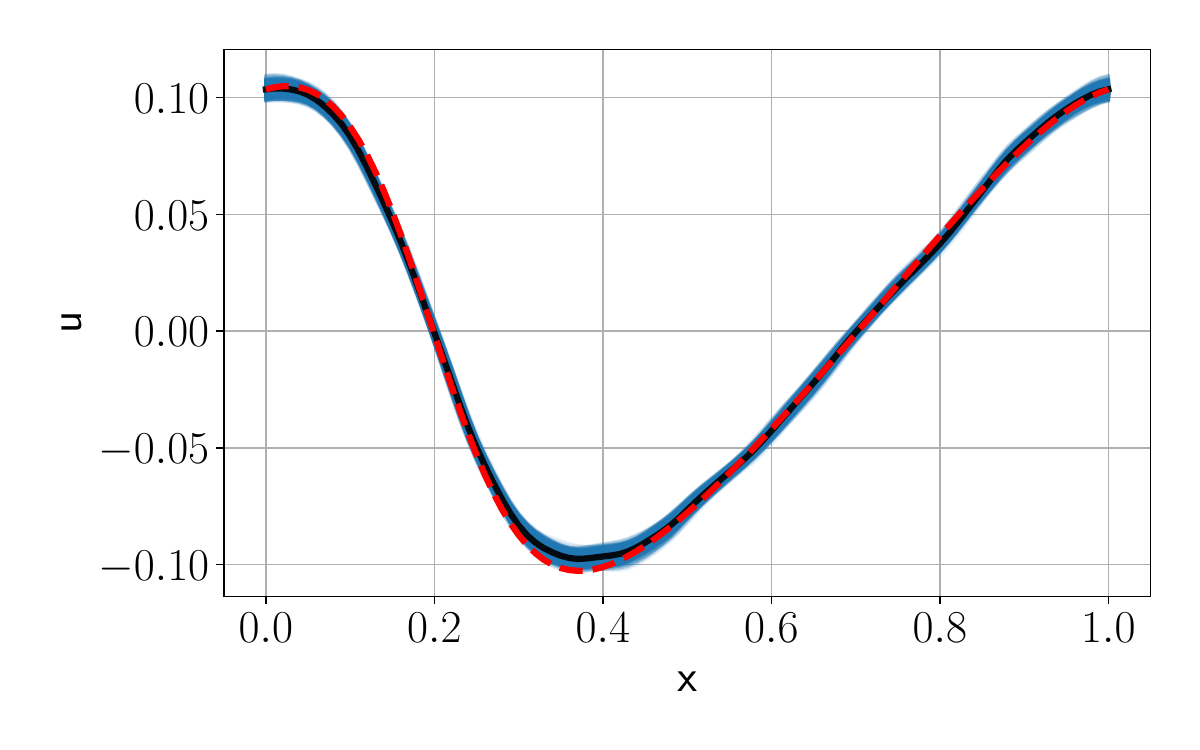}}
        \subfigure[Uncertainty ]{\includegraphics[width=0.45\linewidth]{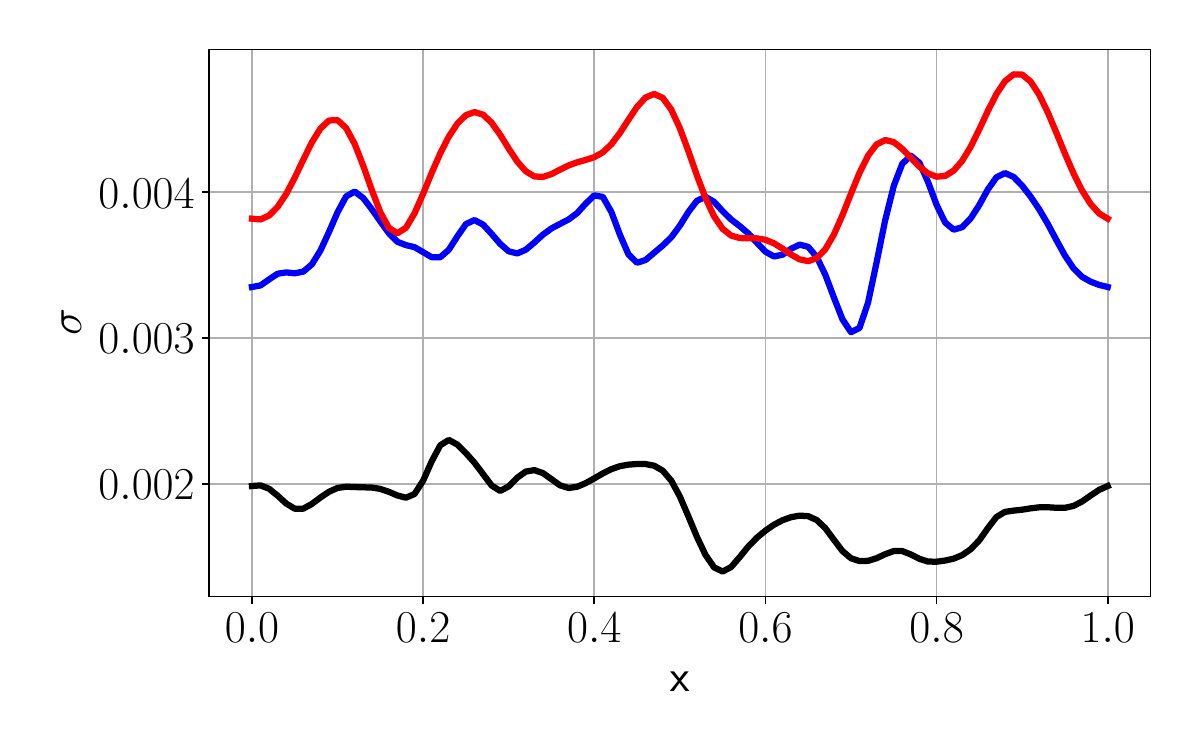}}
        \caption{Comparing predictions for two values of the sensitivity threshold and a randomly chosen subset of the parameters: (a-c) \myline{c0}{} samples, \myline{black}{}  mean prediction, \myline{red}{dashed} true solution. (d) Comparing uncertainties for the three cases \myline{red}{} $\tau=0.9$ \myline{blue}{} $\tau=0.3$, \myline{black}{} Random }
        \label{fig:uncertainty_tau}
    \end{figure}

We conducted experiments with varying sensitivity thresholds, $\tau$, to analyze the performance of the proposed approach on the Burgers dataset. With decreasing $\tau$, the number of parameters considered to perform HMC, i.e., the dimensionality of the parameter space, reduces as expected. Due to this dimensionality reduction, the computational cost to perform HMC also reduces, as summarized in Table~\ref{Tab:cost_tau}. This is evident from the acceptance rate for a fixed step size (column 3) and the step size required to perform HMC for a fixed acceptance rate (column 4). The accuracy slightly improves since more parameters are fixed at mean values of VI with reducing $\tau$. Since the least sensitive parameters are fixed at their mean values, the uncertainties in the prediction also reduce with decreasing $\tau$ as seen in Figure~\ref{fig:uncertainty_tau}. The uncertainties are also compared with randomly choosing 1,279 parameters (the same number of parameters as $\tau = 0.3$) to elucidate the importance of the proposed sensitivity analysis. Importantly, it is evident that the uncertainties in the case of randomly choosing a subset of parameters are \textit{much worse} compared to the uncertainties obtained by the parameters chosen from the sensitivity analysis, even with a small threshold of 0.3.
    \begin{table}[htpb]
        \centering
        \caption{Comparison of the computation cost between varying sensitivity threshold $\tau$ for the Burgers problem. The step size is fixed at $2E-4$ (for column 3), and the acceptance rate is fixed at 80\% (for column 4). }
        \begin{tabular}{cccccc}
        \toprule
          \multirow{2}{*}{$\mathbf{\tau}$} & \multirow{2}{*}{\# of parameters} & \multicolumn{3}{c}{\textbf{Fixed step size}}  &  {\textbf{Fixed acceptance rate}}  \\
          && Acceptance rate & MSE & Time/sample & Required step size  \\ \hline
          0.9 & 28,214 & 87\% & 2.9E-4 & 6.27 s& 5.0E-4 \\ 
          0.75 & 12,944 & 89\% & 2.9E-4 & 6.37 s & 6.9E-4\\
          0.50 & 4,007 & 92 \% & 2.7E-4 & 6.21 s&7.7E-4\\
          0.30 & 1,279 &94 \% &2.0E-4& 6.05 s&8.9E-4\\
            \bottomrule
        \end{tabular}
        \label{Tab:cost_tau}
    \end{table}

\newpage
 \bibliographystyle{elsarticle-num} 
 \bibliography{references}





\end{document}